\documentclass[runningheads]{llncs}
\usepackage{graphicx}

\usepackage{tikz}
\usepackage{comment}
\usepackage{amsmath,amssymb} 
\usepackage{color}
\usepackage{booktabs}
\usepackage[accsupp]{axessibility}  

\usepackage{threeparttable}
\usepackage{bbding}
\usepackage{multirow}
\usepackage{amssymb}
\usepackage{subfigure}
\usepackage{mcode}
\usepackage{wrapfig}
\usepackage[width=122mm,left=12mm,paperwidth=146mm,height=193mm,top=12mm,paperheight=217mm]{geometry}
\usepackage[pagebackref=true,breaklinks=true,letterpaper=true,colorlinks,bookmarks=true]{hyperref}
\usepackage{url}

\newcommand{\et}{\emph{et al.}}
\newcommand{\eg}{e.g.}
\newcommand{\ie}{i.e.}

\begin{document}
\pagestyle{headings}
\mainmatter
\title{Self-Promoted  Supervision  for Few-Shot Transformer} 
\def\ECCVSubNumber{****}
\titlerunning{Self-Promoted  Supervision  for Few-Shot Transformer}
%
\author{Bowen Dong\inst{1} \and
Pan Zhou\inst{2} \and
Shuicheng Yan\inst{2} \and
Wangmeng Zuo\inst{1}
}
\authorrunning{Dong et al.}
%
\institute{Harbin Institute of Technology \and
National University of Singapore\\
\email{\{cndongsky,panzhou3,shuicheng.yan\}@gmail.com, wmzuo@hit.edu.cn}}
\maketitle

\vspace{-1em}
\begin{abstract}
The few-shot learning ability of vision transformers (ViTs) is rarely investigated though heavily desired. In this work, we empirically find that with the same few-shot learning frameworks, \eg~Meta-Baseline,  replacing the widely  used CNN  feature extractor with  a ViT model often severely impairs  few-shot classification performance.  Moreover, our empirical study shows that in the absence of inductive bias,  ViTs often learn the low-qualified token dependencies under few-shot learning regime where only a few labeled training data are available,  which largely contributes to the above performance degradation.  To alleviate this issue, for the first time, we propose a simple yet effective few-shot training framework for ViTs, namely Self-promoted sUpervisioN (SUN).   
Specifically, besides the conventional global supervision for global semantic learning,  SUN  further pretrains the ViT on the few-shot learning dataset and then uses it to generate individual location-specific supervision for  guiding each patch token.  This  location-specific  supervision  tells the ViT which patch tokens are similar or dissimilar and thus   accelerates  token dependency learning. Moreover, it models the local semantics in each patch token to improve the object grounding and recognition capability which helps  learn generalizable patterns. To improve the quality of location-specific supervision, we further propose two techniques:~1) background patch filtration to filtrate background patches out and assign them into an extra background class; and 2) spatial-consistent augmentation to introduce sufficient diversity for data augmentation while keeping the accuracy of the generated local supervisions.    Experimental results show that SUN using ViTs significantly surpasses other  few-shot learning frameworks with ViTs and is the first one that achieves higher performance than those CNN state-of-the-arts. Our code is publicly available at \url{https://github.com/DongSky/few-shot-vit}.
  
\end{abstract}

\section{Introduction}
Vision transformers (ViTs) have achieved great success in the computer vision field, and even surpass corresponding state-of-the-art CNNs on many vision tasks, \eg~image classification~\cite{dosovitskiy2021an,pmlr-v139-touvron21a,Touvron_2021_ICCV,Wang_2021_ICCV,Liu_2021_ICCV,zhang2021aggregating}, object detection~\cite{detr20,YOLOS} and  segmentation~\cite{Zheng_2021_CVPR,Strudel_2021_ICCV}. 
One key factor contributing to ViTs' success is their powerful self-attention mechanism~\cite{vaswani2017attention}, which does not introduce any inductive bias and can better capture the long-range dependencies among local features in the data than the convolution mechanism in CNNs.  
This motivates us to investigate two important problems.
First, we wonder whether ViTs can perform well under few-shot learning setting or not, which aims to recognize objects from novel categories with only a handful of labeled samples as reference. 
Second, if no, how to improve the few-shot learning ability of ViTs?
The few-shot learning ability is heavily desired for machine intelligence, since in practice, many tasks actually have only a few labeled data due to data scarcity  
\begin{figure*}[tb]
	\label{fig:performance}
	\centering
	\subfigure[{1-shot}]{
		\includegraphics[width=2.25in]{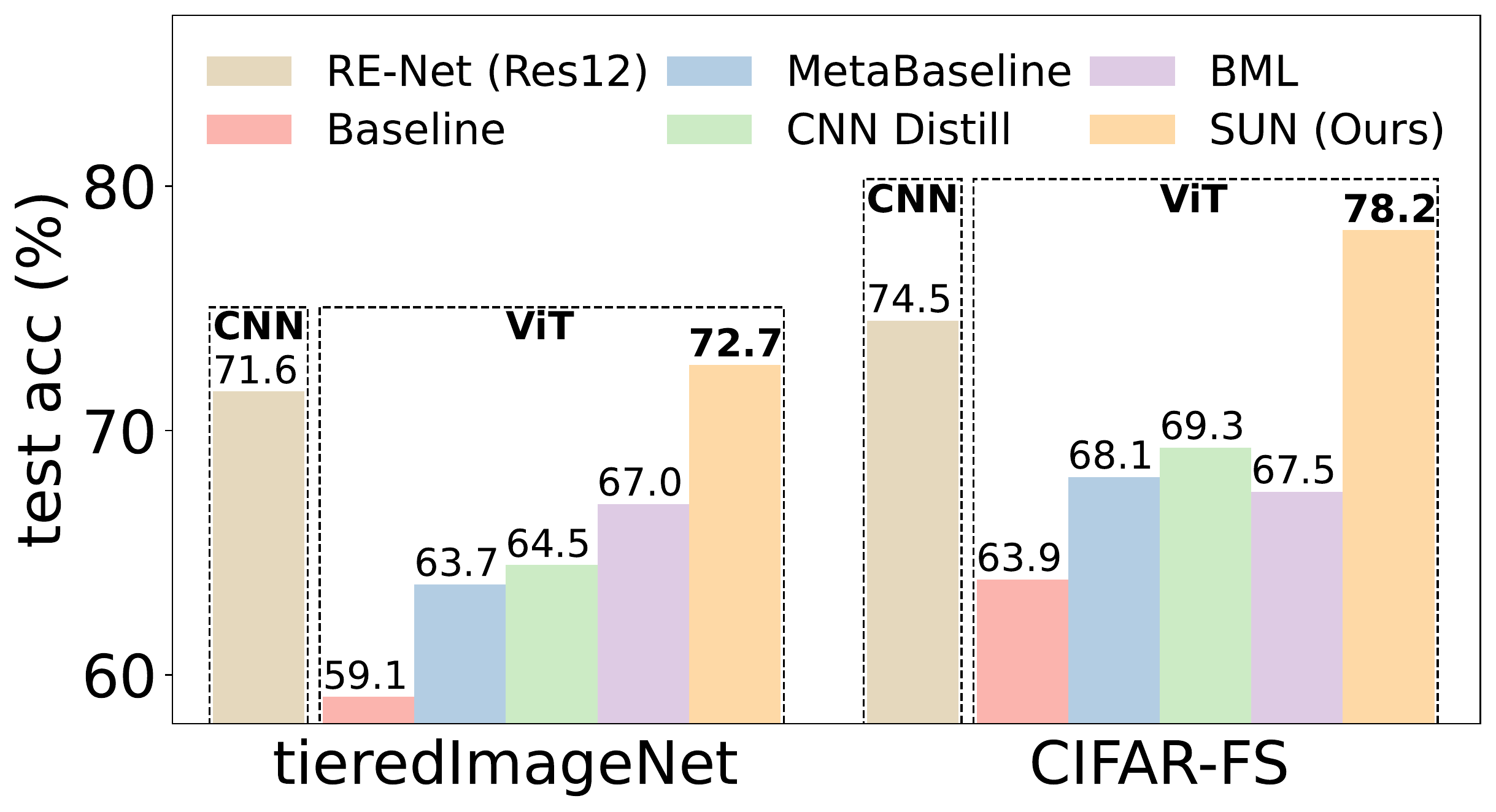}
		\label{fig:per_tiered}
	} \subfigure[{5-shot}]{
		\includegraphics[width=2.25in]{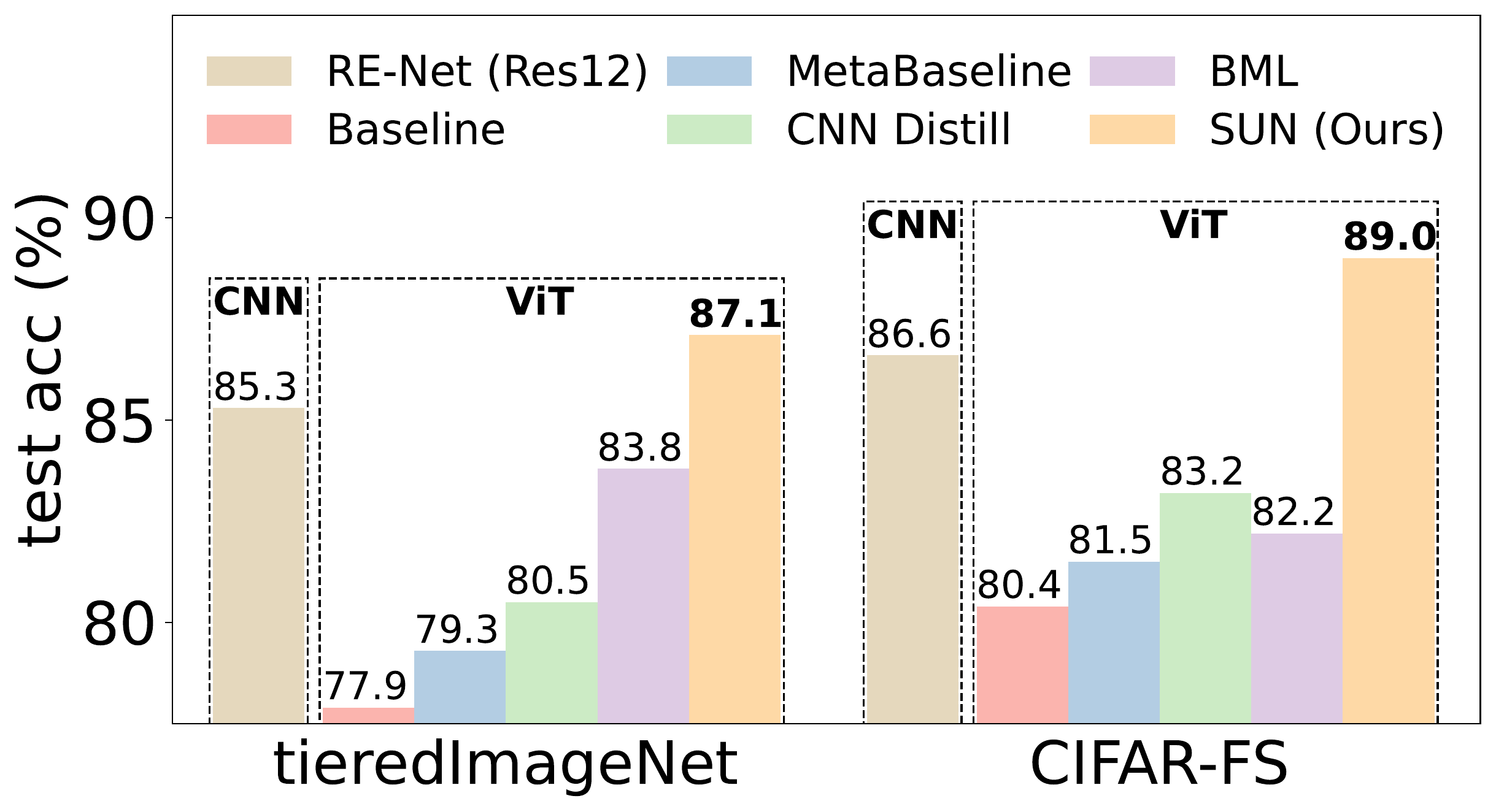}
		\label{fig:per_cifarfs}
	} \\
	\vspace{-1.5em}
	\caption{Performance comparisons of different few-shot classification frameworks.  
	Except the state-of-the-art RE-Net~\cite{Kang_2021_ICCV} which uses ResNet-12~\cite{He2016DeepRL} as the feature extractor, all methods use the same NesT transformer~\cite{zhang2021aggregating} as the feature extractor.
	With same ViT backbone, our SUN significantly surpasses other  baselines and even achieves higher performance than SoTA baseline with CNN backbone. Best viewed in color.}
	\vspace{-2em}
\end{figure*}
(\eg~disease data), high equipment cost to capture data and manual labeling cost (\eg~pixel-level labeling). 
In this work, we are particularly interested in few-shot classification~\cite{vinyals2016matching,FinnAL17,snell2017prototypical} which is a benchmark task to evaluate few-shot learning capacity of a method. 

For the first question, unfortunately, we empirically find that for representative few-shot learning frameworks, \eg~Meta-Baseline~\cite{Chen_2021_ICCV_metabaseline}, replacing the CNN feature extractor by ViTs severely impairs  few-shot classification performance. 
The most possible reason is the lack of inductive bias in ViTs---in absence of any prior inductive bias,  ViTs needs a large amount of data to learn the dependencies among local features, \ie~patch tokens.
Our analysis in Sec.~\ref{sec:analysis} shows that introducing CNN-alike   inductive bias can partly accelerate the token dependency learning in ViTs and thus partially mitigate this effect, which also accords with the observations in~\cite{li2021vision,zhang2021bootstrapping} under supervised learning.
For example, in Fig.~\ref{fig:performance}, the ``Baseline''~\cite{chen2019closerfewshot} and ``Meta-Baseline''~\cite{Chen_2021_ICCV_metabaseline} use vanilla ViTs, while ``CNN Distill'' introduces CNN-alike inductive bias via distilling a pretrained CNN teacher network into the ViT and achieves higher performance.
However, CNN-alike inductive bias cannot be inherently suitable for ViTs and cannot well enhance and accelerate token dependency learning in ViTs,  because ViTs and CNNs have different network architectures, and also the inductive bias in CNNs often cannot well capture long-range dependencies.  
This naturally motivates us to consider how to improve the few-shot learning ability of ViTs.

We therefore propose an effective few-shot learning  framework for ViTs, namely, \emph{Self-promoted sUpervisioN} (SUN for short). 
SUN enables a ViT model to generalize well on novel unseen categories with a few labeled training data.
The key idea of SUN is to enhance and accelerate dependency learning among different tokens, \ie,  patch tokens, in ViTs. 
This idea is intuitive, since if ViTs can learn the token dependencies fast and accurately, and thus naturally can learn from less labeled training data, which is consistent with few-shot learning scenarios. 
In particular, at the meta-training phase, SUN provides global supervision to global feature embedding and further employs individual location-specific supervision to guide each patch token.
Here SUN first trains the ViT on the   training data in few-shot learning, and uses it to generate patch-level pseudo labels as location-specific supervision. 
This is why we call our method ``Self-Promoted Supervision" as it uses the same ViT to generate local supervision.  
To improve the quality of patch-level supervision, we further propose two techniques, namely 1) background patch filtration and 2) spatial-consistent augmentation. 
The former aims to alleviate effects of  bad cases where background patches are wrongly assigned to a semantic class and have incorrect local supervision; the latter is to introduce sufficient diversity to data augmentation while keeping the accuracy of generated   location-specific supervisions.
Our location-specific dense supervision benefits ViT on few-shot learning tasks from two aspects.
Firstly, considering the  location-specific supervisions on all tokens are consistent, \ie~similar pseudo labels for similar local tokens,  this dense supervision tells ViT which patch tokens are similar or dissimilar and thus can accelerate ViT to learn high-quality token dependencies. 
Secondly, the local semantics in each patch token are also well modeled to improve the object grounding and recognition capabilities of ViTs.
Actually, as shown in~\cite{zhong2020random,jiang2021all}, modeling semantics in local tokens can avoid learning skewed and non-generalizable patterns and thus substantially improve the model generalization performance which is also heavily desired in few-shot learning. 
So both aspects can improve the few-shot learning capacity of ViTs.
Next, at the meta-tuning phase, similar to existing methods~\cite{Chen_2021_ICCV_metabaseline,ye2020fewshot}, SUN adapts the knowledge of ViT trained at the meta-training phase to new tasks by fine-tuning ViT on the corresponding training data.

Extensive experimental results demonstrate that with the same ViT architecture, our SUN framework  significantly outperforms all baseline frameworks on the few-shot classification tasks,  as shown in Fig.~\ref{fig:performance}. 
Moreover, to the best of our knowledge,  SUN with ViT backbones is the first work that applies ViT  backbones to few-shot classification, and is also the first ViT method that achieves comparable even higher performance than state-of-the-art baselines using similar-sized CNN backbones, as illustrated in  Fig.~\ref{fig:performance}. 
This well shows the superiority of SUN, since CNN has high inductive bias and is actually much more suitable than ViT on few-shot learning problems. 
Additionally, our SUN also provides a simple yet solid baseline for few-shot classification. 

\vspace{-1em}
\section{Related Work}

\vspace{-0.5em}
\subsubsection{Vision Transformer.} 
Unlike CNNs that use convolutions to capture local information in images, ViTs~\cite{dosovitskiy2021an} model long-range interaction explicitly, and  have shown great potential for vision tasks.   
However, to achieve comparable performance with CNN of a similar model size trained on ImageNet~\cite{deng2009imagenet} , ViTs need huge-scale datasets to train~\cite{deng2009imagenet,sun2017revisiting},
greatly increasing computational cost.  
To alleviate this issue and further improve  performance, many works~\cite{pmlr-v139-touvron21a,jiang2021all,Yuan_2021_ICCV,xiao2021early,Liu_2021_ICCV,zhang2021aggregating}

\noindent\cite{Wu_2021_ICCV_CvT,yu2021metaformer,tolstikhin2021mlp,li2021localvit,liu2021pay,Wang_2021_ICCV,Xu_2021_ICCV,YuanFHLZCW21} propose effective solutions. 
However, with these advanced techniques, ViTs still need to train on large-scale datasets, \eg~ImageNet~\cite{deng2009imagenet} or JFT-300M~\cite{sun2017revisiting}, and often fail on small datasets~\cite{zhang2021aggregating,li2021vision,zhang2021bootstrapping}. 
Recently, a few works~\cite{li2021vision,cao2022training} also propose ViTs for small datasets.
For instance, Li~\et~\cite{li2021vision} initialized ViT from a well-trained CNN to borrow inductive bias in CNN to ViT.
Meanwhile, Cao~\et~\cite{cao2022training} first used instance discriminative loss to pretrain ViT and then fine-tuned it on the target dataset.
In comparison, we propose self-promoted supervision to provide dense local supervisions, enhancing token dependency learning and alleviating the data-hungry issue.  
Moreover, we are interested in few-shot learning problems that focus on initializing ViTs with small training data from base classes and generalizing on novel unseen categories with a few labeled data. 

\vspace{-1.5em}
\subsubsection{Few-Shot Classification.}  
Few-shot classification methods can be coarsely divided into three groups: 1) \emph{optimization-based  methods}~\cite{FinnAL17,rusu2018metalearning,Lee2019MetaLearningWD,zhou2019metalearning,zhou2021task} for fast adaptation to new tasks; 
2) \emph{memory-based methods}~\cite{Ravi2017OptimizationAA,mishra2018a,graves2014neural} for storing important training samples; 
and 3) \emph{metric-based methods}~\cite{snell2017prototypical,Chen_2021_ICCV_metabaseline,vinyals2016matching,ye2020fewshot} which learn the similarity metrics among samples.
This work follows the pipeline of metric-based methods for the simplicity and high accuracy.  
It is worth noting that some recent works~\cite{Doersch_NEURIPS2020_CTX,Hou_NEURIPS2019_CAN,ye2020fewshot,zhmoginov2022hypertransformer,liu2021a} also integrate transformer layers with metric-based methods for few shot learning. 
%
%
Different from previous works~\cite{Doersch_NEURIPS2020_CTX,Hou_NEURIPS2019_CAN,ye2020fewshot,zhmoginov2022hypertransformer,liu2021a} which leverage transformer layers as few-shot classifiers, 
our work focuses on improving the few-shot classification accuracy of ViT backbones.  

\vspace{-1.0em}
\section{Empirical Study of ViTs for Few-Shot Classification}\label{sec:preliminaries}

Here we first introduce the few-shot classification task. 
Given a labeled base dataset $\mathbb{D}_{\text{base}}$ which contains base classes $C_{\text{base}}$, this task aims to learn a meta-learner $f$ (feature extractor) such that $f$ can be fast adapted to unseen classes $C_{\text{novel}}$ (\ie~$C_{\text{base}}\cap C_{\text{novel}}=\varnothing$) in which each class has a few labeled training samples. 
This task is a benchmark to evaluate few-shot learning capacity of a method, since it needs to well learn the knowledge in the base data $\mathbb{D}_{\text{base}}$ and then fast adapt the learnt knowledge to the new data $C_{\text{novel}}$ via only a few training data. 
Next, we investigate the ViTs' performance on this important task and also analyze the potential reasons for their performance. 

\noindent\textbf{Performance Investigation of ViTs. }\label{sec:preexp} 
We investigate the few-shot learning performance of various ViTs, including LV-ViT~\cite{jiang2021all} (single stage ViT),  Le-ViT~\cite{Graham_2021_ICCV} (multi-stage ViT), Visformer~\cite{Chen_2021_ICCV} (CNN-enhanced ViT), Swin~\cite{Liu_2021_ICCV} and NesT~\cite{zhang2021aggregating} (locality-enhanced ViT).
For a fair comparison, we scale the depth and width of these ViTs such that their model sizes are similar to ResNet-12~\cite{He2016DeepRL} ($\sim$12.5M parameters) which is the most commonly used architecture and achieves (nearly) state-of-the-art performance on few-shot classification tasks.  

\begin{table}[t]
    \centering
    \caption{Classification accuracy $(\%)$ of the Meta-Baseline few-shot learning framework using ViTs and ResNet-12 as a feature extractor on \emph{mini}ImageNet.
    }
    \label{table:preexp}
    \vspace{-1em}
    \small
    \setlength{\tabcolsep}{4.6pt} 
   \renewcommand{\arraystretch}{2.9}
 { \fontsize{8.3}{3}\selectfont{
    \begin{threeparttable}
    \begin{tabular}{llcccccc}
\toprule
  &  \multirow{2}{*}{{Backbone}}&\multirow{2}{*}{{Params}}&\multicolumn{2}{c}{\textbf{Meta-Training Phase}}&\multicolumn{2}{c}{\textbf{Meta-Tuning Phase}}
    \\ \cline{4-7}
   & &&{{5-way 1-shot}}&{{5-way 5-shot}}&{{5-way 1-shot}}&{{5-way 5-shot}}
    \\ \toprule
CNN &     ResNet-12&12.5M&\textbf{60.00$\pm$0.44}&\textbf{80.55$\pm$0.31}&\textbf{64.53$\pm$0.45}&\textbf{81.41$\pm$0.31}
    \\ \hline
&    {LV-ViT}&13.5M&{43.08$\pm$0.38}&{59.03$\pm$0.39}&{43.07$\pm$0.39}&{58.87$\pm$0.39}\\
   & {Le-ViT}&12.6M&{38.89$\pm$0.39}&{53.51$\pm$0.37}&{40.65$\pm$0.40}&{54.23$\pm$0.38}\\
    ViT &{Swin}&12.3M&{48.26$\pm$0.42}&{65.72$\pm$0.38}&{54.63$\pm$0.45}&{70.60$\pm$0.38}\\
    & {Visformer}&12.4M&{43.55$\pm$0.38}&{60.49$\pm$0.39}&{47.61$\pm$0.43}&{63.00$\pm$0.39}\\
    & {NesT}&12.6M&{49.23$\pm$0.43}&{66.57$\pm$0.39}&{54.57$\pm$0.46}&{69.85$\pm$0.38} \\
       \bottomrule
    \end{tabular}
    \end{threeparttable}}}
    \vspace{-1.5em}
\end{table}
  \begin{figure*}[tb]
  	\label{fig:pre_acc}
  	\centering
  	\subfigure[base \emph{train} acc]{
  		\includegraphics[width=1.1in]{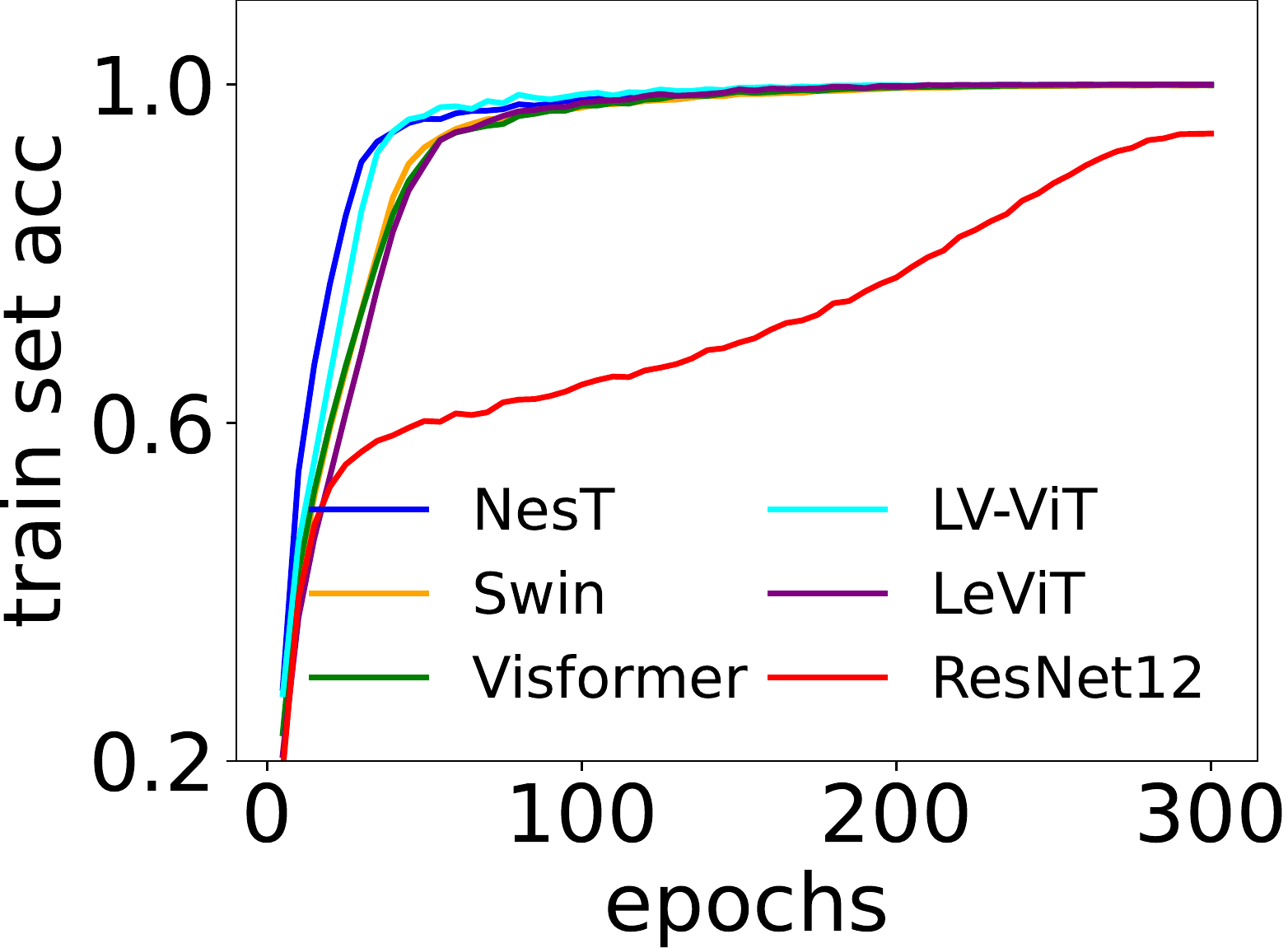}
  		\label{fig:pre_train}
  	} \hspace{-2mm} \subfigure[base \emph{val} acc]{
  		\includegraphics[width=1.1in]{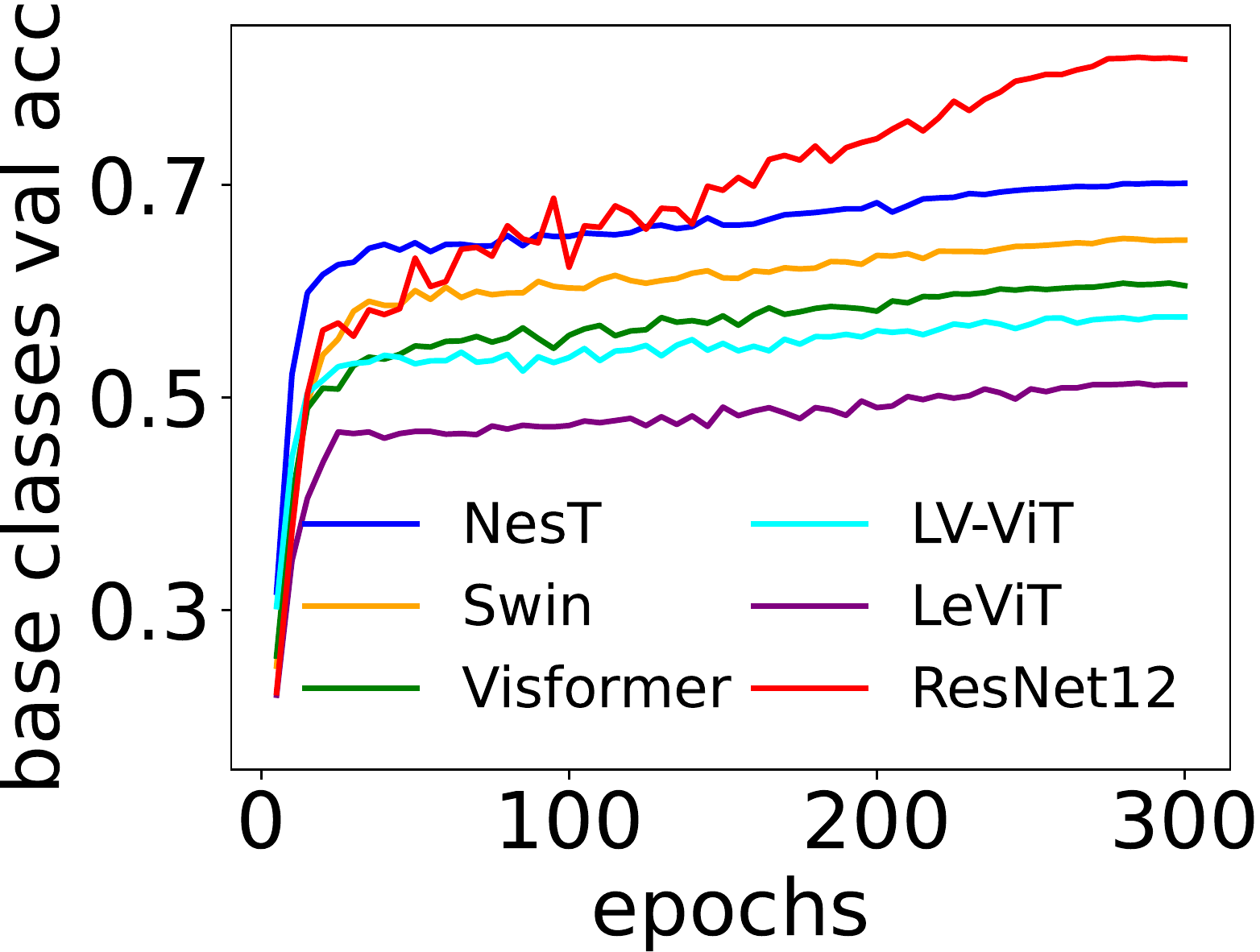}
  		\label{fig:pre_val}
  	} \hspace{-2mm} \subfigure[\emph{train}  1-shot acc]{
  		\includegraphics[width=1.1in]{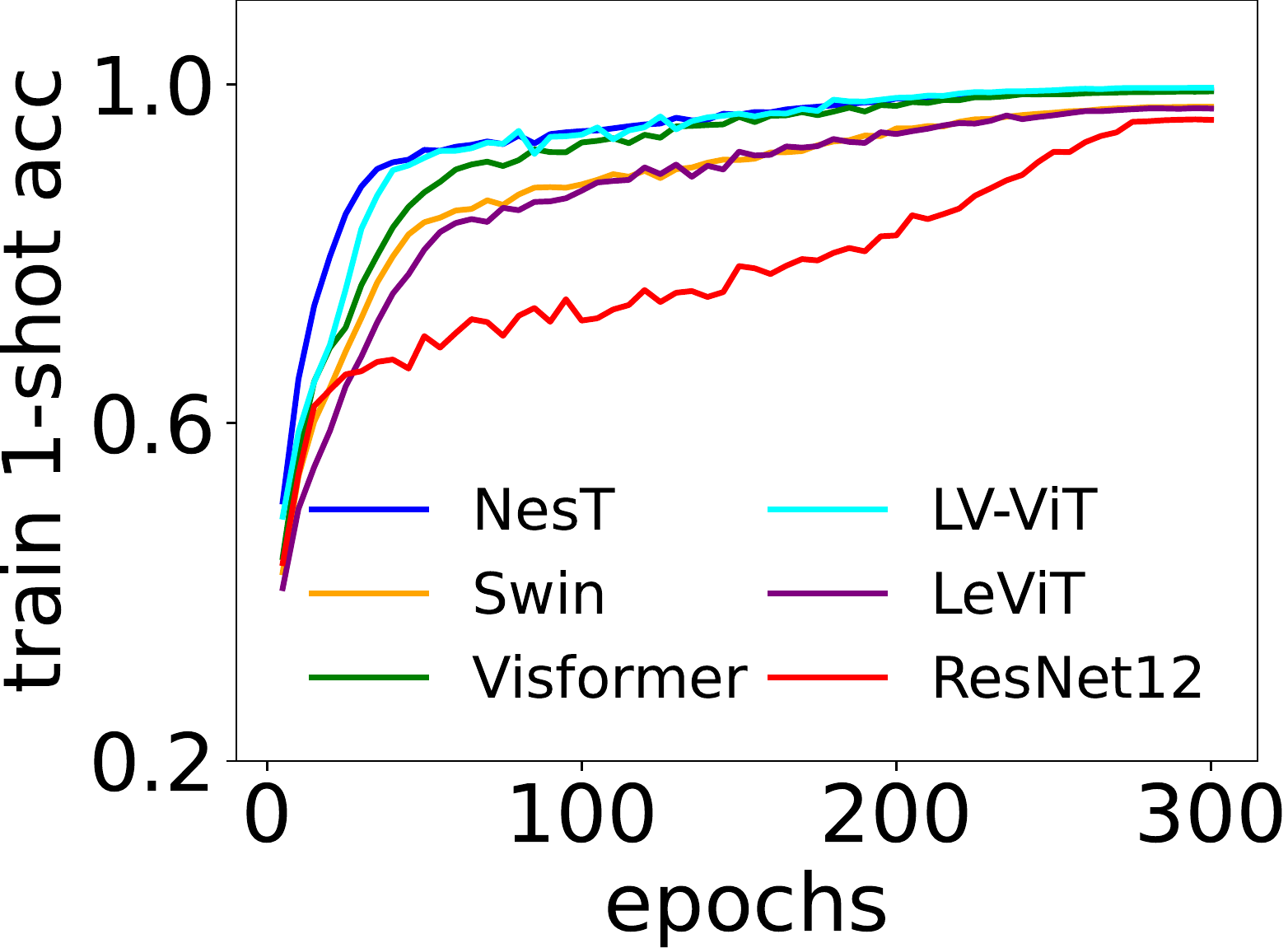}
  		\label{fig:pre_1shot}
  	} \hspace{-2mm} \subfigure[\emph{test} 1-shot acc]{
  		\includegraphics[width=1.1in]{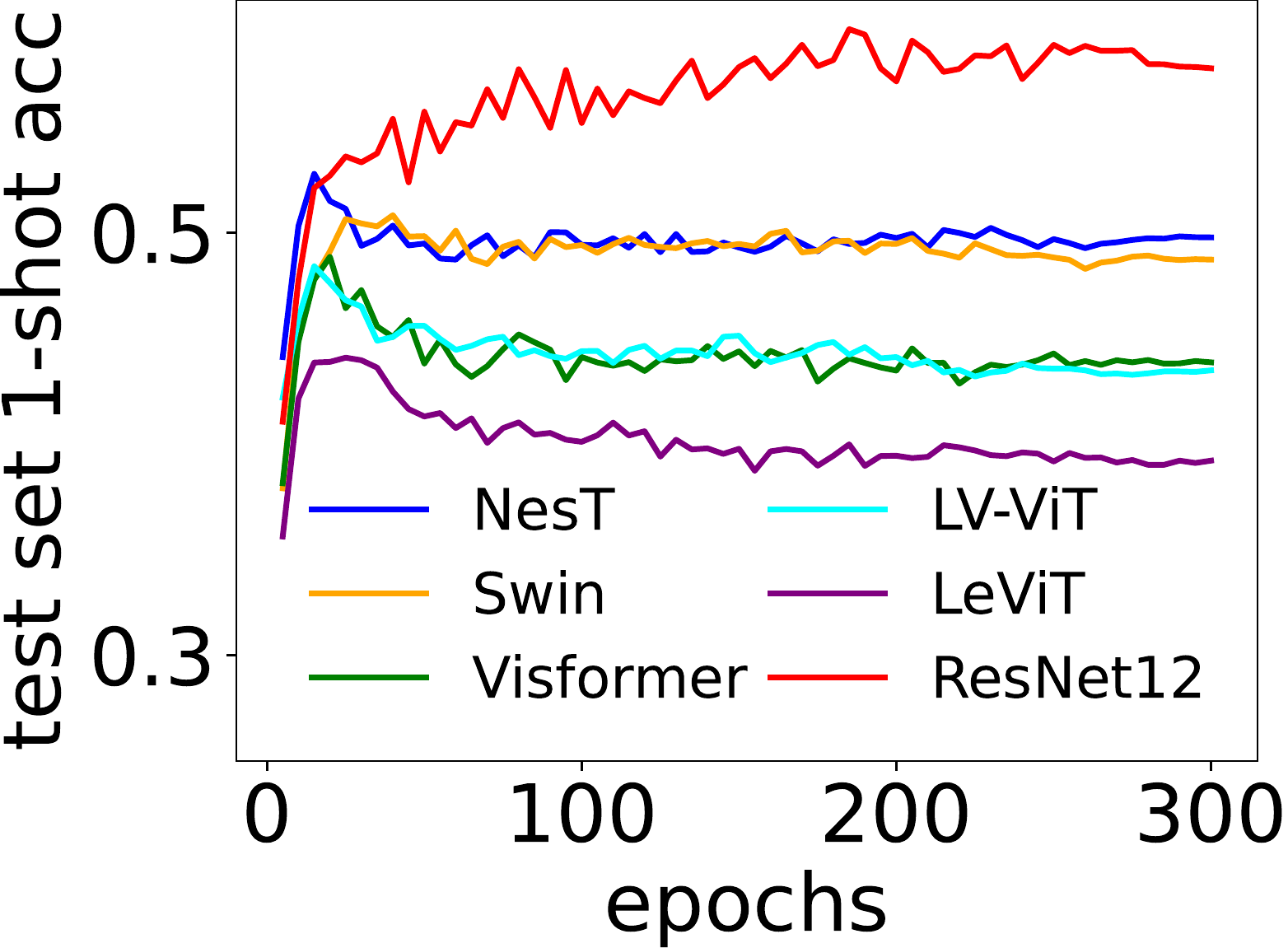}
  		\label{fig:pre_5shot}
  	} \\
  	\vspace{-1.5em}
  	\caption{Accuracy curve comparison of different feature extractors in Meta-Baseline.
    }
  	\label{fig:all}
  	\vspace{-2.5em}
  \end{figure*}

Here we choose a simple yet effective Meta-Baseline~\cite{Chen_2021_ICCV_metabaseline} as the few-shot learning framework to test the performance of the above five ViTs. 
Specifically, in the meta-training phase, Meta-Baseline performs conventional supervised classification; in meta-tuning phase, given a training task, it computes the class prototypes by averaging the sample features from the same class on the support set, and then minimizes the cosine distance between the sample feature and the corresponding class prototypes.
For inference, similar to meta-tuning phase, given a test task $\tau$, it first calculates the class prototypes on the support set of $\tau$, and assigns a class label to a test sample according to cosine distance between the test sample feature and the  class prototypes.  
See more details in Sec.~\ref{meta-tune}.  
 
Table~\ref{table:preexp} summarizes the experimental results on \emph{mini}ImageNet~\cite{vinyals2016matching}.  
Here, ``Meta-Training Phase" means we test ViTs after meta-training phase in Meta-Baseline, while ``Meta-Tuning Phase" denotes the test performance of ViTs after meta-tuning phase.   
Frustratingly, one can observe that even with the same Meta-Baseline framework, all ViTs perform much worse than ResNet-12 after meta-training  phase.
Moreover, after meta-tuning phase, all ViTs still suffer from much worse few-shot learning performance. 
These observations inspire us to investigate what happens during the meta-training period of ViTs and why ViTs perform much worse than CNNs. 

\noindent\textbf{Analysis. }\label{sec:analysis}
To analyze performance of ViTs in the meta-training phase, Fig.~\ref{fig:all}  plots accuracy curves on training and validation sets in base dataset $\mathbb{D}_{\text{base}}$, and reports the 5-way  1-shot accuracy on the training and test sets of the novel classes.  
According to Fig.~\ref{fig:pre_train} and~\ref{fig:pre_val}, all ViTs converge well on training and validation sets in the base dataset $\mathbb{D}_{\text{base}}$. 
Fig.~\ref{fig:pre_1shot} shows that ViTs also converge well on the training data of novel categories during the whole meta-training phase, while Fig.~\ref{fig:pre_5shot} indicates that ViTs  generalize poorly to the test data of novel categories.   
For instance, though NesT~\cite{zhang2021aggregating} achieves $\sim$52\% accuracy in the first 30 meta-training epochs, its accuracy drops to 49.2\% rapidly around the 30th  meta-training epochs.  
In contrast, ResNet-12 maintains nearly 60\%  1-shot accuracy in the last 100 meta-training epochs. 
These observations show that compared to CNN feature extractors, ViTs often suffer from generalization issues on novel categories though they enjoy good generalization ability on $\mathbb{D}_{\text{base}}$.

\begin{table}[t]
    \centering
    \caption{Empirical analysis of ViTs on two datasets. (Top) shows the effect of inductive bias ({IB}) to few-shot accuracy of ViTs; (Middle) studies whether CNN-alike inductive bias ({CaIB}) can enhance token dependency learning; and (Bottom) analyzes local attention ({LA}) in enhancing token dependency learning. 
    }
    \label{table:advanced_analysis}
    \vspace{-1em}
    \small
    \setlength{\tabcolsep}{2.8pt} 
   \renewcommand{\arraystretch}{3}
 { \fontsize{8.3}{3}\selectfont{
    \begin{threeparttable}
    \begin{tabular}{lcc|cccc}
    \toprule
    \multirow{2}{*}{\textbf{Method}}&\multirow{2}{*}{\textbf{ViT}}&\multirow{2}{*}{\textbf{Var}}&\multicolumn{2}{c}{\textbf{\emph{mini}ImageNet}}&\multicolumn{2}{c}{\textbf{\emph{tiered}ImageNet}}
    \\ \cline{4-7}
    &&&{{5-way 1-shot}}&{{5-way 5-shot}}&{{5-way 1-shot}}&{{5-way 5-shot}}
    \\ \hline     \hline
    {(Top)}&&{IB}\\
    \hline

    {Meta-Baseline}~\cite{Chen_2021_ICCV_metabaseline}&NesT&&{54.57$\pm$0.46}&{69.85$\pm$0.38}&\textbf{63.73$\pm$0.47}&{79.33$\pm$0.38} \\
    \cite{Chen_2021_ICCV_metabaseline}+{CNN}~\cite{weng2021semisupervised}&NesT&\checkmark&\textbf{57.91$\pm$0.48}&\textbf{73.31$\pm$0.38}&{63.46$\pm$0.51}&\textbf{80.13$\pm$0.38} \\
    \hline
    \hline
    {(Middle)}&&{CaIB}\\
    \hline
    {Meta-Baseline}~\cite{Chen_2021_ICCV_metabaseline}&NesT&&{54.57$\pm$0.46}&{69.85$\pm$0.38}&{63.73$\pm$0.47}&{79.33$\pm$0.38} \\
    \cite{Chen_2021_ICCV_metabaseline}+{CNN Distill}&NesT&\checkmark&\textbf{55.79$\pm$0.45}&\textbf{71.81$\pm$0.37}&\textbf{64.48$\pm$0.50}&\textbf{80.43$\pm$0.37} \\
    \hline
    \hline
    {(Bottom)}&&{LA}\\
    \hline
    {Meta-Baseline}~\cite{Chen_2021_ICCV_metabaseline}&LeViT&&{38.89$\pm$0.39}&{53.51$\pm$0.37}&{55.75$\pm$0.49}&{71.99$\pm$0.39} \\
    {Meta-Baseline}~\cite{Chen_2021_ICCV_metabaseline}&Swin&\checkmark&\textbf{54.63$\pm$0.45}&\textbf{70.60$\pm$0.38}&{62.68$\pm$0.50}&{78.52$\pm$0.38} \\
    {Meta-Baseline}~\cite{Chen_2021_ICCV_metabaseline}&NesT&\checkmark&{54.57$\pm$0.46}&{69.85$\pm$0.38}&\textbf{63.73$\pm$0.47}&\textbf{79.33$\pm$0.38} \\
    \toprule
    \end{tabular}
    \end{threeparttable}}}
    \vspace{-2.5em}
\end{table}

Next, we empirically analyze why ViTs perform worse than CNNs on few-shot classification. 
The most possible explanation is that the lack of inductive bias leads to the performance degeneration. 
To illustrate this point, we introduce inductive bias (in CNN) into ViT through three ways. \textbf{a)} For each stage, we use a ViT branch and a CNN branch to independently extract image features, and combine their features for fusion.  
See  implementation details in  Appendix~\ref{sec:implement_analysis}.
In this way, this mechanism inherits the inductive bias from the CNN branch.  
Table~\ref{table:advanced_analysis} (Top) shows that by introducing this CNN inductive bias, the new NesT (\ie~``\cite{Chen_2021_ICCV_metabaseline}+CNN") largely  surpasses the vanilla NesT  (``Meta-Baseline~\cite{Chen_2021_ICCV_metabaseline}"). 
\textbf{b)} We train a CNN model on $\mathbb{D}_{\text{base}}$, and use it to teach ViT via  knowledge distillation~\cite{hinton2015distill}. 
This method, \ie~``\cite{Chen_2021_ICCV_metabaseline}+CNN Distill'' in  Table~\ref{table:advanced_analysis} (Middle),  can well introduce the CNN-alike inductive bias into ViT to enhance dependency learning, 
 and improves the vanilla``Meta-Baseline~\cite{Chen_2021_ICCV_metabaseline}'' by a significant margin.  
 To analyze the quality of token dependency, we   visualize the attention maps of the last block of vanilla ViT and CNN-distilled ViT in Fig.~\ref{fig:vis_distill_attn}. One can find that CNN-distilled ViT can learn better token dependency than vanilla ViT, since the former captures almost all semantic tokens while the later one only captures a few semantic patches. 
%
\textbf{c)} Local attention enjoys better few-shot learning performance. 
Table~\ref{table:advanced_analysis} (Bottom) shows that with the same Meta-Baseline~\cite{Chen_2021_ICCV_metabaseline} framework,  Swin and NesT achieve much  higher   accuracy than LeViT, since local attention in Swin and NesT introduces (CNN-alike) inductive bias and can enhance token dependency learning than the global attention in LeViT.    
All these results show that inductive bias benefits  few-shot classification.
%
%

%
%

\begin{figure}
  \centering
  \includegraphics[width=\textwidth]{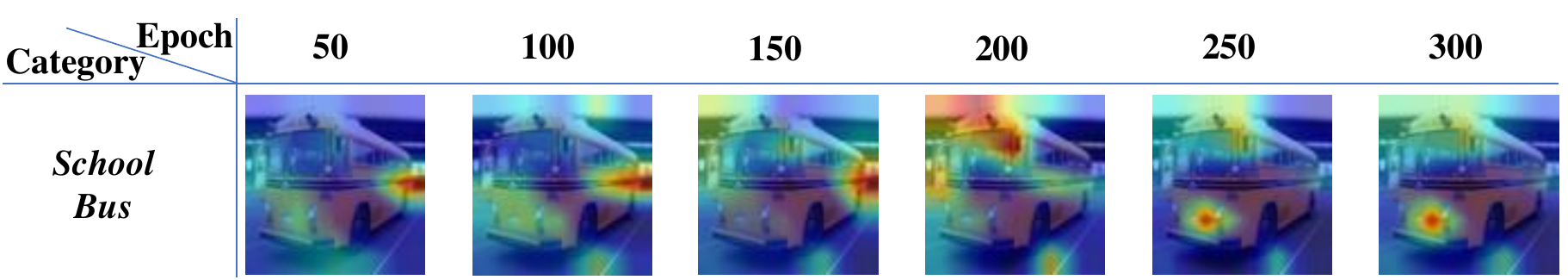}
  \vspace{-2.5em}
  \caption{Visualization of attention maps from vanilla ViTs with different training epochs.}
  \label{fig:per_epoch_baseline}
\end{figure}

\begin{figure}[tb]
	\vspace{-1em}
	\centering
  \subfigure[Attention maps]{\includegraphics[width=.244\textwidth]{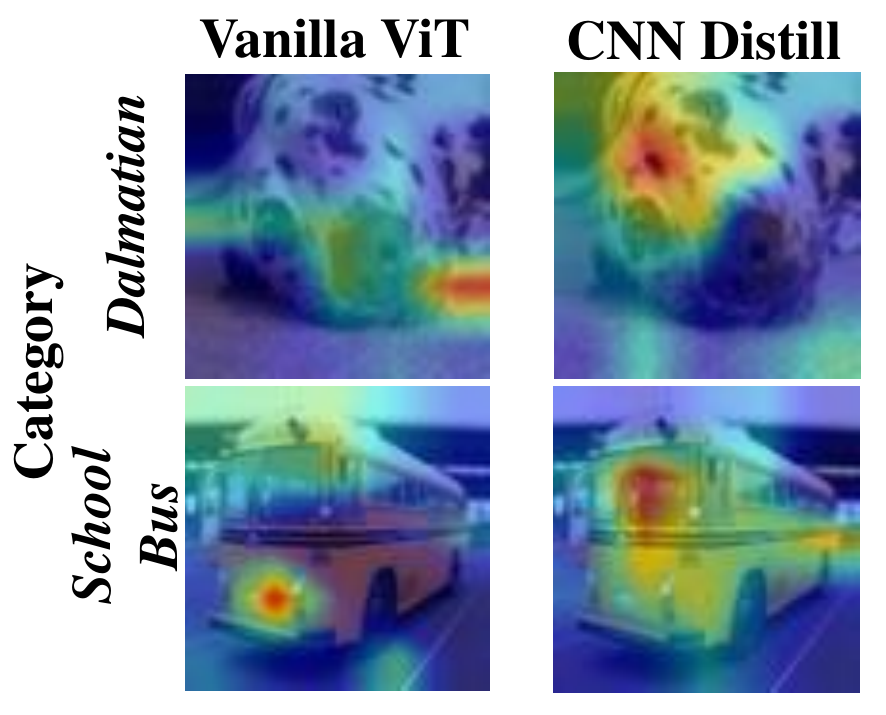}
		\label{fig:vis_distill_attn}}\hspace{-2mm}
  \subfigure[base \emph{train} acc]{\includegraphics[width=.244\textwidth]{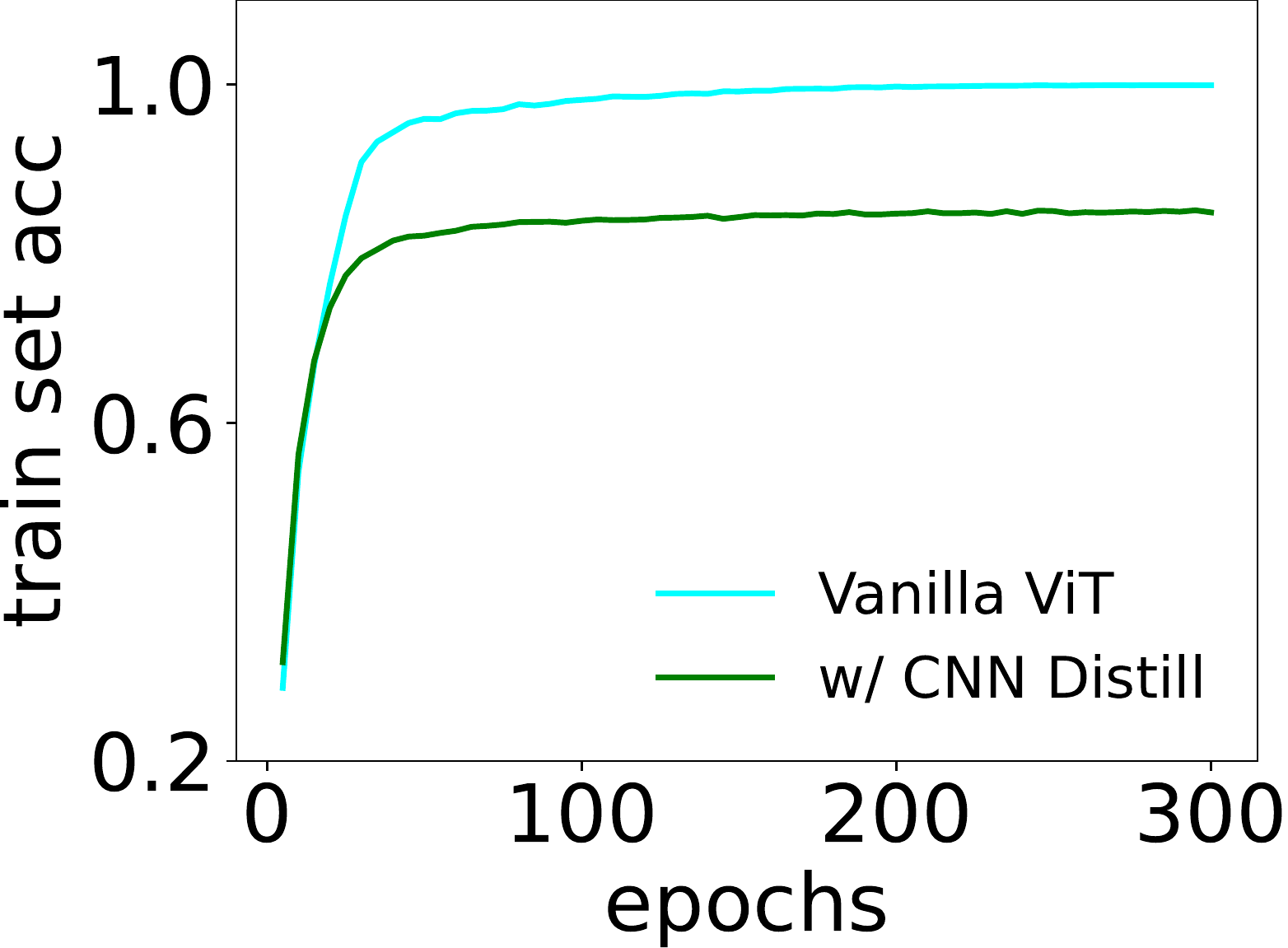}
		\label{fig:pre_train_acc_distill}}\hspace{-2mm}
    \subfigure[attn dist (middle)]{\includegraphics[width=.244\textwidth]{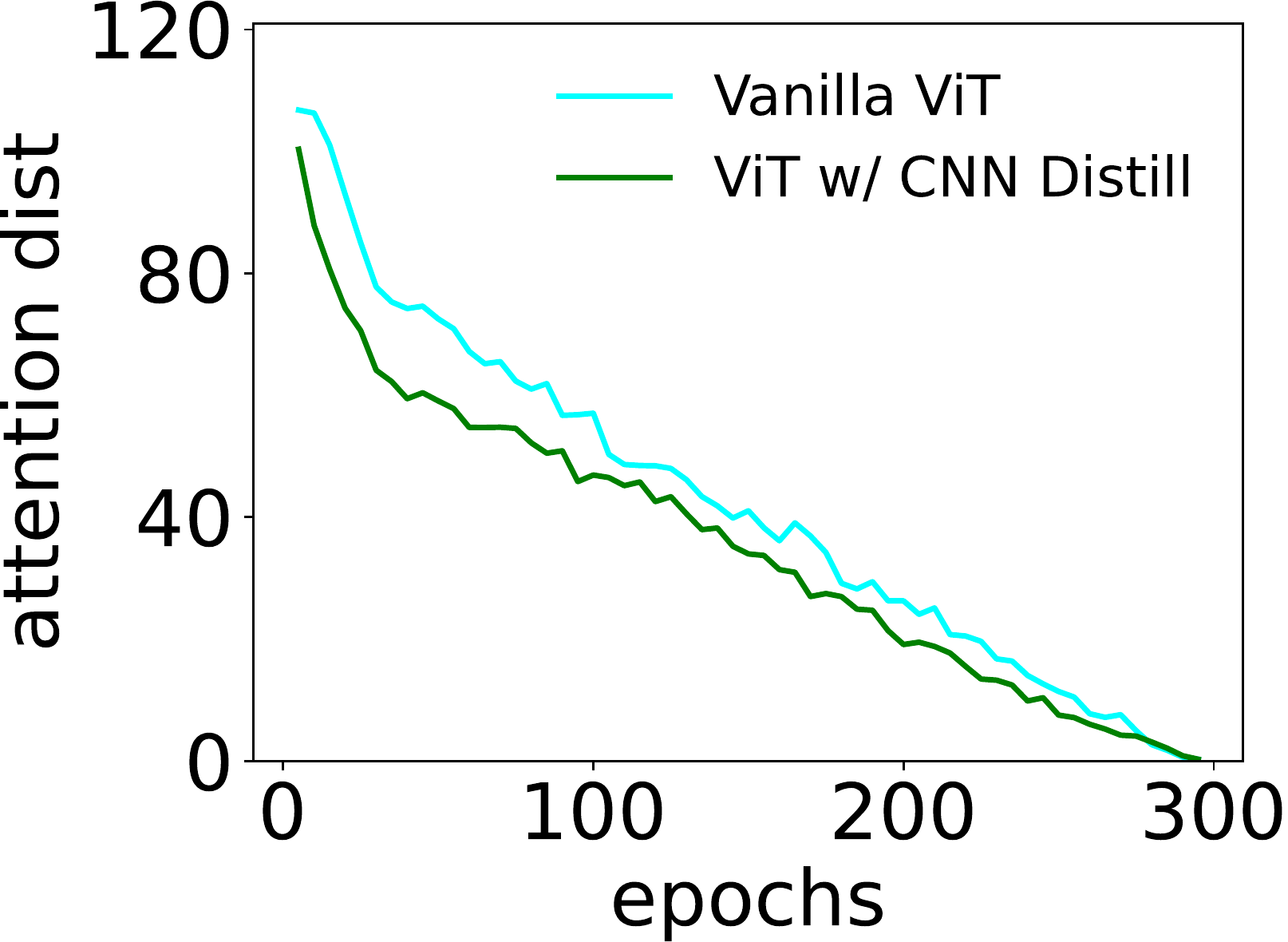}
		\label{fig:pre_attn_dis_layer4}}\hspace{-2mm}
    \subfigure[attn dist (deep)]{\includegraphics[width=.244\textwidth]{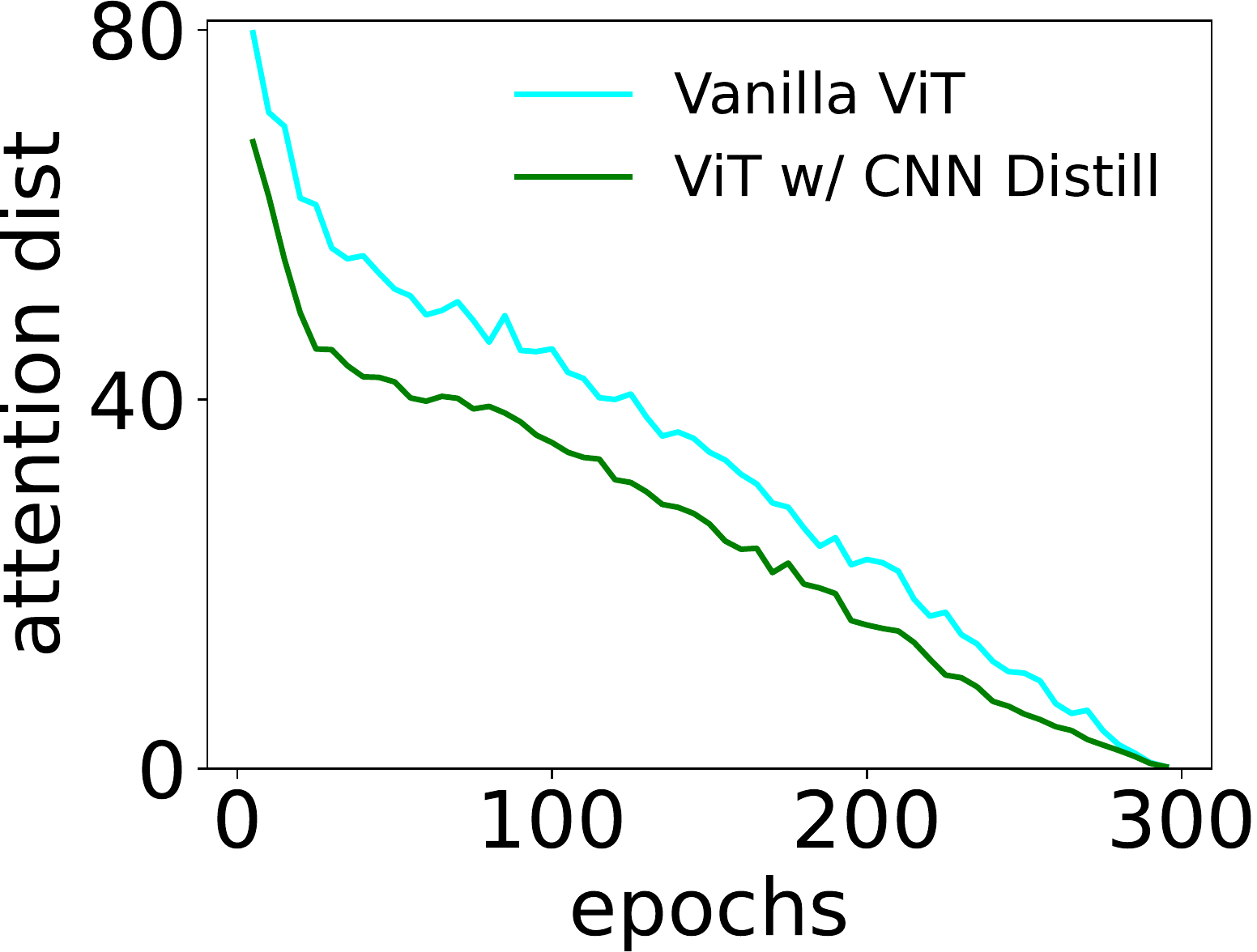}
		\label{fig:pre_attn_dis_layer6}}
	\vspace{-1em}
	\caption{\textbf{(a)} Attention maps from vanilla ViT and CNN-distilled ViT. \textbf{(b)} training accuracy curves between ViT and CNN-distilled ViT. \textbf{(c\&d)} Attention score distance between ViT and CNN-distilled ViT at the middle layer (middle) and last layer (right).}
	\label{fig:pre_attn_dis}
	\vspace{-2em}
\end{figure}

Moreover, we also observe that the convergence speed  is not equivalent to the quality of token dependency learning. In comparison to vanilla ViT, CNN-distilled ViT can learn higher-qualified token dependency as shown in Fig.~\ref{fig:vis_distill_attn}, but has  slower convergence speed and lower training accuracy as illustrated in Fig.~\ref{fig:pre_train_acc_distill}. So one can conclude that the convergence speed actually does not reflect the quality of token dependency learning. Besides, for vanilla ViT,  its  training accuracy increases rapidly in the first 30 epochs but becomes saturated in the following epochs in Fig.~\ref{fig:pre_train_acc_distill}, while its attention maps at the last block still evolve in  Fig.~\ref{fig:per_epoch_baseline}. Indeed,  Fig.~\ref{fig:pre_attn_dis_layer4} (and \ref{fig:pre_attn_dis_layer6}) indicates that  attention score distance descends stably for all training epochs. So these results all show that convergence speed does not well reflect the quality of token dependency learning. 


\begin{figure*}
  \begin{center}
  \includegraphics[width=4.8in]{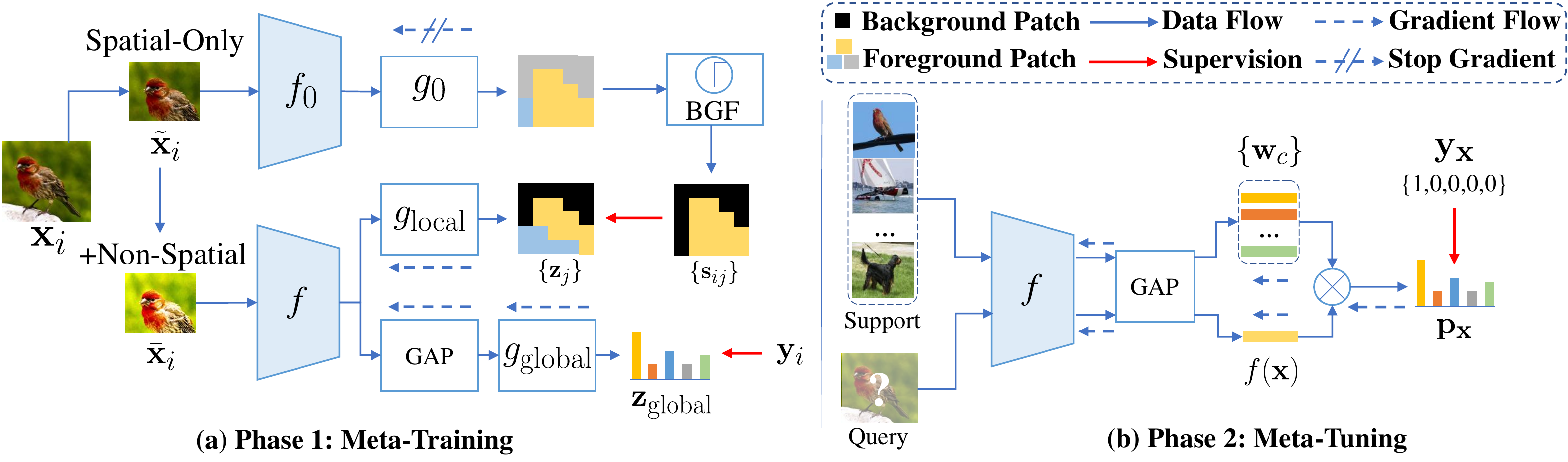}
  \end{center}
  \vspace{-2em}
  \caption{Training pipeline of ViTs with self-promoted supervision (SUN). 
  For meta-training phase (a), given an image $\mathbf{x}_{i}$, we use spatial-consistent augmentation to generate two crops $\bar{\mathbf{x}}_{i}$ and $\tilde{\mathbf{x}}_{i}$.  
  Given  $\tilde{\mathbf{x}}_{i}$, teacher (ViT $f_{0}$ \& and  classifier $g_{0}$) with background filtration (BGF) generates its location-specific supervision $\{\mathbf{s}_{ij}\}$. 
  Given $\bar{\mathbf{x}}_{i}$, the meta-learner $f$ extracts its token features, and then global classifier $g_{\text{global}}$ and local classifier $g_{\text{local}}$ respectively predict the global semantic label $\tilde{\mathbf{y}}_{i}$ of  $\bar{\mathbf{x}}_{i}$ and the semantic labels  $\{\tilde{\mathbf{s}}_{ij}\}$ of all patches.  Finally, SUN uses ground-truth labels $\mathbf{y}_{i}$ and the location-specific label $\{\mathbf{s}_{ij}\}$ to supervise  $\tilde{\mathbf{y}}_{i}$  and $\{\tilde{\mathbf{s}}_{ij}\}$ for  optimizing $f$ and $g_{\text{global/local}}$. 
  During meta-tuning phase (b), SUN adopts an existing few-shot method (\eg, Meta-Baseline~\cite{Chen_2021_ICCV_metabaseline}) to fine-tune $f$ and then predicts the label for each query $\mathbf{x}$ with a support set $\mathbf{S}$.  
  }
  \vspace{-2em}
  \label{fig:pipeline}
  \end{figure*}
\vspace{-1em}
\section{Self-Promoted Supervision for Few-Shot Classification}\label{sec:SUN}
 \vspace{-0.2em}
From the observations and analysis in Sec.~\ref{sec:preliminaries}, one knows that 1) directly replacing the CNN feature extractor with ViT in existing few-shot learning frameworks often leads to severe performance degradation; and 2) the few-shot learning frameworks that boost token dependency learning to remedy lack of inductive bias can improve performance of ViT on few-shot learning tasks.  
Accordingly, here we propose a Self-promoted sUpervisioN (SUN) framework which aims to improve the few-shot learning ability of ViTs via enhancing and accelerating token dependency learning in ViTs via a dense location-specific supervision. Similar to conventional few-shot learning frameworks, \eg~Meta-Baseline, our SUN framework also consists of two phases, \ie, meta-training  and meta-tuning which are illustrated in Fig.~\ref{fig:pipeline}.  In the following, we will introduce these phases. 

\vspace{-1em}
\subsection{Meta-Training} \label{sec:meta-training}
For meta-training phase  in Fig.~\ref{fig:pipeline}(a), similar to Meta-Baseline, our target is to learn a meta learner $f$ mentioned in Sec.~\ref{sec:preliminaries} such that $f$ is able to fast adapt itself to new classes with a few training data. Here $f$ is actually a feature extractor and is a ViT model in this work.  
The main idea of SUN is a dense location-specific supervision which aims to enhance and accelerate token dependency learning in ViT and thus boost the learning efficiency on the limited training data. 

Specifically, to train the meta-learner $f$ on the base dataset $\mathbb{D}_{\text{base}}$, SUN uses global supervision, \ie, the ground-truth label,  to guide the global average token of all patch tokens in $f$ for learning the global semantics of the whole image. More importantly, SUN further employs individual location-specific supervision to supervise each patch token in $f$. To generate patch-level pseudo labels as location-specific supervision, SUN first conducts the supervised classification pretraining to optimize a teacher model $f_g$ consisting of a ViT $f_0$ with the same architecture as $f$ and a classifier $g_0$ on the base dataset $\mathbb{D}_{\text{base}}$. This is reasonable, since teacher $f_0$ is trained on $\mathbb{D}_{\text{base}}$ and its predicted patch-level pseudo labels are semantically consistent with the ground-truth label of the whole image, \ie, similar pseudo labels for similar local and class tokens. This is why our method is so called \emph{self-promoted} since $f_{0}$ has the same architecture as $f$.  
Formally, for each training image $\mathbf{x}_{{i}}$ in the base dataset $\mathbb{D}_{\text{base}}$, SUN first calculates the per-token classification confidence score $\mathbf{\hat{s}}_{i}$ as follows: 
\vspace{-0.3em}
\begin{equation}\label{eq:clsmap}\centering
	\mathbf{\hat{s}}_{i} = [\mathbf{\hat{s}}_{i1}, \mathbf{\hat{s}}_{i2}, \cdots, \mathbf{\hat{s}}_{iK}]=  f_g(\mathbf{x}_{{i}}) = [ g(\mathbf{z}_{\text{1}}), g(\mathbf{z}_{\text{2}}), \cdots,  g(\mathbf{z}_{{K}})] \in \mathbb{R}^{c\times K}, 
	\vspace{-0.2em}
\end{equation} 
where $\mathbf{z}=f(\mathbf{x}_{{i}}) = [\mathbf{z}_{\text{cls}}, \mathbf{z}_{\text{1}}, \mathbf{z}_{\text{2}}, \cdots,  \mathbf{z}_{{K}}]$ denotes the class and patch tokens of image $\mathbf{x}_{{i}}$, and $\mathbf{\hat{s}}_{ij}$ is the pseudo label of the $j$-th local patch $\mathbf{x}_{{ij}}$ in sample $\mathbf{x}_{{i}}$.  Accordingly, for each patch $\mathbf{x}_{{ij}}$, the position with relative high confidence in $\mathbf{\hat{s}}_{ij}$ indicates that this patch contains the semantics of corresponding categories.  

To fully exploit the power of local supervision, we further propose  background filtration (BGF) technique which targets at classifying background patches into a new unique class and improving quality of patch-level supervision. This technique is necessary, since being trained on $\mathbb{D}_{\text{base}}$ which has no background class, the teacher $f_0$ always wrongly assigns background patches into a base (semantic) class instead of background class and provides inaccurate location-specific supervision. To tackle this issue, we filtrate the local patches with very low confidence score as background patches and classify them into a new unique class.  
Specifically, given a batch of $m$ patches with confidence score $\{\mathbf{\hat{s}}_{ij}\}$, we first select the maximum score for each patch, and then sort the $m$ scores of patches by an ascending order. Next, we view the top $p\%$ patches with the lowest scores as background patches. Meanwhile, we add a new unique class, \ie, the background class, into the base classes $C_{\text{base}}$. Accordingly, we increase one dimension for the generated pseudo label $	\mathbf{\hat{s}}_{i} \in\mathbb{R}^{c}$ to obtain $\mathbf{s}_{ij}\in\mathbb{R}^{c+1}$.   For background patches, the last positions of their pseudo label $\mathbf{s}_{ij}\in\mathbb{R}^{c+1}$ are one and remaining positions are zero. For non-background patches, we only set the last positions of their pseudo label as  zero and do not change other values.   In this way,  given one image $\mathbf{x}_{{i}}$ with ground-truth label $\mathbf{y}_{i}$,  we  define its overall training loss:
 \vspace{-0.4em}
 \begin{equation}\label{eq:denseloss}\centering
 	\mathcal{L}_{\scriptsize{\mbox{SUN}}} = H(g_{\scriptsize{\text{global}}}(\mathbf{z}_{\text{global}}), \mathbf{y}_{i}) + \lambda \sum\nolimits_{j=1}^{K} H(g_{\scriptsize{\text{local}}}(\mathbf{z}_{{j}}), \mathbf{s}_{ij}),
 	\vspace{-0.3em}
 \end{equation}
where $\mathbf{z}=f(\mathbf{x}_{{i}}) = [\mathbf{z}_{\text{cls}}, \mathbf{z}_{\text{1}}, \mathbf{z}_{\text{2}}, \cdots,  \mathbf{z}_{{K}}]$ denotes the class and patch tokens of image $\mathbf{x}_{{i}}$, and $\mathbf{z}_{\text{global}}$ is a global average pooling of all  patch tokens $\{\mathbf{z}_{{j}}\}_{j=1}^K$.  Here $H$ denotes the cross entropy loss, and $g_{\scriptsize{\text{global}}}$ and $g_{\scriptsize{\text{local}}}$ are two trainable classifiers for global semantic classification and local patch classification respectively.  For $\lambda$, we set it as  $0.5$ in all experiments for simplicity. 

Now we discuss the two benefits of the dense supervision on few-shot learning tasks. Firstly, all location-specific supervisions are generated by the teacher $f_0$ and thus guarantees similar pseudo labels for similar local tokens. This actually   tells ViT which patch tokens are similar or dissimilar and thus can accelerate  token dependency learning.  Secondly, in contrast to the global supervision on the whole image, the location-specific supervisions are at a fine-grained level, namely the patch level, and thus help ViTs to easily discover the target objects and improve the recognition accuracy. This point is also consistent with some previous literature. For instance, works~\cite{zhong2020random,jiang2021all} demonstrate that modeling semantics in local tokens can avoid learning skewed and non-generalizable patterns and thus substantially improve the model generalization performance. Therefore, both aspects are heavily desired in few-shot learning problems and can increase the few-shot learning capacity  of ViTs.

To further improve the robustness of local supervision from teacher $f_0$ while keeping sufficient data diversity, we propose a ``Spatial-Consistent Augmentation'' (SCA) strategy to improve generalization.   SCA consists of a \emph{spatial-only} augmentation and  a \emph{non-spatial} augmentation. In spatial-only augmentation, we only introduce spatial transformation (\eg, random crop and resize, flip and rotation) to augment the input images $\mathbf{x}_{i}$ and obtain $\mathbf{\tilde{x}}_{i}$. For non-spatial augmentation, it only leverages  non-spatial  augmentations, \eg~color jitter, on  $\mathbf{\tilde{x}}_{i}$ to obtain $\mathbf{\bar{x}}_{i}$. During the meta-training phase, we feed $\mathbf{\tilde{x}}_{i}$ into the teacher $f_0$  to generate $\mathbf{s}_{ij}$ used in Eqn.~\eqref{eq:denseloss}, and feed $\mathbf{\bar{x}}_{i}$ into the target meta-leaner $f$. In this way, the samples used to train meta-learner $f$ are of high diversity, while still enjoying very accurate location-specific supervisions   $\mathbf{s}_{ij}$ since the teacher $f_0$ uses a weak augmentation $\mathbf{\tilde{x}}_{i}$ to generate the location-specific supervisions. This also helps improve the generalization ability of ViTs.

\vspace{-1.0em}
\subsection{Meta-Tuning}  \label{meta-tune}
For meta-tuning phase in Fig.~\ref{fig:pipeline}(b), our target is to finetune the meta-learner $f$ via training it  on multiple ``$N$-way $K$-shot'' tasks $\{\tau\}$ sampled from the base dataset $\mathbb{D}_{\text{base}}$, 
such that $f$ can be adapted to a new task which contains unseen classes $C_{\text{novel}}$ (\ie~$C_{\text{novel}}\notin \mathbb{D}_{\text{base}}$) with only a few labeled training samples. To this end, without loss of generality, we follow  a simple yet effective meta-tuning method in Meta Baseline~\cite{Chen_2021_ICCV_metabaseline}.  Besides, we also investigate different  meta-tuning methods, such as FEAT~\cite{ye2020fewshot} and DeepEMD~\cite{Zhang_2020_CVPR}, in the Appendix~\ref{sec:sun-f} and \ref{sec:sun-d}. With the same FEAT and DeepEMD, our SUN framework still shows superiority over other few-shot learning frameworks. See more details in the Appendix~\ref{sec:sun-f} and \ref{sec:sun-d}.  For completeness,  we introduce  the  meta-tuning  in Meta-Baseline in the following. %
Specifically, given a  task $\tau$ with support set $\mathbf{S}$, SUN calculates the classification  prototype  $\mathbf{w}_{k}$ of class $k$ via $\mathbf{w}_{k}=\sum\nolimits_{\mathbf{x}\in\mathbf{S}_{k}}\mcode{GAP}(f(\mathbf{x}))/\left | \mathbf{S}_{k} \right |$, where $\mathbf{S}_{k}$ denotes the support samples from class $c$ and $\mcode{GAP}$ means global average pooling operation. 
Then for each query image $\mathbf{x}$, meta-learner $f$ calculates the classification confidence score of the $k$-th class:
\begin{equation}\label{eq:p_x_mb}\centering
    \mathbf{p}_{k} = \frac{\exp(\gamma\cdot \mcode{cos}(\mcode{GAP}(f(\mathbf{x})),  \mathbf{w}_{k}))}{\sum\nolimits_{{k}'}\exp(\gamma\cdot    \mcode{cos}(\mcode{GAP}(f(\mathbf{x})),  \mathbf{w}_{k})  )}, \vspace{-0.2em}
\end{equation} 
where  $\mcode{cos}$ denotes cosine similarity and $\gamma$ is a temperature parameter. Finally, it minimizes the cross-entropy loss $\mathcal{L}_{\text{few-shot}} = H(\mathbf{p}_{\mathbf{x}}, \mathbf{y}_{\mathbf{x}})$ to fine-tune meta-leaner $f$ on the various sampled tasks $\{\tau\}$, where $\mathbf{p}_{\mathbf{x}}=[\mathbf{p}_1, \cdots, \mathbf{p}_c]$ is the prediction and $\mathbf{y}_{\mathbf{x}}$ is   ground-truth label of $\mathbf{x}$.  After this meta-tuning, given a new test task $\tau'$ with support set $\mathbf{S}'$, we follow the above step to compute its classification prototypes,  and then use Eqn.~\eqref{eq:p_x_mb} to predict the labels of the test samples. 

\vspace{-1.0em}
\section{Experiments}
\vspace{-0.2em}
We first introduce our  training details in Sec.~\ref{sec:implementation}. Next, we test our SUN on four different ViTs (\ie, LV-ViT, Swin Transformer, Visformer and NesT) in Sec.~\ref{comp:different_vits}. Finally,  we choose NesT~\cite{zhang2021aggregating} as our ViT feature extractor to  compare with other baselines in  Sec.~\ref{sec:comp_samevit} and also to conduct ablation study in Sec.~\ref{sec:ablation}.  Following~\cite{chen2019closerfewshot,Chen_2021_ICCV_metabaseline,Zhou_2021_ICCV}, we evaluate SUN with various ViTs on three widely used few-shot benchmarks, \ie, \emph{mini}ImageNet~\cite{vinyals2016matching}, \emph{tiered}ImageNet~\cite{ren2018metalearning} and CIFAR-FS~\cite{bertinetto2018metalearning} whose details are deferred to the supplement. For fairness,  for each benchmark, we follow~\cite{chen2019closerfewshot,Chen_2021_ICCV_metabaseline,Zhou_2021_ICCV} and report the average accuracy  with 95\% confidence interval on 2,000 tasks under 5-way 1-shot and 5-way 5-shot classification settings.


\vspace{-1em}
\subsection{Training Details} \label{sec:implementation}
For meta-training phase, we use AdamW~\cite{loshchilov2018decoupled} with a learning rate of 5e-4 and a cosine learning rate scheduler~\cite{Loshchilov2017SGDRSG} to train our meta learner $f$ for 800, 800, 300 epochs on \emph{mini}ImageNet, CIFAR-FS and \emph{tiered}ImageNet, respectively.  For augmentation, we use Spatial-Consistent Augmentation in Sec.~\ref{sec:SUN}. 
For location-specific supervision on each patch, we only keep the top-$k$ (\eg, $k=5$) highest confidence in $\mathbf{\hat{s}}_{ij}$  to reduce the label noise.  
To train the teacher $f_g$, we employ the same augmentation strategy in~\cite{pmlr-v139-touvron21a}, including random crop, random augmentation, mixup, etc.  
Meanwhile we adopt AdamW~\cite{loshchilov2018decoupled} with the same parameters as above to train 300 epochs. For meta-tuning, we simply utilize the optimizer and the hyper-parameters used in Meta Baseline, \eg, SGD with learning rate of 1e-3 to finetune the meta-learner $f$ for 40 epochs. 
Moreover, we use relatively large drop path rate 0.5 to avoid overfitting for all training. This greatly differs from conventional setting on drop path rate where it often uses 0.1.  Following conventional supervised setting~\cite{xiao2021early,jiang2021all,yu2021improving} , we also use a three-layer convolution block~\cite{He2016DeepRL} with residual connection to compute  patch embedding. This  conventional stem  has only  $\sim$0.2M  parameters and  is much smaller than ViT backbone. 

\begin{table}[t]
    \centering
    \caption{Comparison between SUN and Meta Baseline on \emph{mini}ImageNet. 
    }
    \label{table:compatibility}
    \vspace{-1em}
    \small
    \setlength{\tabcolsep}{4.0pt} 
   \renewcommand{\arraystretch}{3}
 { \fontsize{8.3}{3}\selectfont{
    \begin{threeparttable}
    \begin{tabular}{lcccccc}
    \toprule
    \multirow{2}{*}{{Backbone}}&\multirow{2}{*}{{Params}}&\multicolumn{2}{c}{\textbf{Meta-Baseline}~\cite{chen2019closerfewshot}}&\multicolumn{2}{c}{\textbf{SUN (Ours)}}
    \\ \cline{3-6}
    &&{{5-way 1-shot}}&{{5-way 5-shot}}&{{5-way 1-shot}}&{ {5-way 5-shot}} \\ 
    \hline
    {LV-ViT}&12.6M&{43.08$\pm$0.38}&{59.03$\pm$0.39}&\textbf{59.00$\pm$0.44}&\textbf{75.47$\pm$0.34} \\
    {Swin}&12.6M&{54.63$\pm$0.45}&{70.60$\pm$0.38}&\textbf{64.94$\pm$0.46}&\textbf{80.40$\pm$0.32} \\
    {Visformer}&12.5M&{47.61$\pm$0.43}&{63.00$\pm$0.39}&\textbf{67.80$\pm$0.45}&\textbf{83.25$\pm$0.29}\\
    {NesT}&12.8M&{54.57$\pm$0.46}&{69.85$\pm$0.38}&\textbf{66.54$\pm$0.45}&\textbf{82.09$\pm$0.30} \\ 
    \toprule
    \end{tabular}
    \end{threeparttable}}}
    \vspace{-1.3em}
\end{table}
\begin{figure}
  \centering
  \includegraphics[width=\textwidth]{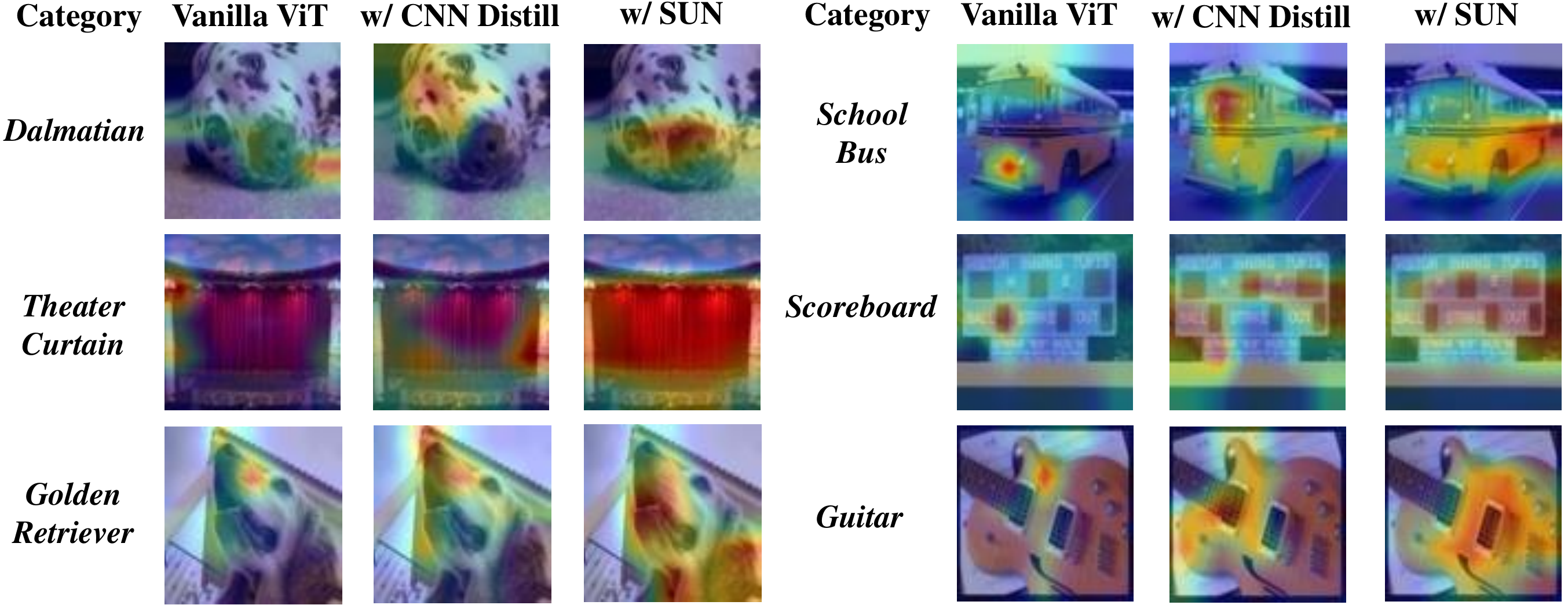}
  \vspace{-2.5em}
  \caption{Visualization of attention maps from vanilla ViT, CNN-distilled ViT and ViT with SUN. ViT with SUN performs better, thus obtains better token dependency. }
  \label{fig:attn_comp_SUN}
  \vspace{-2em}
\end{figure}

\subsection{Comparison on Different ViTs}\label{comp:different_vits}
\vspace{-0.2em}
We evaluate our SUN on four different ViTs, \ie, LV-ViT, Swin, Visformer and NesT, which cover most of existing ViT types. Table~\ref{table:compatibility} shows that on \emph{mini}ImageNet,  SUN significantly surpasses  Meta-Baseline~\cite{Chen_2021_ICCV_metabaseline} on the four ViTs. Specifically, it makes 15.9\%, 10.3\%, 20.2\%, and 12.0\% improvement  over Meta-Baseline  under 5-way 1-shot  setting. 
These results demonstrate the superior generalization ability of our SUN on novel categories, and also its good compatibility with different kinds of ViTs.  
Besides, among all ViTs, Visformer achieves the best performance, since it uses many CNN modules and can better handle the few-shot learning task.  
In the following Sec.~\ref{sec:comp_samevit} and \ref{sec:ablation}, we use the 
second best NesT as our ViT feature extractor, since NesT does not involve CNN modules and only introduce modules to focus on local features and thus better reveals few-shot learning ability of conventional ViT architecture. And in Sec.~\ref{sec:comp_SoTA}, we use both NesT and Visformer to make comparisons with the state-of-the-art CNN-based few-shot classification methods.

\noindent\textbf{SUN v.s. ViT in dependency learning. } 
Following the analysis in Sec.~\ref{sec:analysis}, we visualize the attention maps of vanilla ViT, CNN-distilled ViT and ViT with SUN to compare the quality of token dependency learning. Fig.~\ref{fig:attn_comp_SUN} demonstrate the comparison result. ViT with SUN can capture more semantic tokens than vanilla ViT and CNN-distilled ViT on various categories. The result indicates that ViT with SUN can learn better token dependency during the training. 

\begin{table*}[t]
    \centering
    \caption{Comparison of different  ViT based few-shot classification frameworks  on \emph{mini}ImageNet. ``AC'' means introducing  CNN modules in ViT backbone to fuse features.  ``Distill'' is knowledge distillation.  ``FT'' is using  meta-tuning  to finetune meta learner (see Sec.~\ref{meta-tune}). Except \cite{li2021vision}, all methods use NesT as ViT  feature extractor. }
    \label{table:compfstrans}
    \vspace{-1em}
    \small
    \setlength{\tabcolsep}{3.8pt} 
   \renewcommand{\arraystretch}{3}
 { \fontsize{8.3}{3}\selectfont{
    \begin{threeparttable}
    \begin{tabular}{lcccccc}
    \toprule
    \multirow{2}{*}{\textbf{Method}}&\multirow{2}{*}{\textbf{Backbone}}&\multirow{2}{*}{\textbf{AC}}&\multirow{2}{*}{\textbf{Distill}}&\multirow{2}{*}{\textbf{FT}}&\multicolumn{2}{c}{\textbf{\emph{mini}ImageNet}}
    \\ \cline{6-7}
    &&&&&{{5-way 1-shot}}&{{5-way 5-shot}}
    \\ \hline
    {Meta-Baseline}~\cite{Chen_2021_ICCV_metabaseline}&\emph{ResNet-12}& & &\checkmark&{64.53$\pm$0.45}&{81.41$\pm$0.31} \\
    \hline
    {Baseline}~\cite{chen2019closerfewshot}&\emph{ViT}& & & &{49.23$\pm$0.43}&{66.57$\pm$0.39} \\
    {Meta-Baseline}~\cite{Chen_2021_ICCV_metabaseline}&\emph{ViT}& & &\checkmark&{54.57$\pm$0.46}&{69.85$\pm$0.38} \\
    {Re-Parameter}~\cite{li2021vision}&\emph{ViT}&\checkmark& &\checkmark&{46.59$\pm$0.44}&{62.56$\pm$0.41} \\
    \cite{Chen_2021_ICCV_metabaseline}+{CNN-Distill}~\cite{weng2021semisupervised}&\emph{ViT}&\checkmark&\checkmark&\checkmark&{55.79$\pm$0.45}&{71.81$\pm$0.37} \\
    {Semiformer}~\cite{weng2021semisupervised}&\emph{CNN+ViT}&\checkmark&\checkmark& &{56.62$\pm$0.46}&{72.91$\pm$0.39} \\
    \cite{Chen_2021_ICCV_metabaseline}+{Semiformer}~\cite{weng2021semisupervised}&\emph{CNN+ViT}&\checkmark&\checkmark&\checkmark&{57.91$\pm$0.48}&{73.31$\pm$0.38} \\
    \cite{Chen_2021_ICCV_metabaseline}{+DrLoc}~\cite{liu2021efficient}&\emph{ViT}& & &\checkmark&{57.85$\pm$0.46}&{74.03$\pm$0.35} \\
    {BML}~\cite{Zhou_2021_ICCV}&\emph{ViT}& &\checkmark& &{59.35$\pm$0.45}&{76.00$\pm$0.35} \\
    \hline
    {SUN (Ours)}&\emph{ViT}& &\checkmark&\checkmark&\textbf{66.54$\pm$0.45}&\textbf{82.09$\pm$0.30} \\
    \toprule
    \end{tabular}
    \end{threeparttable}}}
    \vspace{-1.5em}
\end{table*}

\vspace{-1.2em}
\subsection{Comparison among Different Few-shot Learning Frameworks}\label{sec:comp_samevit}
In Table~\ref{table:compfstrans},  we compare SUN with other few-shot learning frameworks on the same ViT architecture.  In Table~\ref{table:compfstrans},  the method``\cite{Chen_2021_ICCV_metabaseline}+CNN-Distill" pretrains a CNN to distill a ViT;  ``Semiformer"~\cite{weng2021semisupervised} introduces  CNN modules into ViT; ``Re-Parameter"~\cite{li2021vision}  initializes a ViT by  a trained CNN ; ``\cite{Chen_2021_ICCV_metabaseline}+DrLoc" predicts the relative positions between local token pairs; ``BML"~\cite{Zhou_2021_ICCV} introduces mutual learning based few-shot learning framework. 
All these methods propose their own techniques to improve few-shot learning performance of ViT via introducing CNN-alike inductive bias or constructing tasks to better learn token dependency.  
Except~\cite{li2021vision} that uses its CNN-transformer-mixed architecture, all others  adopt    NesT~\cite{zhang2021aggregating} as a ViT feature extractor or use ResNet-12 as the CNN feature extractor.   Table~\ref{table:compfstrans} shows that our SUN significantly surpasses all these methods in terms of accuracy on novel categories. 
Specifically, our SUN outperforms the runner-up by 7.19\% and 6.09\% under 5-way  1-shot  and 5-shot settings, respectively.    
These  results show that even on the same ViT, SUN achieves much better few-shot accuracy on novel categories than other few-shot learning frameworks. 

\begin{table}[t]
  \centering
  \caption{Comparison with SoTA  few-shot learning methods under  5-way few-shot classification setting. The results of the best 2 methods are in bold font.}
  	\vspace{-1em}
  \label{table:SoTAfs}
      \setlength{\tabcolsep}{0.8pt} 
 \renewcommand{\arraystretch}{2.4}
{ \fontsize{6.5}{3}\selectfont{
  \begin{threeparttable}
  \begin{tabular}{lccccccc}
  \toprule
  \multirow{2}{*}{\textbf{Method}}&\multirow{2}{*}{\shortstack{\textbf{Classifier}\\\textbf{Params}}}&\multicolumn{2}{c}{\textbf{\emph{mini}ImageNet}}&\multicolumn{2}{c}{\textbf{\emph{tiered}ImageNet}}&\multicolumn{2}{c}{\textbf{CIFAR-FS}}
  \\ \cline{3-8}
  &&{{1-shot}}&{{5-shot}}&{{1-shot}}&{{5-shot}}&{{1-shot}}&{{5-shot}}
  \\ \toprule
 \multicolumn{7}{l}{ \textbf{ResNet-12/18 as feature extractor}  }\\
  
  {MetaOptNet}~\cite{Lee2019MetaLearningWD}&0&64.09$\pm$0.62&80.00$\pm$0.45&65.81$\pm$0.74&81.75$\pm$0.53&72.00$\pm$0.70&84.20$\pm$0.50
  \\ 
  {DeepEMD}~\cite{Zhang_2020_CVPR}&0&65.91$\pm$0.82&82.41$\pm$0.56&71.16$\pm$0.80&86.03$\pm$0.58&46.47$\pm$0.70&63.22$\pm$0.71
  \\ 
  {FEAT}~\cite{ye2020fewshot}&1.05M&66.78$\pm$0.20&82.05$\pm$0.14&70.80$\pm$0.23&84.79$\pm$0.16&-&-
  \\ 
  {TADAM}~\cite{oreshkin2018tadam}&1.23M&58.50$\pm$0.30&76.70$\pm$0.30&-&-&-&-
  \\ 
  {Rethink-Distill}~\cite{tian2020rethink}&225K&64.82$\pm$0.60&82.14$\pm$0.43&71.52$\pm$0.69&86.03$\pm$0.49&73.90$\pm$0.80&86.90$\pm$0.50
  \\
  {DC}~\cite{Lifchitz_2019_CVPR}&224K&61.26$\pm$0.20&79.01$\pm$0.13&-&-&-&-
  \\
  {CloserLook++}~\cite{chen2019closerfewshot}&131K&51.87$\pm$0.77&75.68$\pm$0.63&-&-&-&-
  \\
  {Meta-Baseline}~\cite{Chen_2021_ICCV_metabaseline}&0&63.17$\pm$0.23&79.26$\pm$0.17&68.62$\pm$0.27&83.29$\pm$0.18&-&-
  \\
  {Neg-Cosine}~\cite{liu2020negative}&131K&63.85$\pm$0.81&81.57$\pm$0.56&-&-&-&-
  \\
  {AFHN}~\cite{li2020adversarial}&359K&62.38$\pm$0.72&78.16$\pm$0.56&-&-&68.32$\pm$0.93&81.45$\pm$0.87
  \\
  {Centroid}~\cite{afrasiyabi2020associative}&10K&59.88$\pm$0.67&80.35$\pm$0.73&69.29$\pm$0.56&85.97$\pm$0.49&-&- \\
  {RE-Net}~\cite{Kang_2021_ICCV}&430K&{67.60$\pm$0.44}&{82.58$\pm$0.30}&{71.61$\pm$0.51}&{85.28$\pm$0.35}&{74.51$\pm$0.46}&{86.60$\pm$0.32} \\
  {TPMN}~\cite{Wu_2021_ICCV}&16M&\textbf{67.64$\pm$0.63}&\textbf{83.44$\pm$0.43}&{72.24$\pm$0.70}&{86.55$\pm$0.63}&{75.50$\pm$0.90}&{87.20$\pm$0.60} \\
  \toprule
   \multicolumn{7}{l}{ \textbf{NesT ViT as feature extractor}  }\\
  {CloserLook++}~\cite{chen2019closerfewshot}&180K&{49.23$\pm$0.43}&{66.57$\pm$0.39}&{59.13$\pm$0.46}&{77.88$\pm$0.39}&{63.89$\pm$0.49}&{80.43$\pm$0.37}
  \\
  {Meta-Baseline}~\cite{Chen_2021_ICCV_metabaseline}&0&{54.57$\pm$0.46}&{69.85$\pm$0.38}&{63.73$\pm$0.47}&{79.33$\pm$0.38}&{68.05$\pm$0.48}&{81.53$\pm$0.36}
  \\
  {BML}~\cite{Zhou_2021_ICCV}&180K&{59.35$\pm$0.45}&{76.00$\pm$0.35}&{66.98$\pm$0.50}&{83.75$\pm$0.34}&{67.51$\pm$0.48}&{82.17$\pm$0.36} \\
  {SUN}&0&{66.54$\pm$0.45}&{82.09$\pm$0.30}&\textbf{72.93$\pm$0.50}&\textbf{86.70$\pm$0.33}&\textbf{78.17$\pm$0.46}&\textbf{88.98$\pm$0.33} \\
  \toprule
\multicolumn{7}{l}{ \textbf{Visformer ViT as feature extractor}  }\\
  {SUN}&0&\textbf{67.80$\pm$0.45}&\textbf{83.25$\pm$0.30}&\textbf{72.99$\pm$0.50}&\textbf{86.74$\pm$0.33}&\textbf{78.37$\pm$0.46}&\textbf{88.84$\pm$0.32} \\
  \toprule
  \end{tabular}
  \end{threeparttable}}}
  \vspace{-2.5em}
\end{table}

\vspace{-1.2em}
\subsection{Comparison with State-of-The-Arts}\label{sec:comp_SoTA}
\vspace{-0.3em}
Here  we compare SUN with state-of-the-arts (SoTAs), including CNN based methods and ViT based one,   on \emph{mini}ImageNet~\cite{vinyals2016matching}, \emph{tiered}ImageNet~\cite{ren2018metalearning} and  CIFAR-FS~\cite{bertinetto2018metalearning}.  
Table~\ref{table:SoTAfs} reports the evaluation results.  
Without introducing extremely complex few-shot learning methods like~\cite{Wu_2021_ICCV,Kang_2021_ICCV}, our SUN  achieves comparable performance with SoTA on \emph{mini}ImageNet~\cite{vinyals2016matching}, and sets new SoTAs on \emph{tiered}ImageNet~\cite{ren2018metalearning} and  CIFAR-FS~\cite{bertinetto2018metalearning}. 
Specifically, on \emph{tiered}ImageNet under 5-way 1-shot and 5-shot settings, our  SUN (Visformer) respectively obtains 72.99\% and 86.74\%,  and respectively improves $\sim$0.8\% and $\sim$0.2\% over the SoTA TPMN~\cite{Wu_2021_ICCV}.     
On CIFAR-FS dataset, our SUN (NesT) obtains 78.17\% and 88.98\% in terms of 1-shot accuracy and 5-shot accuracy, which significantly outperforms all the state-of-the-art methods by at least 2.5\% in terms of 1-shot accuracy.  
Meanwhile, our SUN (Visformer) also obtains 67.80\% 1-shot accuracy on \emph{mini}ImageNet \emph{test} set and surpasses the SoTA CNN-based few-shot classification methods by $\sim$1.2\%. 
These results well demonstrate the effectiveness of ViTs with our SUN in few-shot learning. 

\vspace{-1.2em}
\subsection{Ablation Study}\label{sec:ablation}
\noindent{\bf{Effect of each Phase in SUN. }} 
We investigate the effect of each training phase in SUN. For fairness, we always use the same ViT feature extractor~\cite{zhang2021aggregating} as meta-learner $f$ for few-shot classification. 
Table~\ref{table:ablation_phase} shows that on \emph{mini}ImageNet dataset,  compared to the baseline (denoted by ``Base'' in Table~\ref{table:ablation_phase}), 
the teacher ViT $f_g$ of meta-training phase significantly outperforms the baseline by 13.2\% and 12.9\% in terms of 1-shot and 5-shot accuracy. 
By introducing the location-specific supervision, type (b) obtains $\sim$0.7\% improvement for 1-shot accuracy. Then after introducing SCA and BGF, type (d) can further lead to a gain of $\sim$1.8\% 1-shot accuracy as well as $\sim$1.0\% 5-shot accuracy. A possible reason is that the SCA improves the quality of location-specific supervision from the pretrained transformer while keeping the strong augmentation ability for the target transformer in the meta-training phase. Meanwhile, BGF eliminates the negative effect from mislabeling background patches and improves the feature representation quality. 
These two techniques significantly improve location-specific supervision  and   improve  the generalization ability of ViT feature extractors.

\begin{table}[t]
    \centering
    \caption{Ablation study of  SUN on \emph{mini}ImageNet. ``Local'' means location-specific supervision.   {SCA} is spatial-consistent augmentation.  {BGF} means background filtration. And $f_{g}$ means the teacher ViT model in SUN meta-training phase.}
    \label{table:ablation_phase}
     \vspace{-1em}
    \setlength{\tabcolsep}{3.8pt} 
   \renewcommand{\arraystretch}{3}
 { \fontsize{8.3}{3}\selectfont{
    \begin{threeparttable}
    \begin{tabular}{l|cccc|c|cc}
    \toprule
    \multirow{2}{*}{\textbf{Type}}&\multicolumn{4}{c}{\textbf{Meta-Training}}&\multicolumn{1}{c}{\textbf{Meta-Tuning}}&\multicolumn{2}{c}{\textbf{\emph{mini}ImageNet}}
    \\ \cline{2-8}
    &$f_{g}$&Local&SCA&BGF&Meta-Baseline~\cite{Chen_2021_ICCV_metabaseline}&{{5-way 1-shot}}&{{5-way 5-shot}}
    \\ \hline
    {Base}&&&&&&{49.23$\pm$0.43}&{66.57$\pm$0.39} \\
    \hline
    {(a)}&\checkmark&&&&&{62.40$\pm$0.44}&{79.45$\pm$0.32} \\
    {(b)}&\checkmark&\checkmark&&&&{63.06$\pm$0.45}&{79.91$\pm$0.32} \\
    {(c)}&\checkmark&\checkmark&\checkmark&&&{64.50$\pm$0.45}&{80.59$\pm$0.31} \\
    {(d)}&\checkmark&\checkmark&\checkmark&\checkmark&&{64.84$\pm$0.45}&{80.96$\pm$0.32} \\
    {(e)}&\checkmark&\checkmark&\checkmark&\checkmark&\checkmark&\textbf{66.54$\pm$0.45}&\textbf{82.09$\pm$0.30} \\
    \toprule
    \end{tabular}
    \end{threeparttable}}}
     \vspace{-1.4em}
  \end{table}
\begin{table}[t]
    \caption{Effect of drop path rate in teacher ViT model  on \emph{mini}ImageNet. }
    \vspace{-2em}
    \begin{center}
    \label{tab:ablation_dpr}
        \setlength{\tabcolsep}{3.6pt} 
   \renewcommand{\arraystretch}{3}
 { \fontsize{8.3}{3}\selectfont{
    \begin{tabular}{c|c|c|c|c|c}
    \toprule
   drop path rate $p_{dpr}$ & 0.1 & 0.2 & 0.3 & \textbf{0.5} & 0.8 \\
    \hline
    {5-way 1-shot} & 60.10$\pm$0.45 & 60.47$\pm$0.45 & 62.02$\pm$0.45 & \textbf{62.40$\pm$0.44} & 61.73$\pm$0.44 \\
    {5-way 5-shot} & 77.44$\pm$0.33 & 77.97$\pm$0.34 & 79.29$\pm$0.31 & \textbf{79.45$\pm$0.32} & 78.12$\pm$0.32 \\
    \toprule
    \end{tabular}}}
    \end{center}
    \vspace{-3em}
    \end{table}
\begin{table}[h]
\caption{Ablation of meta-tuning, where ``SUN-*'' means ours and others are SoTAs. 
}
 \vspace{-1.0em}
\centering
\resizebox{\columnwidth}{!}{
\begin{tabular}{c | c c  c  c | c  c  c }
\hline
Methods & MetaBaseline~\cite{Chen_2021_ICCV_metabaseline} & FEAT~\cite{ye2020fewshot} & DeepEMD~\cite{Zhang_2020_CVPR} & COSOC~\cite{luo2021rectifying} & SUN-M & SUN-F & SUN-D \\
\hline
5-way 1-shot & 64.53$\pm$0.45 & 66.78$\pm$0.20 & 68.77$\pm$0.29 & {69.28$\pm$0.49} & 67.80$\pm$0.45 & 66.90$\pm$0.44 & \textbf{69.56$\pm$0.44}   \\
5-way 5-shot & 81.41$\pm$0.31 & 82.05$\pm$0.14 
& 84.13$\pm$0.53 &  {85.16$\pm$0.42} & 83.25$\pm$0.30 & 82.63$\pm$0.30 & \textbf{85.38$\pm$0.49} \\
\hline
\end{tabular}
 }
\label{table:ablation_mata_tuning}
 \vspace{-1.5em}
\end{table}
\begin{figure*}[!htb]
  \label{fig:ablation_curve}
  \centering
  \subfigure[base \emph{train} acc]{
          \includegraphics[width=1.1in]{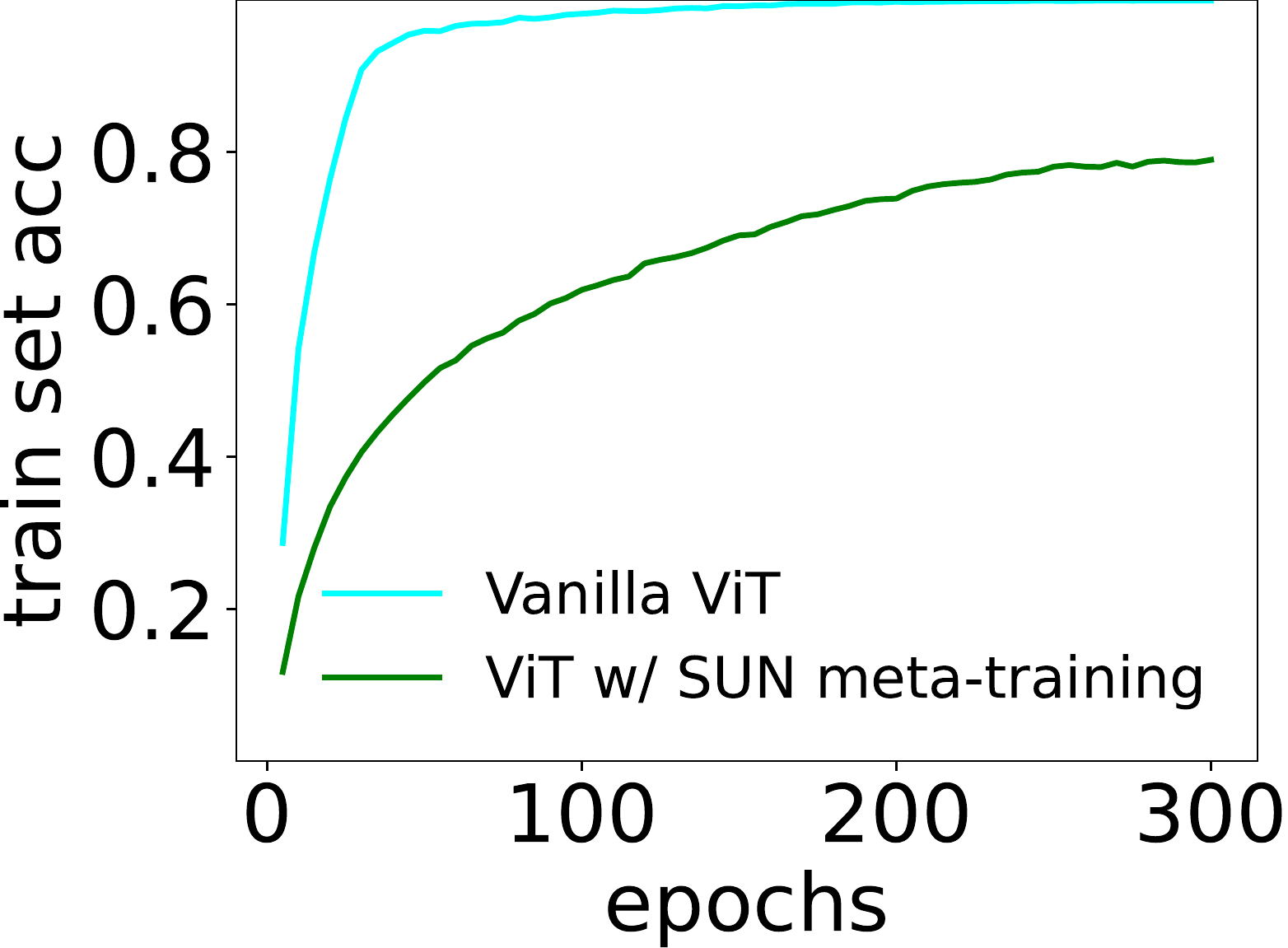}
          \label{fig:ablation_train}
  } \hspace{-2mm} \subfigure[base \emph{val} acc]{
      \includegraphics[width=1.1in]{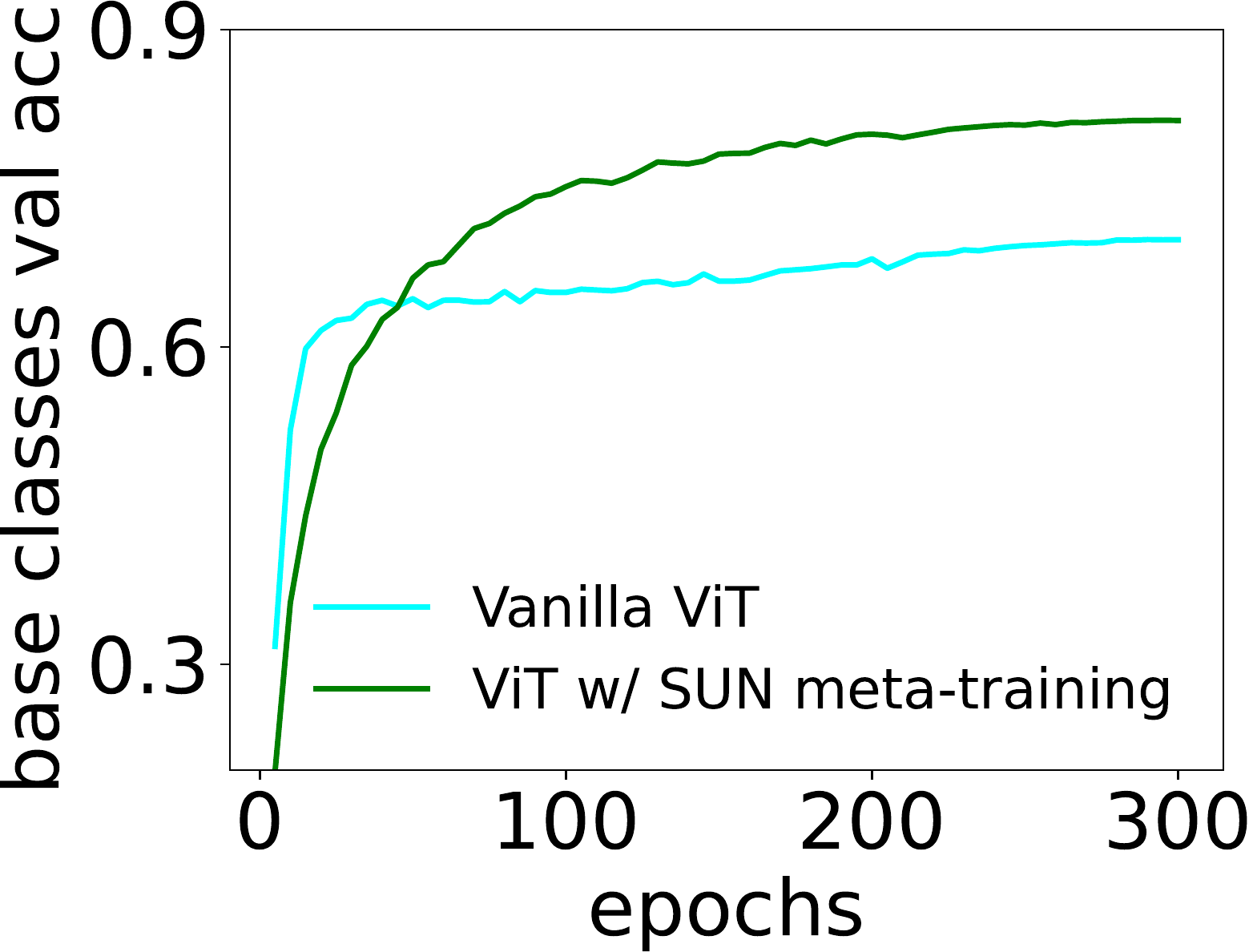}
      \label{fig:ablation_val}
  } \hspace{-2mm} \subfigure[\emph{train} 1-shot acc]{
      \includegraphics[width=1.1in]{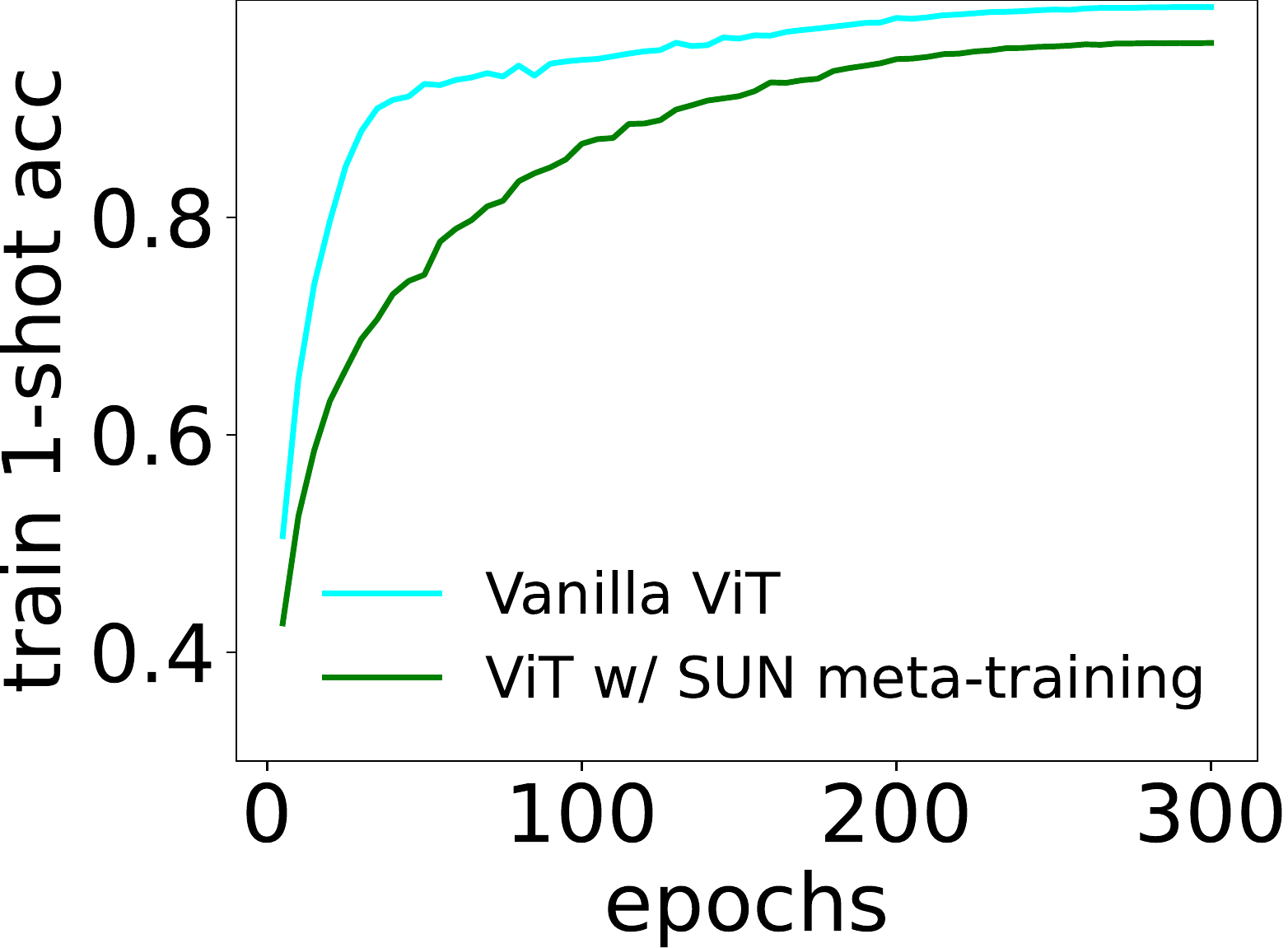}
          \label{fig:ablation_1shot}
  } \hspace{-2mm} \subfigure[\emph{test} 1-shot acc]{
      \includegraphics[width=1.1in]{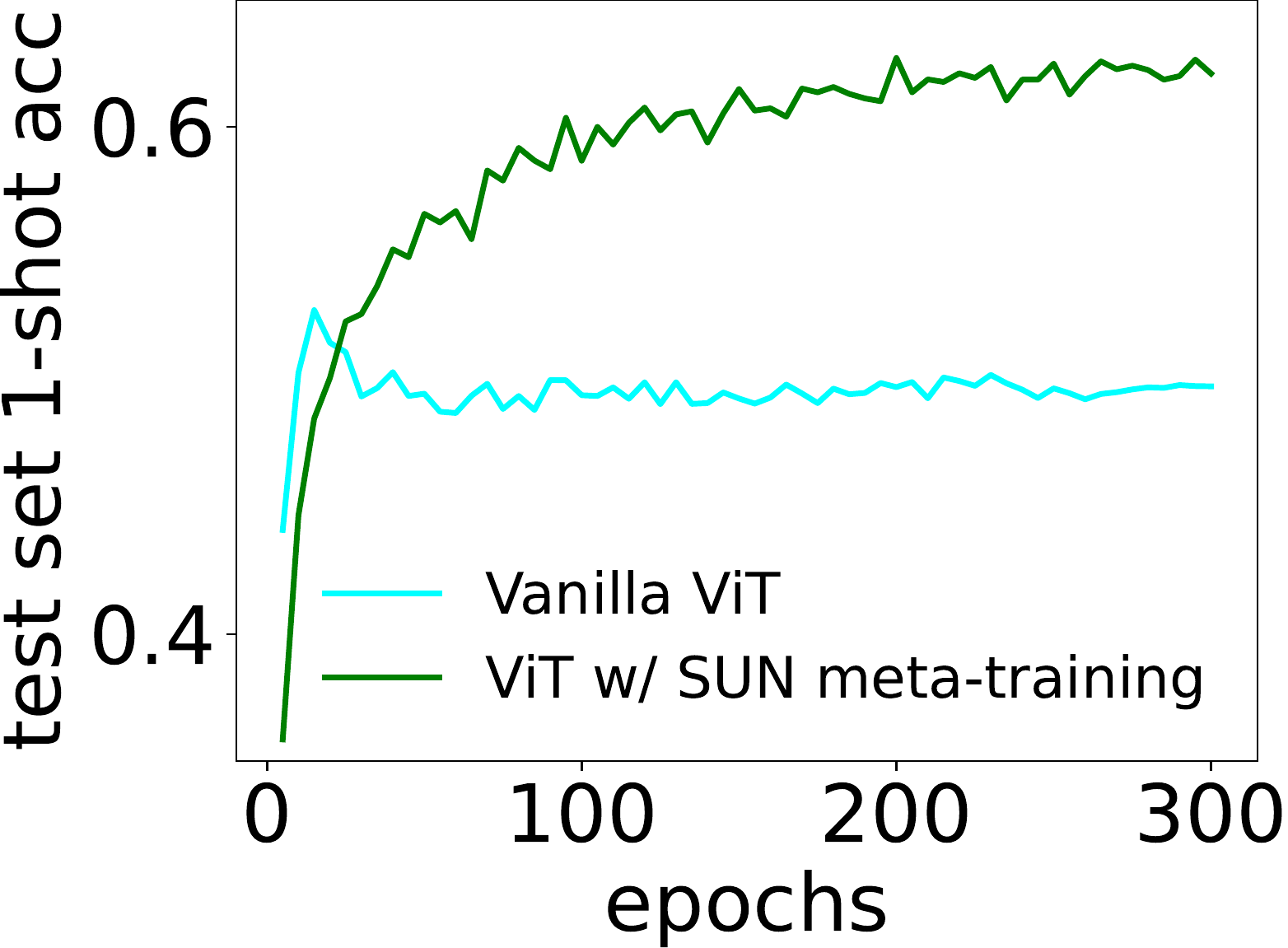}
          \label{fig:ablation_5shot}
  } \\
   \vspace{-1.5em}
  \caption{Accuracy of ViTs (w/ or w/o SUN meta-training phase) on \emph{mini}ImageNet. ViTs with SUN meta-training generalize better on both base and novel categories.}
  \vspace{-2em}
\end{figure*}

To further analyze our meta-training phase,  we plot the accuracy curves of the  meta-training method  in~\cite{chen2019closerfewshot} and our  meta-training phase  in Fig.~\ref{fig:ablation_train}$\sim$\ref{fig:ablation_5shot}. 
As shown in Fig.~\ref{fig:ablation_1shot} and~\ref{fig:ablation_5shot}, ViT with SUN achieves higher classification accuracy on novel classes.  Besides, as shown in Fig.~\ref{fig:ablation_val},  SUN also obtains $\sim$11\%  improvement on base classes than meta-training method~\cite{chen2019closerfewshot}. 
This observation inspires us to rethink the training paradigm of ViTs for few-shot classification, 
such that ViTs can generalize well on both base and novel categories. 

We also investigate  meta-tuning phase. As shown in type (e) in Table~\ref{table:ablation_phase}, the meta-learner $f$ after meta-training phase can be naturally adopted to the existing few-shot learning methods and obtain further improvement. 
Specifically, our SUN (denoted by (e)) also surpasses $f$ after meta-training phase by 1.7\% and 1.1\% in terms of 1-shot and 5-shot accuracy. 

\noindent{\bf{Meta-Tuning Methods. }} We also evaluate our SUN with different meta-tuning methods. To empirically analyze our SUN, we fix Visformer~\cite{Chen_2021_ICCV} as ViT backbone and choose Meta-Baseline~\cite{Chen_2021_ICCV_metabaseline}, FEAT~\cite{ye2020fewshot} and a dense prediction method DeepEMD~\cite{Zhang_2020_CVPR}. As shown in Table~\ref{table:ablation_mata_tuning}, all SUN-M/F/D improve the accuracy and perform better than corresponding CNN-based baselines. Moreover, SUN-D achieves 69.56\% 5-way 1-shot accuracy, even slightly surpasses the SoTA COSOC~\cite{luo2021rectifying}.

\noindent{\bf{Drop Path Rate Analysis. }}
As mentioned in Sec.~\ref{sec:implementation}, we also show the effect of drop path rate $p_{dpr}$. Here we use the teacher ViT model $f_{0}$ in meta-training phase to investigate.   Table~\ref{tab:ablation_dpr} shows  that  a relatively large drop path rate (\eg, $p_{dpr}=0.5$) gives highest accuracy, since it can well mitigate over-fitting and is suitable for few-shot learning problems where training samples are limited.


\vspace{-1em}
\section{Conclusion}
\vspace{-0.5em}
In this work, we first empirically showed  that  replacing the CNN  feature extractor as  a ViT model leads to severe performance degradation on few-shot learning tasks, and  the slow token  dependency learning on limited training data largely contributes to this performance degradation.  Then  for the first time, we proposed a  simple yet effective few-shot training framework for ViTs, \ie~Self-promoted sUpervisioN (SUN), to resolve this issue.  By firstly pretraining the ViT on the few-shot learning dataset, SUN adopts it to generate individual location-specific supervision for 1) guiding the ViT on which patch tokens are similar or dissimilar and thus accelerating token dependency learning; 2) improving the object grounding and recognition ability. Experimental results on few-shot classification tasks demonstrate the superiority of SUN in accuracy over state-of-the-arts.

%

%

\clearpage
%
%
\bibliographystyle{splncs04}
\bibliography{egbib}

\begin{thebibliography}{10}
\providecommand{\url}[1]{\texttt{#1}}
\providecommand{\urlprefix}{URL }
\providecommand{\doi}[1]{https://doi.org/#1}

\bibitem{afrasiyabi2020associative}
Afrasiyabi, A., Lalonde, J.F., Gagn{\'e}, C.: Associative alignment for
  few-shot image classification. In: Vedaldi, A., Bischof, H., Brox, T., Frahm,
  J.M. (eds.) Computer Vision -- ECCV 2020. pp. 18--35. Springer International
  Publishing, Cham (2020)

\bibitem{bao2022beit}
Bao, H., Dong, L., Piao, S., Wei, F.: {BE}it: {BERT} pre-training of image
  transformers. In: International Conference on Learning Representations
  (2022), \url{https://openreview.net/forum?id=p-BhZSz59o4}

\bibitem{bertinetto2018metalearning}
Bertinetto, L., Henriques, J.F., Torr, P., Vedaldi, A.: Meta-learning with
  differentiable closed-form solvers. In: International Conference on Learning
  Representations (2019), \url{https://openreview.net/forum?id=HyxnZh0ct7}

\bibitem{cao2022training}
Cao, Y.H., Yu, H., Wu, J.: Training vision transformers with only 2040 images
  (2022)

\bibitem{detr20}
Carion, N., Massa, F., Synnaeve, G., Usunier, N., Kirillov, A., Zagoruyko, S.:
  End-to-end object detection with transformers. In: Vedaldi, A., Bischof, H.,
  Brox, T., Frahm, J. (eds.) Computer Vision - {ECCV} 2020 - 16th European
  Conference, Glasgow, UK, August 23-28, 2020, Proceedings, Part {I}. Lecture
  Notes in Computer Science, vol. 12346, pp. 213--229. Springer (2020).
  \doi{10.1007/978-3-030-58452-8\_13},
  \url{https://doi.org/10.1007/978-3-030-58452-8\_13}

\bibitem{caron2021emerging}
Caron, M., Touvron, H., Misra, I., J\'egou, H., Mairal, J., Bojanowski, P.,
  Joulin, A.: Emerging properties in self-supervised vision transformers. In:
  Proceedings of the International Conference on Computer Vision (ICCV) (2021)

\bibitem{chen2019closerfewshot}
Chen, W.Y., Liu, Y.C., Kira, Z., Wang, Y.C., Huang, J.B.: A closer look at
  few-shot classification. In: International Conference on Learning
  Representations (2019)

\bibitem{Chen_2021_ICCV_metabaseline}
Chen, Y., Liu, Z., Xu, H., Darrell, T., Wang, X.: Meta-baseline: Exploring
  simple meta-learning for few-shot learning. In: Proceedings of the IEEE/CVF
  International Conference on Computer Vision (ICCV). pp. 9062--9071 (October
  2021)

\bibitem{Chen_2021_ICCV}
Chen, Z., Xie, L., Niu, J., Liu, X., Wei, L., Tian, Q.: Visformer: The
  vision-friendly transformer. In: Proceedings of the IEEE/CVF International
  Conference on Computer Vision (ICCV). pp. 589--598 (October 2021)

\bibitem{deng2009imagenet}
Deng, J., Dong, W., Socher, R., Li, L.J., Li, K., Fei-Fei, L.: Imagenet: A
  large-scale hierarchical image database. In: 2009 IEEE conference on computer
  vision and pattern recognition. pp. 248--255. Ieee (2009)

\bibitem{Doersch_NEURIPS2020_CTX}
Doersch, C., Gupta, A., Zisserman, A.: Crosstransformers: spatially-aware
  few-shot transfer. In: Larochelle, H., Ranzato, M., Hadsell, R., Balcan,
  M.F., Lin, H. (eds.) Advances in Neural Information Processing Systems.
  vol.~33, pp. 21981--21993. Curran Associates, Inc. (2020),
  \url{https://proceedings.neurips.cc/paper/2020/file/fa28c6cdf8dd6f41a657c3d7caa5c709-Paper.pdf}

\bibitem{dosovitskiy2021an}
Dosovitskiy, A., Beyer, L., Kolesnikov, A., Weissenborn, D., Zhai, X.,
  Unterthiner, T., Dehghani, M., Minderer, M., Heigold, G., Gelly, S.,
  Uszkoreit, J., Houlsby, N.: An image is worth 16x16 words: Transformers for
  image recognition at scale. In: International Conference on Learning
  Representations (2021)

\bibitem{YOLOS}
Fang, Y., Liao, B., Wang, X., Fang, J., Qi, J., Wu, R., Niu, J., Liu, W.: You
  only look at one sequence: Rethinking transformer in vision through object
  detection. Advances in Neural Information Processing Systems 34
  pre-proceedings  (2021)

\bibitem{FinnAL17}
Finn, C., Abbeel, P., Levine, S.: Model-agnostic meta-learning for fast
  adaptation of deep networks. In: Proceedings of the 34th International
  Conference on Machine Learning (2017)

\bibitem{Graham_2021_ICCV}
Graham, B., El-Nouby, A., Touvron, H., Stock, P., Joulin, A., J\'egou, H.,
  Douze, M.: Levit: A vision transformer in convnet's clothing for faster
  inference. In: Proceedings of the IEEE/CVF International Conference on
  Computer Vision (ICCV). pp. 12259--12269 (October 2021)

\bibitem{graves2014neural}
Graves, A., Wayne, G., Danihelka, I.: Neural turing machines. arXiv preprint
  arXiv:1410.5401  (2014)

\bibitem{He2016DeepRL}
He, K., Zhang, X., Ren, S., Sun, J.: Deep residual learning for image
  recognition. 2016 IEEE Conference on Computer Vision and Pattern Recognition
  (CVPR) pp. 770--778 (2016)

\bibitem{hinton2015distill}
Hinton, G., Vinyals, O., Dean, J.: Distilling the knowledge in a neural
  network. In: NIPS Deep Learning and Representation Learning Workshop (2015),
  \url{http://arxiv.org/abs/1503.02531}

\bibitem{Hou_NEURIPS2019_CAN}
Hou, R., Chang, H., MA, B., Shan, S., Chen, X.: Cross attention network for
  few-shot classification. In: Wallach, H., Larochelle, H., Beygelzimer, A.,
  d\textquotesingle Alch\'{e}-Buc, F., Fox, E., Garnett, R. (eds.) Advances in
  Neural Information Processing Systems. vol.~32. Curran Associates, Inc.
  (2019),
  \url{https://proceedings.neurips.cc/paper/2019/file/01894d6f048493d2cacde3c579c315a3-Paper.pdf}

\bibitem{jiang2021all}
Jiang, Z., Hou, Q., Yuan, L., Daquan, Z., Shi, Y., Jin, X., Wang, A., Feng, J.:
  All tokens matter: Token labeling for training better vision transformers.
  In: Advances in Neural Information Processing Systems (2021)

\bibitem{Kang_2021_ICCV}
Kang, D., Kwon, H., Min, J., Cho, M.: Relational embedding for few-shot
  classification. In: Proceedings of the IEEE/CVF International Conference on
  Computer Vision (ICCV). pp. 8822--8833 (October 2021)

\bibitem{Krizhevsky2009LearningML}
Krizhevsky, A.: Learning multiple layers of features from tiny images (2009)

\bibitem{Lee2019MetaLearningWD}
Lee, K., Maji, S., Ravichandran, A., Soatto, S.: Meta-learning with
  differentiable convex optimization. 2019 IEEE/CVF Conference on Computer
  Vision and Pattern Recognition (CVPR) pp. 10649--10657 (2019)

\bibitem{li2020adversarial}
Li, K., Zhang, Y., Li, K., Fu, Y.: Adversarial feature hallucination networks
  for few-shot learning. In: 2020 IEEE/CVF Conference on Computer Vision and
  Pattern Recognition (CVPR) (2020)

\bibitem{li2021vision}
Li, S., Chen, X., He, D., Hsieh, C.J.: Can vision transformers perform
  convolution? (2021)

\bibitem{li2021localvit}
Li, Y., Zhang, K., Cao, J., Timofte, R., Van~Gool, L.: Localvit: Bringing
  locality to vision transformers. arXiv preprint arXiv:2104.05707  (2021)

\bibitem{Lifchitz_2019_CVPR}
Lifchitz, Y., Avrithis, Y., Picard, S., Bursuc, A.: Dense classification and
  implanting for few-shot learning. In: Proceedings of the IEEE/CVF Conference
  on Computer Vision and Pattern Recognition (CVPR) (June 2019)

\bibitem{liu2020negative}
Liu, B., Cao, Y., Lin, Y., Li, Q., Zhang, Z., Long, M., Hu, H.: Negative margin
  matters: Understanding margin in few-shot classification. In: European
  Conference on Computer Vision. pp. 438--455 (2020)

\bibitem{liu2021pay}
Liu, H., Dai, Z., So, D., Le, Q.: Pay attention to mlps. Advances in Neural
  Information Processing Systems  \textbf{34} (2021)

\bibitem{liu2021a}
Liu, L., Hamilton, W.L., Long, G., Jiang, J., Larochelle, H.: A universal
  representation transformer layer for few-shot image classification. In:
  International Conference on Learning Representations (2021),
  \url{https://openreview.net/forum?id=04cII6MumYV}

\bibitem{liu2021efficient}
Liu, Y., Sangineto, E., Bi, W., Sebe, N., Lepri, B., Nadai, M.D.: Efficient
  training of visual transformers with small datasets. In: Beygelzimer, A.,
  Dauphin, Y., Liang, P., Vaughan, J.W. (eds.) Advances in Neural Information
  Processing Systems (2021), \url{https://openreview.net/forum?id=SCN8UaetXx}

\bibitem{Liu_2021_ICCV}
Liu, Z., Lin, Y., Cao, Y., Hu, H., Wei, Y., Zhang, Z., Lin, S., Guo, B.: Swin
  transformer: Hierarchical vision transformer using shifted windows. In:
  Proceedings of the IEEE/CVF International Conference on Computer Vision
  (ICCV). pp. 10012--10022 (October 2021)

\bibitem{Loshchilov2017SGDRSG}
Loshchilov, I., Hutter, F.: Sgdr: Stochastic gradient descent with warm
  restarts. International Conference on Learning Representations  (2017)

\bibitem{loshchilov2018decoupled}
Loshchilov, I., Hutter, F.: Decoupled weight decay regularization. In:
  International Conference on Learning Representations (2019)

\bibitem{luo2021rectifying}
Luo, X., Wei, L., Wen, L., Yang, J., Xie, L., Xu, Z., Tian, Q.: Rectifying the
  shortcut learning of background for few-shot learning. In: Beygelzimer, A.,
  Dauphin, Y., Liang, P., Vaughan, J.W. (eds.) Advances in Neural Information
  Processing Systems (2021), \url{https://openreview.net/forum?id=N1i6BJzouX4}

\bibitem{vandermaaten08tsne}
van~der Maaten, L., Hinton, G.: Visualizing data using t-sne. Journal of
  Machine Learning Research  \textbf{9}(86),  2579--2605 (2008),
  \url{http://jmlr.org/papers/v9/vandermaaten08a.html}

\bibitem{mishra2018a}
Mishra, N., Rohaninejad, M., Chen, X., Abbeel, P.: A simple neural attentive
  meta-learner. In: International Conference on Learning Representations
  (2018), \url{https://openreview.net/forum?id=B1DmUzWAW}

\bibitem{oreshkin2018tadam}
Oreshkin, B., Rodr\'{\i}guez~L\'{o}pez, P., Lacoste, A.: Tadam: Task dependent
  adaptive metric for improved few-shot learning. In: Bengio, S., Wallach, H.,
  Larochelle, H., Grauman, K., Cesa-Bianchi, N., Garnett, R. (eds.) Advances in
  Neural Information Processing Systems. vol.~31. Curran Associates, Inc.
  (2018),
  \url{https://proceedings.neurips.cc/paper/2018/file/66808e327dc79d135ba18e051673d906-Paper.pdf}

\bibitem{Ravi2017OptimizationAA}
Ravi, S., Larochelle, H.: Optimization as a model for few-shot learning. In:
  ICLR (2017)

\bibitem{ren2018metalearning}
Ren, M., Ravi, S., Triantafillou, E., Snell, J., Swersky, K., Tenenbaum, J.B.,
  Larochelle, H., Zemel, R.S.: Meta-learning for semi-supervised few-shot
  classification. In: International Conference on Learning Representations
  (2018), \url{https://openreview.net/forum?id=HJcSzz-CZ}

\bibitem{rusu2018metalearning}
Rusu, A.A., Rao, D., Sygnowski, J., Vinyals, O., Pascanu, R., Osindero, S.,
  Hadsell, R.: Meta-learning with latent embedding optimization. In:
  International Conference on Learning Representations (2019),
  \url{https://openreview.net/forum?id=BJgklhAcK7}

\bibitem{snell2017prototypical}
Snell, J., Swersky, K., Zemel, R.: Prototypical networks for few-shot learning.
  In: Proceedings of the 31st International Conference on Neural Information
  Processing Systems. pp. 4080--4090 (2017)

\bibitem{Strudel_2021_ICCV}
Strudel, R., Garcia, R., Laptev, I., Schmid, C.: Segmenter: Transformer for
  semantic segmentation. In: Proceedings of the IEEE/CVF International
  Conference on Computer Vision (ICCV). pp. 7262--7272 (October 2021)

\bibitem{sun2017revisiting}
Sun, C., Shrivastava, A., Singh, S., Gupta, A.: Revisiting unreasonable
  effectiveness of data in deep learning era. In: Proceedings of the IEEE
  international conference on computer vision. pp. 843--852 (2017)

\bibitem{sun2019mtl}
Sun, Q., Liu, Y., Chua, T., Schiele, B.: Meta-transfer learning for few-shot
  learning. In: The IEEE Conference on Computer Vision and Pattern Recognition
  (CVPR) (June 2019)

\bibitem{tian2020rethink}
Tian, Y., Wang, Y., Krishnan, D., Tenenbaum, J.B., Isola, P.: Rethinking
  few-shot image classification: a good embedding is all you need? ECCV  (2020)

\bibitem{tolstikhin2021mlp}
Tolstikhin, I.O., Houlsby, N., Kolesnikov, A., Beyer, L., Zhai, X.,
  Unterthiner, T., Yung, J., Steiner, A., Keysers, D., Uszkoreit, J., et~al.:
  Mlp-mixer: An all-mlp architecture for vision. Advances in Neural Information
  Processing Systems  \textbf{34} (2021)

\bibitem{pmlr-v139-touvron21a}
Touvron, H., Cord, M., Douze, M., Massa, F., Sablayrolles, A., Jegou, H.:
  Training data-efficient image transformers and distillation through
  attention. In: International Conference on Machine Learning. vol.~139, pp.
  10347--10357 (July 2021)

\bibitem{Touvron_2021_ICCV}
Touvron, H., Cord, M., Sablayrolles, A., Synnaeve, G., J\'egou, H.: Going
  deeper with image transformers. In: Proceedings of the IEEE/CVF International
  Conference on Computer Vision (ICCV). pp. 32--42 (October 2021)

\bibitem{vaswani2017attention}
Vaswani, A., Shazeer, N., Parmar, N., Uszkoreit, J., Jones, L., Gomez, A.N.,
  Kaiser, {\L}., Polosukhin, I.: Attention is all you need. In: Advances in
  neural information processing systems. pp. 5998--6008 (2017)

\bibitem{vinyals2016matching}
Vinyals, O., Blundell, C., Lillicrap, T., Wierstra, D., et~al.: Matching
  networks for one shot learning. In: Advances in Neural Information Processing
  Systems. pp. 3630--3638 (2016)

\bibitem{Wang_2021_ICCV}
Wang, W., Xie, E., Li, X., Fan, D.P., Song, K., Liang, D., Lu, T., Luo, P.,
  Shao, L.: Pyramid vision transformer: A versatile backbone for dense
  prediction without convolutions. In: Proceedings of the IEEE/CVF
  International Conference on Computer Vision (ICCV). pp. 568--578 (October
  2021)

\bibitem{weng2021semisupervised}
Weng, Z., Yang, X., Li, A., Wu, Z., Jiang, Y.G.: Semi-supervised vision
  transformers (2021)

\bibitem{Wu_2021_ICCV_CvT}
Wu, H., Xiao, B., Codella, N., Liu, M., Dai, X., Yuan, L., Zhang, L.: Cvt:
  Introducing convolutions to vision transformers. In: Proceedings of the
  IEEE/CVF International Conference on Computer Vision (ICCV). pp. 22--31
  (October 2021)

\bibitem{Wu_2021_ICCV}
Wu, J., Zhang, T., Zhang, Y., Wu, F.: Task-aware part mining network for
  few-shot learning. In: Proceedings of the IEEE/CVF International Conference
  on Computer Vision (ICCV). pp. 8433--8442 (October 2021)

\bibitem{xiao2021early}
Xiao, T., Dollar, P., Singh, M., Mintun, E., Darrell, T., Girshick, R.: Early
  convolutions help transformers see better. In: Beygelzimer, A., Dauphin, Y.,
  Liang, P., Vaughan, J.W. (eds.) Advances in Neural Information Processing
  Systems (2021)

\bibitem{Xu_2021_ICCV}
Xu, W., Xu, Y., Chang, T., Tu, Z.: Co-scale conv-attentional image
  transformers. In: Proceedings of the IEEE/CVF International Conference on
  Computer Vision (ICCV). pp. 9981--9990 (October 2021)

\bibitem{ye2020fewshot}
Ye, H.J., Hu, H., Zhan, D.C., Sha, F.: Few-shot learning via embedding
  adaptation with set-to-set functions. In: IEEE/CVF Conference on Computer
  Vision and Pattern Recognition (CVPR). pp. 8808--8817 (2020)

\bibitem{yu2021improving}
Yu, P., Chen, Y., Jin, Y., Liu, Z.: Improving vision transformers for
  incremental learning (2021)

\bibitem{yu2021metaformer}
Yu, W., Luo, M., Zhou, P., Si, C., Zhou, Y., Wang, X., Feng, J., Yan, S.:
  Metaformer is actually what you need for vision. arXiv preprint
  arXiv:2111.11418  (2021)

\bibitem{Yuan_2021_ICCV}
Yuan, L., Chen, Y., Wang, T., Yu, W., Shi, Y., Jiang, Z.H., Tay, F.E., Feng,
  J., Yan, S.: Tokens-to-token vit: Training vision transformers from scratch
  on imagenet. In: Proceedings of the IEEE/CVF International Conference on
  Computer Vision (ICCV). pp. 558--567 (October 2021)

\bibitem{YuanFHLZCW21}
Yuan, Y., Fu, R., Huang, L., Lin, W., Zhang, C., Chen, X., Wang, J.: Hrformer:
  High-resolution transformer for dense prediction. NeurIPS  (2021)

\bibitem{Zhang_2020_CVPR}
Zhang, C., Cai, Y., Lin, G., Shen, C.: Deepemd: Few-shot image classification
  with differentiable earth mover's distance and structured classifiers. In:
  IEEE/CVF Conference on Computer Vision and Pattern Recognition (CVPR) (June
  2020)

\bibitem{zhang2021bootstrapping}
Zhang, H., Duan, J., Xue, M., Song, J., Sun, L., Song, M.: Bootstrapping vits:
  Towards liberating vision transformers from pre-training (2021)

\bibitem{zhang2021aggregating}
Zhang, Z., Zhang, H., Zhao, L., Chen, T., , Arik, S.O., Pfister, T.: Nested
  hierarchical transformer: Towards accurate, data-efficient and interpretable
  visual understanding. In: AAAI Conference on Artificial Intelligence (AAAI)
  (2022)

\bibitem{Zheng_2021_CVPR}
Zheng, S., Lu, J., Zhao, H., Zhu, X., Luo, Z., Wang, Y., Fu, Y., Feng, J.,
  Xiang, T., Torr, P.H., Zhang, L.: Rethinking semantic segmentation from a
  sequence-to-sequence perspective with transformers. In: Proceedings of the
  IEEE/CVF Conference on Computer Vision and Pattern Recognition (CVPR). pp.
  6881--6890 (June 2021)

\bibitem{zhmoginov2022hypertransformer}
Zhmoginov, A., Sandler, M., Vladymyrov, M.: Hypertransformer: Model generation
  for supervised and semi-supervised few-shot learning (2022)

\bibitem{zhong2020random}
Zhong, Z., Zheng, L., Kang, G., Li, S., Yang, Y.: Random erasing data
  augmentation. In: Proceedings of the AAAI conference on artificial
  intelligence. vol.~34, pp. 13001--13008 (2020)

\bibitem{zhou2019metalearning}
Zhou, P., Yuan, X., Xu, H., Yan, S., Feng, J.: Efficient meta learning via
  minibatch proximal update. In: Neural Information Processing Systems (2019)

\bibitem{zhou2021task}
Zhou, P., Zou, Y., Yuan, X.T., Feng, J., Xiong, C., Hoi, S.: Task similarity
  aware meta learning: Theory-inspired improvement on maml. In: Uncertainty in
  Artificial Intelligence. pp. 23--33. PMLR (2021)

\bibitem{Zhou_2021_ICCV}
Zhou, Z., Qiu, X., Xie, J., Wu, J., Zhang, C.: Binocular mutual learning for
  improving few-shot classification. In: Proceedings of the IEEE/CVF
  International Conference on Computer Vision (ICCV). pp. 8402--8411 (October
  2021)

\end{thebibliography}
\clearpage
\appendix
\begin{center}
  \Large
  \textbf{Appendix}
\end{center}

The content of Appendix is summarized as follows: 0) in Sec.~\ref{sec:justification}, we discuss the justification of using ViT in few-shot learning scenarios; 1) in Sec.~\ref{sec:network_arch}, we list the network architectures we used in the experiment; 2) in Sec.~\ref{sec:moredetails}, we state implementation and training details to ensure that our SUN can be reproduced; 3) in Sec.~\ref{sec:sun-f} and Sec.~\ref{sec:sun-d}, we introduce the detail of SUN with FEAT (SUN-F) and DeepEMD (SUN-D), then demonstrating its performance; 4) in Sec.~\ref{sec:more_ablation}, we conduct more ablation study to analyze other components of SUN; 5) and in Sec.~\ref{sec:visualization}, we conduct t-SNE visualization to qualitatively evaluate ViT with SUN, thus demonstrating the effectiveness of SUN. 
\begin{table}\label{table:models}
  \centering
  \caption{Detailed layer specification of the ViTs we used~\cite{jiang2021all,Liu_2021_ICCV,zhang2021aggregating,Chen_2021_ICCV} for few-shot classification. All ViTs are scaled to the size with $\sim$12.5M parameters such that they own approximately same parameter number with the widely-used ResNet-12. 
  Specifically, $k\times k$ means convolution operation with kernel size of $k$, \emph{d} means channel dimension, \emph{s} means stride, ``res'' means the residual connection from the input via a $3\times 3$ convolution with \emph{s}=2, MaxPool means max pooling operation, MHSA means self-attention layer, S-MHSA means shifted self-attention layer proposed by~\cite{Liu_2021_ICCV}, FFN means feed-forward layer and C-FFN means FFN with $3\times 3$ depthwise convolution. 
  We strictly follow the corresponding official implementation to build these ViTs. And to simplify the layer description, we omit normalization layers in the table. 
  }
  \setlength{\tabcolsep}{2.5pt} 
     \renewcommand{\arraystretch}{4}
   { \fontsize{6.5}{3}\selectfont{
  \begin{tabular}{c|c|c|c|c|c}
  \toprule
    {Stage}& {Layers} &{LV-ViT}~\cite{jiang2021all}&{Swin Transformer}~\cite{Liu_2021_ICCV}&{NesT}~\cite{zhang2021aggregating}&{Visformer}~\cite{Chen_2021_ICCV} \\
  \midrule
  \multirow{3}{*}{1} & {\shortstack{{Patch}\\{Embedding}\\$ $\\$ $\\$ $\\$ $ }} 
  & \shortstack{{$3\times 3$, \emph{d}=96, \emph{s}=2}\\{$3\times 3$, \emph{d}=96}\\{$3\times 3$, \emph{d}=96 (+res)}\\{MaxPool, 2x2}\\{$4\times 4$, \emph{d}=384, \emph{s}=4}}  
  & \shortstack{{$3\times 3$, \emph{d}=64, \emph{s}=2}\\{$3\times 3$, \emph{d}=64}\\{$3\times 3$, \emph{d}=144 (+res)}\\{MaxPool, 2x2}\\$ $} 
  & \shortstack{{$3\times 3$, \emph{d}=64, \emph{s}=2}\\{$3\times 3$, \emph{d}=64}\\{$3\times 3$, \emph{d}=128 (+res)}\\{MaxPool, 2x2}\\$ $} 
  & \shortstack{{$3\times 3$, \emph{d}=32, \emph{s}=2}\\{$3\times 3$, \emph{d}=32}\\{$3\times 3$, \emph{d}=128 (+res)}\\{MaxPool, 2x2}\\$ $} \\
  \cline{2-6} & {\shortstack{{ViT}\\{Blocks}\\$ $}} 
  & \shortstack{$ $\\{MHSA, \emph{d}=384}\\{FFN, ratio=3}\\$ $}
  & \shortstack{$ $\\{MHSA, \emph{d}=144}\\{S-MHSA, \emph{d}=144}\\{FFN, ratio=4}} 
  & \shortstack{{MHSA, \emph{d}=384}\\{FFN, ratio=4}\\$ $} 
  & \shortstack{$ $\\{MHSA, \emph{d}=384}\\{C-FFN, ratio=4}\\$ $} \\
  \cline{2-6} & {Blocks Num} 
  & 8
  & 2 
  & 2 
  & 4 \\
  \hline
  \multirow{3}{*}{2} & {\shortstack{{Patch}\\{Embedding}}} 
  & \shortstack{{/}\\$ $}
  & \shortstack{{$2\times 2$, \emph{d}=288, \emph{s}=2}\\$ $} 
  & \shortstack{$ $\\{$3\times 3$, \emph{d}=384}\\{MaxPool, $2\times 2$}} 
  & \shortstack{{$2\times 2$, \emph{d}=256, \emph{s}=2}\\$ $} \\
  \cline{2-6} & {\shortstack{{ViT}\\{Blocks}\\$ $}} 
  & \shortstack{$ $\\{/}\\$ $\\$ $}
  & \shortstack{$ $\\{MHSA, \emph{d}=288}\\{S-MHSA, \emph{d}=288}\\{FFN, ratio=4}} 
  & \shortstack{{MHSA, \emph{d}=384}\\{FFN, ratio=4}\\$ $} 
  & \shortstack{$ $\\{MHSA, \emph{d}=256}\\{C-FFN, ratio=4}\\$ $} \\
  \cline{2-6} & {Blocks Num} 
  & /
  & 3 
  & 2 
  & 2 \\
  \hline
  \multirow{3}{*}{3} & {\shortstack{{Patch}\\{Embedding}}} 
  & \shortstack{{/}\\$ $}
  & \shortstack{{$2\times 2$, \emph{d}=576, \emph{s}=2}\\$ $} 
  & \shortstack{$ $\\{$3\times 3$, \emph{d}=512}\\{MaxPool, $2\times 2$}} 
  & \shortstack{{$2\times 2$, \emph{d}=512, \emph{s}=2}\\$ $} \\
  \cline{2-6} & {\shortstack{{ViT}\\{Blocks}\\$ $}} 
  & \shortstack{$ $\\{/}\\$ $\\$ $}
  & \shortstack{$ $\\{MHSA, \emph{d}=576}\\{S-MHSA, \emph{d}=576}\\{FFN, ratio=4}} 
  & \shortstack{{MHSA, \emph{d}=512}\\{FFN, ratio=4}\\$ $} 
  & \shortstack{$ $\\{MHSA, \emph{d}=512}\\{C-FFN, ratio=4}\\$ $} \\
  \cline{2-6} & {Blocks Num} 
  & /
  & 2 
  & 2 
  & 3 \\
  \midrule
  \multicolumn{2}{c|}{Parameters~(M)}& 12.6 &  12.6 & 12.8 & 12.5 \\
  \bottomrule
  \end{tabular}}}
  \end{table}

\section{Why Using ViT for Few-Shot Learning}\label{sec:justification}
The justification of using ViTs for few-shot learning are two-fold: \textbf{a)} though ViT cannot well handle few-shot learning as CNN now, it has three adantages over CNN. 1) ViTs often achieve better performance than CNNs of the same model size when training data is at moderate-scale; 2) ViTs exhibit great potential to unify vision and language models, while existing CNNs mainly work well on vision tasks; 3) ViTs are highly parallelized, and can be more ficiently trained and tested than CNNs. So we hope to build a few-shot ViT training framework to release the above powers in few shot learning. 
Meanwhile, \textbf{b)} our SUN framework uses few-shot learning as an example to prove that ViTs can indeed perform well on such scenarios. 

\section{Network Architecture}\label{sec:network_arch}
We evaluate our SUN on four different ViTs, \ie, LV-ViT~\cite{jiang2021all} (standard ViT), Swin Transformer~\cite{Liu_2021_ICCV} (shifting-window ViT), Visformer~\cite{Chen_2021_ICCV} (CNN-enhanced ViT) and NesT~\cite{zhang2021aggregating} (locality-enhanced ViT), which cover most of existing ViT types. For a fair comparison, we scale the depth and width of these ViTs such that their model sizes are similar to ResNet-12~\cite{He2016DeepRL} ($\sim$12.5M parameters) which is the most commonly used architecture and achieves (nearly) state-of-the-art performance on few-shot classification tasks. 

To make our SUN easier to reproduce, we list the network architectures of ViTs we used in this paper. The detailed layer specification of these ViTs is shown as Table~\ref{table:models}. Specifically, the single stage LV-ViT includes a three-layer overlapped patch embedding with residual connection and eight stacking standard transformer encoder blocks. Given input image with $80\times 80$ resolution, it obtains a 384-dimension feature embedding. And for the multi-stage NesT, which is used in the whole Sec.~5, consists of three stages and each stage contains two transformer encoder layers. Given input image with $80\times 80$ resolution, it obtains a 512-dimensional feature embedding.

\section{More Training Implementation Details}\label{sec:moredetails}
\subsection{Datasets Details}
Following~\cite{chen2019closerfewshot,Chen_2021_ICCV_metabaseline,Zhou_2021_ICCV}, we evaluate ViTs with SUN on three widely used few-shot benchmarks, \ie, CIFAR-FS~\cite{bertinetto2018metalearning}, \emph{mini}ImageNet~\cite{vinyals2016matching} and \emph{tiered}ImageNet~\cite{ren2018metalearning} datasets.

\subsubsection{\emph{mini}ImageNet}\cite{vinyals2016matching} contains 100 different categories chosen from ImageNet-1k dataset~\cite{deng2009imagenet}, and each category includes 600 samples. Here we follow~\cite{chen2019closerfewshot,Chen_2021_ICCV_metabaseline,Zhou_2021_ICCV} and split \emph{mini}ImageNet into 64/16/20 classes for \emph{train}, \emph{val} and \emph{test} sets, respectively. 

\subsubsection{\emph{tiered}ImageNet}\cite{ren2018metalearning} is a larger subset of ImageNet-1k dataset, which totally includes 779,165 images from 608 different categories. Specifically, this dataset includes 351 classes for \emph{training} set, 97 classes for \emph{validation} set and 160 classes for \emph{test} set, respectively. 

\subsubsection{CIFAR-FS}\cite{bertinetto2018metalearning} is built upon CIFAR-100 dataset~\cite{Krizhevsky2009LearningML}, which is divided into 64, 16 and 20 categories for training, validation and testing, respectively. Analogous to \emph{mini}ImageNet, each category includes 600 different images. 

\subsection{Implementation Details of our Analysis in Sec.~\ref{sec:comp_samevit}}\label{sec:implement_analysis}

\subsubsection{\cite{Chen_2021_ICCV_metabaseline}+CNN~\cite{weng2021semisupervised}.}
Here we aim to introduce inductive bias via explicitly introducing CNN layers.
Generally, adding CNN layers has two methods: 1) directly inserting CNN layers after the transformer layers (like Visformer~\cite{Chen_2021_ICCV}); 2) adding independent CNN modules, and fuse the features from both transforme layer and CNN modules. Here we mainly discuss 2). 
%
%
Specifically, with given transformer layer $T$ and a CNN module $C$ (the input and output dimensions of $T$ and $C$ are same), given the input image feature $\mathbf{z}$, the updated feature is obtained by $\mathbf{z}'=T(\mathbf{z})+C(\mathbf{z})$. 
%
We use this combination to replace each transformer layer in the original ViT, and obtain \cite{Chen_2021_ICCV_metabaseline}+CNN.

\subsubsection{\cite{Chen_2021_ICCV_metabaseline}+DrLoc~\cite{liu2021efficient}.}
Here we introduce the implementation detail of \cite{Chen_2021_ICCV_metabaseline}+DrLoc~\cite{liu2021efficient} mentioned in Sec.~5.3 of the main paper. 
With given meta-learner $f$, we additionally introduced a three-layer MLP projection head $h$ and the output dimension of $h$ is 2, which indicates the relative distance between two local tokens on $x$-axis and $y$-axis. 
For given local patches $[\mathbf{z}_{\text{cls}}, \mathbf{z}_{\text{1}}, \mathbf{z}_{\text{2}}, \cdots,  \mathbf{z}_{{K}}]$, we first sample $m$ (\eg, $m=64$) different token pairs $\{(\mathbf{z}_{{i}}, \mathbf{z}_{{j}})\}$. 
For each $(\mathbf{z}_{{i}}, \mathbf{z}_{{j}})$, we calculate the relative distance by $(\delta x', \delta y') = h(\text{concat}(\mathbf{z}_{{i}}, \mathbf{z}_{{j}}))$. 
Since the ground-truth of relative distance $(\delta x, \delta y)$ can be directly calculated, the overall training objective can be written as:
\begin{equation}\label{eq:loss_drloc}
  \mathcal{L}_{\scriptsize{\mbox{drloc}}} = H(g_{\scriptsize{\text{global}}}(\mathbf{z}_{\text{global}}), \mathbf{y}_{i}) + L_{\text{1}}(\delta x', \delta x) + L_{\text{1}}(\delta y', \delta y).
\end{equation}

\subsubsection{\cite{Chen_2021_ICCV_metabaseline}+CNN Distill.}
Here we aim to introduce CNN-alike inductive bias from a pretrained CNN. 
Specifically, we first train a ResNet-12 feature extractor $f^{\text{CNN}}_{0}$ with corresponding global classifier $g^{\text{CNN}}_{0}$ on $\mathbb{D}_{\text{base}}$, and use them to teach ViT feature extractor $f$ and corresponding global classifier $g_{\text{global}}$ via knowledge distillation~\cite{hinton2015distill}. 
Following the description in Sec.~4.1, given $f^{\text{CNN}}_{0}$ with $g^{\text{CNN}}_{0}$, as well as the target meta-learner $f$ with global classifier $g_{\text{global}}$, we denote the the classification result from the global average pooling of all patch tokens calculated by $g^{\text{CNN}}_{0}$ is $g^{\text{CNN}}_{\scriptsize{0}}(\mathbf{z}_{0,\text{global}})$. 
Analogously, that calculated by target classifier $g_{\text{global}}$ is denoted as $g_{\scriptsize{\text{global}}}(\mathbf{z}_{\text{global}})$. 
Thus, the training objective is formed as:
\begin{equation}\label{eq:distill_loss}
  \mathcal{L}_{\scriptsize{\mbox{distill}}} = H(g_{\scriptsize{\text{global}}}(\mathbf{z}_{\text{global}}), \mathbf{y}_{i}) + \text{JSD}(g^{\text{CNN}}_{\scriptsize{0}}(\mathbf{z}_{0,\text{global}}), g_{\scriptsize{\text{global}}}(\mathbf{z}_{\text{global}})),
\end{equation}
where JSD($\cdot,\cdot$) indicates JS-Divergence between $g^{\text{CNN}}_{\scriptsize{0}}(\mathbf{z}_{0,\text{global}})$ and $g_{\scriptsize{\text{global}}}(\mathbf{z}_{\text{global}})$.
\begin{table}[t]
  \caption{Details of the Spatial-Consistent Augmentation (SCA) proposed in Sec.~4, where \emph{p} means the random probability to perform the corresponding augmentation. }
  \begin{center}
  \label{tab:sca_detail}
      \setlength{\tabcolsep}{3.6pt} 
 \renewcommand{\arraystretch}{4}
{ \fontsize{8.3}{3}\selectfont{
  \begin{tabular}{c|c}
  \toprule
 Spatial-Only Transformation & Non-Spatial Transformation  \\
  \hline
  Random Crop and Resize & ColorJitter (brightness=0.4, contrast=0.4, saturation=0.4)  \\
  Random Horizontal Flip, \emph{p}=0.5 & Gaussian Blur, \emph{p}=0.5 \\
  Random Rotation, \emph{p}=0.2 & Solarization, \emph{p}=0.5 \\
   / & GrayScale, \emph{p}=0.2 \\
   / & Random Erasing \\
  \toprule
  \end{tabular}}}
  \end{center}
  \end{table}
\subsection{More Training Details of SUN}
For meta-training phase, we use AdamW~\cite{loshchilov2018decoupled} with a learning rate of 5e-4 and a cosine learning rate scheduler~\cite{Loshchilov2017SGDRSG} to train our meta learner $f$.  Specifically, we train our model for 800, 800, 300 epochs on \emph{mini}ImageNet, CIFAR-FS and \emph{tiered}ImageNet, respectively.  For data augmentation, we use Spatial-Consistent Augmentation in Sec.~4. 
To completely describe the spatial-consistent augmentation, Table~\ref{tab:sca_detail} lists the detailed augmentation for both spatial-only part and non-spatial part. 
For location-specific supervision on each patch, we only keep the top-$k$ (\eg, $k=5$) highest confidence in $\mathbf{\hat{s}}_{ij}$  to reduce the label noise.  
To train the teacher $f_g$ (the combination of feature extractor $f_{0}$ with global classifier $g_{0}$), we employ the same augmentation strategy in~\cite{pmlr-v139-touvron21a}, including random crop, random augmentation, mixup, etc.  
Meanwhile, we adopt AdamW~\cite{loshchilov2018decoupled} with the same settings as above to train it for 300 epochs. 
%
%
For meta-tuning, we simply utilize the same optimizer and settings as Meta-Baseline, \ie, SGD with learning rate of 1e-3 to finetune the meta-learner $f$ for 40 epochs. 
Moreover, we use relatively large drop path rate 0.5 to avoid overfitting for all training. 
This greatly differs from conventional setting on drop path rate where it often uses 0.1~\cite{pmlr-v139-touvron21a,Touvron_2021_ICCV}.  
Following conventional supervised setting~\cite{xiao2021early,jiang2021all,yu2021improving}, we also use a three-layer convolution block~\cite{He2016DeepRL} with residual connection to compute patch embedding. 
This conventional stem  has only  $\sim$0.2M  parameters and  is much smaller than ViT backbone. 

\begin{table}[t]
  \centering
  \caption{Comparison with SoTA few-shot learning methods under  5-way few-shot classification setting, where SUN-F is our proposed SUN with FEAT~\cite{ye2020fewshot} as meta-tuning phase. The results of the best 2 methods are in bold font.}
  \label{table:SoTAfs_sunf}
      \setlength{\tabcolsep}{0.8pt} 
 \renewcommand{\arraystretch}{2.4}
{ \fontsize{6.5}{3}\selectfont{
  \begin{threeparttable}
  \begin{tabular}{lccccccc}
  \toprule
  \multirow{2}{*}{\textbf{Method}}&\multirow{2}{*}{\shortstack{\textbf{Classifier}\\\textbf{Params}}}&\multicolumn{2}{c}{\textbf{\emph{mini}ImageNet}}&\multicolumn{2}{c}{\textbf{\emph{tiered}ImageNet}}&\multicolumn{2}{c}{\textbf{CIFAR-FS}}
  \\ \cline{3-8}
  &&{{1-shot}}&{{5-shot}}&{{1-shot}}&{{5-shot}}&{{1-shot}}&{{5-shot}}
  \\ \toprule
 \multicolumn{7}{l}{ \textbf{ResNet-12/18 as feature extractor}  }\\
  
  {MetaOptNet}~\cite{Lee2019MetaLearningWD}&0&64.09$\pm$0.62&80.00$\pm$0.45&65.81$\pm$0.74&81.75$\pm$0.53&72.00$\pm$0.70&84.20$\pm$0.50
  \\ 
  {DeepEMD}~\cite{Zhang_2020_CVPR}&0&65.91$\pm$0.82&82.41$\pm$0.56&71.16$\pm$0.80&86.03$\pm$0.58&46.47$\pm$0.70&63.22$\pm$0.71
  \\ 
  {FEAT}~\cite{ye2020fewshot}&1.05M&66.78$\pm$0.20&82.05$\pm$0.14&70.80$\pm$0.23&84.79$\pm$0.16&-&-
  \\ 
  {TADAM}~\cite{oreshkin2018tadam}&1.23M&58.50$\pm$0.30&76.70$\pm$0.30&-&-&-&-
  \\ 
  {Rethink-Distill}~\cite{tian2020rethink}&225K&64.82$\pm$0.60&82.14$\pm$0.43&71.52$\pm$0.69&86.03$\pm$0.49&73.90$\pm$0.80&86.90$\pm$0.50
  \\
  {DC}~\cite{Lifchitz_2019_CVPR}&224K&61.26$\pm$0.20&79.01$\pm$0.13&-&-&-&-
  \\
  {MTL}~\cite{sun2019mtl}&0&61.20$\pm$1.80&75.50$\pm$0.80&-&-&-&-
  \\
  {CloserLook++}~\cite{chen2019closerfewshot}&131K&51.87$\pm$0.77&75.68$\pm$0.63&-&-&-&-
  \\
  {Meta-Baseline}~\cite{Chen_2021_ICCV_metabaseline}&0&63.17$\pm$0.23&79.26$\pm$0.17&68.62$\pm$0.27&83.29$\pm$0.18&-&-
  \\
  {Neg-Cosine}~\cite{liu2020negative}&131K&63.85$\pm$0.81&81.57$\pm$0.56&-&-&-&-
  \\
  {AFHN}~\cite{li2020adversarial}&359K&62.38$\pm$0.72&78.16$\pm$0.56&-&-&68.32$\pm$0.93&81.45$\pm$0.87
  \\
  {Centroid}~\cite{afrasiyabi2020associative}&10K&59.88$\pm$0.67&80.35$\pm$0.73&69.29$\pm$0.56&85.97$\pm$0.49&-&- \\
  {RE-Net}~\cite{Kang_2021_ICCV}&430K&\textbf{67.60$\pm$0.44}&\textbf{82.58$\pm$0.30}&{71.61$\pm$0.51}&{85.28$\pm$0.35}&{74.51$\pm$0.46}&{86.60$\pm$0.32} \\
  {TPMN}~\cite{Wu_2021_ICCV}&16M&\textbf{67.64$\pm$0.63}&\textbf{83.44$\pm$0.43}&\textbf{72.24$\pm$0.70}&\textbf{86.55$\pm$0.63}&\textbf{75.50$\pm$0.90}&\textbf{87.20$\pm$0.60} \\
  \toprule
   \multicolumn{7}{l}{ \textbf{NesT ViT as feature extractor}  }\\
  {CloserLook++}~\cite{chen2019closerfewshot}&180K&{49.23$\pm$0.43}&{66.57$\pm$0.39}&{59.13$\pm$0.46}&{77.88$\pm$0.39}&{63.89$\pm$0.49}&{80.43$\pm$0.37}
  \\
  {Meta-Baseline}~\cite{Chen_2021_ICCV_metabaseline}&0&{54.57$\pm$0.46}&{69.85$\pm$0.38}&{63.73$\pm$0.47}&{79.33$\pm$0.38}&{68.05$\pm$0.48}&{81.53$\pm$0.36}
  \\
  {BML}~\cite{Zhou_2021_ICCV}&180K&{59.35$\pm$0.45}&{76.00$\pm$0.35}&{66.98$\pm$0.50}&{83.75$\pm$0.34}&{67.51$\pm$0.48}&{82.17$\pm$0.36} \\
  {SUN-F (Ours)}&1.05M&{66.60$\pm$0.44}&{81.90$\pm$0.32}&\textbf{72.66$\pm$0.51}&\textbf{87.08$\pm$0.33}&\textbf{77.87$\pm$0.46}&\textbf{88.75$\pm$0.33}\\
  \toprule
  \end{tabular}
  \end{threeparttable}}}
\end{table}

\section{Using FEAT as Meta-Tuning of SUN}\label{sec:sun-f}
As mentioned in Sec.~4.2 of the main paper, we also investigate different meta-tuning methods, such as introducing the task-specific few-shot learning stage of FEAT~\cite{ye2020fewshot} as our meta-tuning phase. 
With the same FEAT, our SUN framework still shows superiority over other few-shot learning frameworks.
%
%
Here we first give a brief introduction of FEAT, and then compare the SUN with FEAT with existing methods to demonstrate the superiority our method.

Generally, the goals of FEAT are two-fold: 1) all feature embeddings from the support set $\mathbf{S}$ of each task $\tau$ should be aligned by a permutation invariant function to obtain more discriminative prototypes, and 2) aligned feature embedding of each query should be similar to embeddings with the same class and dissimilar to those of other classes. 
Thus, it first introduces a self-attention layer $A$ to obtain aligned classification prototypes ${\mathbf{w}}'_{k} = A(\left\{\mathbf{w}_{k}, \forall 1 \leq k \leq c\right\}),$ 
and then uses   Eqn.~\eqref{eq:p_x_feat} to calculate the confidence score ${\mathbf{p}}'_{k}$ for each query $\mathbf{x}$, 
\begin{equation}\label{eq:p_x_feat}\centering
 {\mathbf{p}}'_{k} = \frac{\exp(\gamma\cdot \cos(GAP(f(\mathbf{x})),  {\mathbf{w}}'_{k}))}{\sum\nolimits_{{k}'}\exp(\gamma\cdot    \cos(GAP(f(\mathbf{x})),  {\mathbf{w}}'_{k})  )}, 
\end{equation}
and then obtains the classification prediction  $\mathbf{p}^{\text{F}}_{\text{x}}=[{\mathbf{p}}'_1, \cdots, {\mathbf{p}}'_c]$.
Meanwhile, for each class $k\in c$, FEAT introduces contrastive learning loss among query embeddings for each task $\tau$. Specifically, FEAT calculates the class center $\mathbf{q}_{c}$ as $\mathbf{q}_{c}= \sum A(\left \{ f({\mathbf{x}}'), {\mathbf{x}}'\in\tau_{k} \right \}) /( N_q+N_k)$, where $\tau_{c}$ indicates all query and support images of class $c$, and $A$ is the same self-attention layer mentioned above. Thus for each query $\mathbf{x}$, FEAT also enforces $f(x)$ aligned by $A$ to be close to the corresponding class center $\mathbf{q}_{c}$, then we obtain: 
\begin{equation}\label{eq:p_x_aux}\centering
 \mathbf{p}_{\mathbf{x}}^{\text{aux}} = \left \{\frac{\exp(\gamma\cdot \langle f(\mathbf{x})\cdot \mathbf{q}_{c}) \rangle }{\sum\nolimits_{{c}'}\exp(\gamma\cdot \langle  f(\mathbf{x})\cdot \mathbf{q}_{{c}'} \rangle )}, \forall c\in \mathbf{S}\right \}.
\end{equation} 
Finally, it minimizes the classification loss $\mathcal{L}_{\text{few-shot}} = H(\mathbf{p}^{\text{F}}_{\text{x}}, \text{y}_{\mathbf{x}})+H(\mathbf{p}_{\mathbf{x}}^{\text{aux}}, \text{y}_{\mathbf{x}})$, where $\text{y}_{\mathbf{x}}$ is the classification label of $\mathbf{x}$ w.r.t. $\mathbf{S}$.
By using this meta-tuning method, we term our method ``SUN-F". 

Analogous to ``SUN-M'' stated in Sec.~5.4, we evaluate SUN-F using NesT~\cite{zhang2021aggregating} on three diferent datasets, \ie, \emph{mini}ImageNet~\cite{vinyals2016matching}, \emph{tiered}ImageNet~\cite{ren2018metalearning} and  CIFAR-FS~\cite{bertinetto2018metalearning}. The detailed evaluation results are given in Table~\ref{table:SoTAfs_sunf}. 
\begin{table}[h]
    \caption{Comparison results  
    among DeepEMD~\cite{Zhang_2020_CVPR}, COSOC~\cite{luo2021rectifying} and SUN  
    under 5-way few-shot classification setting 
    on \emph{mini}ImageNet. 
    }
       \centering
       \begin{tabular}{l | c | c | c }
       \hline
       Methods & DeepEMD~\cite{Zhang_2020_CVPR} & COSOC~\cite{luo2021rectifying} & SUN-D (Ours) \\
       \hline
       5-way 1-shot & 68.77$\pm$0.29 & {69.28$\pm$0.49} & \textbf{69.56$\pm$0.44}   \\
       5-way 5-shot 
       & 84.13$\pm$0.53 &  {85.16$\pm$0.42} & \textbf{85.38$\pm$0.49} \\
       \hline
       \end{tabular}
       \label{table:dense_prediction}
     \end{table}
With the same NesT as meta-learner $f$, our SUN-F achieves the best 5-way 1-shot an 5-way 5-shot accuracy on the three datasets. Specifically, SUN-F outperforms BML by 7.6\%, 5.7\%, 10.3\% in terms of 5-way 1-shot accuracy on \emph{mini}ImageNet, \emph{tiered}ImageNet, CIFAR-FS \emph{test} sets, respectively. 
Moreover, SUN-F also performs very competitively in comparison with state-of-the-art CNN-based few-shot learning methods. 
Specifically, on \emph{tiered}ImageNet under 5-way 1-shot and 5-shot settings, our  SUN-F respectively obtains 72.66\% and 87.08\%, and respectively improves $\sim$0.4\% and $\sim$0.5\% over the SoTA TPMN~\cite{Wu_2021_ICCV}.   
On CIFAR-FS dataset, our SUN-F obtains 77.87\% and 88.75\% in terms of 1-shot accuracy and 5-shot accuracy, which significantly outperforms all the state-of-the-art methods by at least 2.3\% in terms of 1-shot accuracy. 
Meanwhile, our SUN-F also obtains 66.60\% 1-shot accuracy on \emph{mini}ImageNet \emph{test} set, which also surpasses most of CNN-based few-shot learning methods. 

\section{Using DeepEMD as Meta-Tuning of SUN}\label{sec:sun-d}
Besides, ViT with SUN inherently supports incorporating with dense prediction methods to conduct meta-tuning phase. There are two ways to leverage dense feature: \textbf{a)} selecting the foreground patch tokens with the local patch scores $g_{\text{local}}(\textbf{z})$, then calculating the global token (via global average pooling) for classification; and \textbf{b)} simply replacing the meta-tuning method with dense prediction methods like DC~\cite{Lifchitz_2019_CVPR} or DeepEMD~\cite{Zhang_2020_CVPR}. Here we mainly discuss b) and use DeepEMD~\cite{Zhang_2020_CVPR} as an example. From Table~\ref{table:dense_prediction}, SUN-D outperfroms DeepEMD by $\sim$1\% on both 5-way 1-shot and 5-shot accuracy. The results indicate that ViTs with SUN can also leverage dense features to conduct few-shot classification and perform better than CNN counterpart. Moreover, the derived SUN-D achieves 69.56\% 5-way 1-shot accuracy which is slightly higher than the state-of-the-art COSOC~\cite{luo2021rectifying}.  
Note that DeepEMD is not applicable to COSOC, since DeepEMD uses dense feature while COSOC uses global tokens; while SUN can incorporate with various methods. 
Thus we believe that SUN can be incorporated with COSOC and achieves better performance than SUN and COSOC.
\section{More Ablation Study}\label{sec:more_ablation}
\subsection{Effect of Patch Embedding} Now we analyze the effect of overlapped patch embedding  in Sec.~\ref{sec:implementation}. 
For comparison, we replace the overlapped patch embedding in ViT by the vanilla non-overlapped patch embedding, and use the same meta-training method.  As shown in  Table~\ref{table:ablation_phase}, compared to type (a) using overlapped embedding, ViT w/o overlapped embedding  achieves 59.70\% and  77.19\%  in terms of 1-shot and 5-shot accuracy respectively. This result shows that overlapped patch embedding benefits the generalization ability on novel categories.
\begin{table}[t]
  \centering
  \caption{Ablation study of training epochs of meta-training phase on \emph{mini}ImageNet and \emph{tiered}ImageNet, where ``epochs'' means training epochs in the meta-training phase. 
  }
  \label{table:epochs}
  \small
  \setlength{\tabcolsep}{4.0pt} 
 \renewcommand{\arraystretch}{4}
{ \fontsize{8.3}{3}\selectfont{
  \begin{threeparttable}
  \begin{tabular}{cccccc}
  \toprule
  \multirow{2}{*}{{Method}}&\multirow{2}{*}{{epochs}}&\multicolumn{2}{c}{\textbf{\emph{mini}ImageNet}}&\multicolumn{2}{c}{\textbf{\emph{tiered}ImageNet}}
  \\ \cline{3-6}
  &&{{5-way 1-shot}}&{ {5-way 5-shot}}&{{5-way 1-shot}}&{ {5-way 5-shot}} \\ 
  \hline
  SUN Meta-Training&300&{63.66$\pm$0.45}&{80.14$\pm$0.32}&{72.26$\pm$0.49}&{86.47$\pm$0.34} \\ 
  SUN Meta-Training&800&\textbf{64.84$\pm$0.45}&\textbf{80.96$\pm$0.32}&\textbf{72.34$\pm$0.49}&\textbf{86.57$\pm$0.34} \\
  \toprule
  \end{tabular}
  \end{threeparttable}}}
\end{table}
\subsection{More Training Epochs in Meta-Training}
Moreover, we also investigate the effect of training epochs during meta-training phase. The motivation comes from two aspects: 1) during meta-training, vanilla ViT faces severe generalization problems after 300 epochs training while ResNet-12 is opposite (see Fig.~2 in the main paper for details), 2) after fixing the generalization issue, ViTs with large-scale training tend to benefit from more training epochs. Thus we aim to investigate whether ViT with SUN benefits from more training epochs. 

Following the widely used 300/800 training epochs in previous ViT works~\cite{caron2021emerging,bao2022beit}, we evaluate our SUN meta-training with 300/800 epochs on \emph{mini}ImageNet (small size) and \emph{tiered}ImageNet (large size). And during testing, we follow~\cite{Chen_2021_ICCV_metabaseline} to evaluate the 5-way 1-shot and 5-way 5-shot accuracy on corresponding meta-learner $f$. 
Evaluation results are given in Table~\ref{table:epochs}. For \emph{mini}ImageNet, after introducing more training epochs, ViT with SUN outperforms that with 300 epochs by 1.2\% and 0.8\% in terms of 5-way 1-shot and 5-way 5-shot accuracy. This observation indicates that more training epochs can improve the generalization ability on novel categories of ViT with SUN. And for the relative larger \emph{tiered}ImageNet, introducing more epochs only leads to 0.1\% accuracy improvement. 
A possible explanation is that larger dataset may inherently benefit to the generalization ability, thus the contribution from more training epochs is limited. 
Therefore, for \emph{mini}ImageNet and CIFAR-FS, we use 800 epochs for meta-training phase; and for \emph{tiered}ImageNet, to trade off the training time as well as the classification accuracy, we keep using 300 epoch during meta-training. 

\begin{table}[t]
  \centering
  \caption{Comparison between SUN meta-training phase without global JS-Divergence constraint and SUN meta-training phase with global JS-Divergence constraint on \emph{mini}ImageNet, where ``JSD'' means further adding JS-Divergence between $g_{\scriptsize{0}}(\mathbf{z}_{0,\text{global}})$ and $g_{\scriptsize{\text{global}}}(\mathbf{z}_{\text{global}})$ in the meta-training phase. 
  }
  \label{table:usingjsd}
  \small
  \setlength{\tabcolsep}{4.0pt} 
 \renewcommand{\arraystretch}{4}
{ \fontsize{8.3}{3}\selectfont{
  \begin{threeparttable}
  \begin{tabular}{cccc}
  \toprule
  \multirow{2}{*}{{Method}}&\multirow{2}{*}{{JSD}}&\multicolumn{2}{c}{\textbf{\emph{mini}ImageNet}}
  \\ \cline{3-4}
  &&{{5-way 1-shot}}&{ {5-way 5-shot}} \\ 
  \hline
  SUN Meta-Training&&\textbf{64.84$\pm$0.45}&\textbf{80.96$\pm$0.32} \\
  SUN Meta-Training&\checkmark&{64.56$\pm$0.44} (-0.28\%)&{80.76$\pm$0.31} (-0.20\%) \\ 
  \toprule
  \end{tabular}
  \end{threeparttable}}}
\end{table}
\subsection{Is Adding JS-Divergence (JSD) Benefit to SUN? }
Denote by $f_{g}$ the given teacher model consisting of a feature extractor $f_{0}$ and a classifier $g_{0}$, and $f$ the target meta-learner with global classifier $g_{\text{global}}$.
Following the description in Sec.~4.1, we denote the the classification result from the global average pooling of all patch tokens calculated by $g_{0}$ is $g_{\scriptsize{0}}(\mathbf{z}_{0,\text{global}})$. 
Analogously, that calculated by target classifier $g_{\text{global}}$ is denoted by $g_{\scriptsize{\text{global}}}(\mathbf{z}_{\text{global}})$. 
Previous knowledge distillation works~\cite{hinton2015distill,tian2020rethink} mainly focus on minimizing the JS-Divergence between $g_{\scriptsize{0}}(\mathbf{z}_{0,\text{global}})$ and $g_{\scriptsize{\text{global}}}(\mathbf{z}_{\text{global}})$ and achieve better classification performance. 
Therefore, we focus on the meta-training phase of our SUN and conduct the ablation study of JS-Divergence on \emph{mini}ImageNet to evaluate whether it is essential for SUN or not. 
Specifically, during meta-training without JSD, we keep using $\mathcal{L}_{\text{SUN}}$ as training loss; while during meta-training with JSD, 
we use $\mathcal{L}_{\text{SUN+JSD}}=\mathcal{L}_{\text{SUN}}+\text{JSD}(g_{\scriptsize{0}}(\mathbf{z}_{0,\text{global}}), g_{\scriptsize{\text{global}}}(\mathbf{z}_{\text{global}}))$ as training loss. 
And during testing, we follow~\cite{Chen_2021_ICCV_metabaseline} to evaluate the 5-way 1-shot and 5-way 5-shot accuracy on corresponding meta-learner $f$. The evaluation results are shown in Table~\ref{table:usingjsd}. After introducing the global JS-Divergence constraint, meta-training with JSD drops 0.3\% and 0.2\% in terms of 5-way 1-shot and 5-way 5-shot accuracy.  
A possible explanation is that adding JS-Divergence may enforce the $g_{\scriptsize{\text{global}}}(\mathbf{z}_{\text{global}})$ to be similar to $g_{\scriptsize{0}}(\mathbf{z}_{0,\text{global}})$, thus somewhat ignores some knowledge from the location-specific supervision, and then slightly impairs the generalization ability on novel classes. 
Thus, in the final SUN framework, the global JS-Divergence constraint is not included.

\section{t-SNE Visualization Results}\label{sec:visualization}
\begin{figure*}[!htb]
  \centering
  \subfigure[\emph{train}, \emph{mini}, Baseline]{
          \includegraphics[width=1.5in]{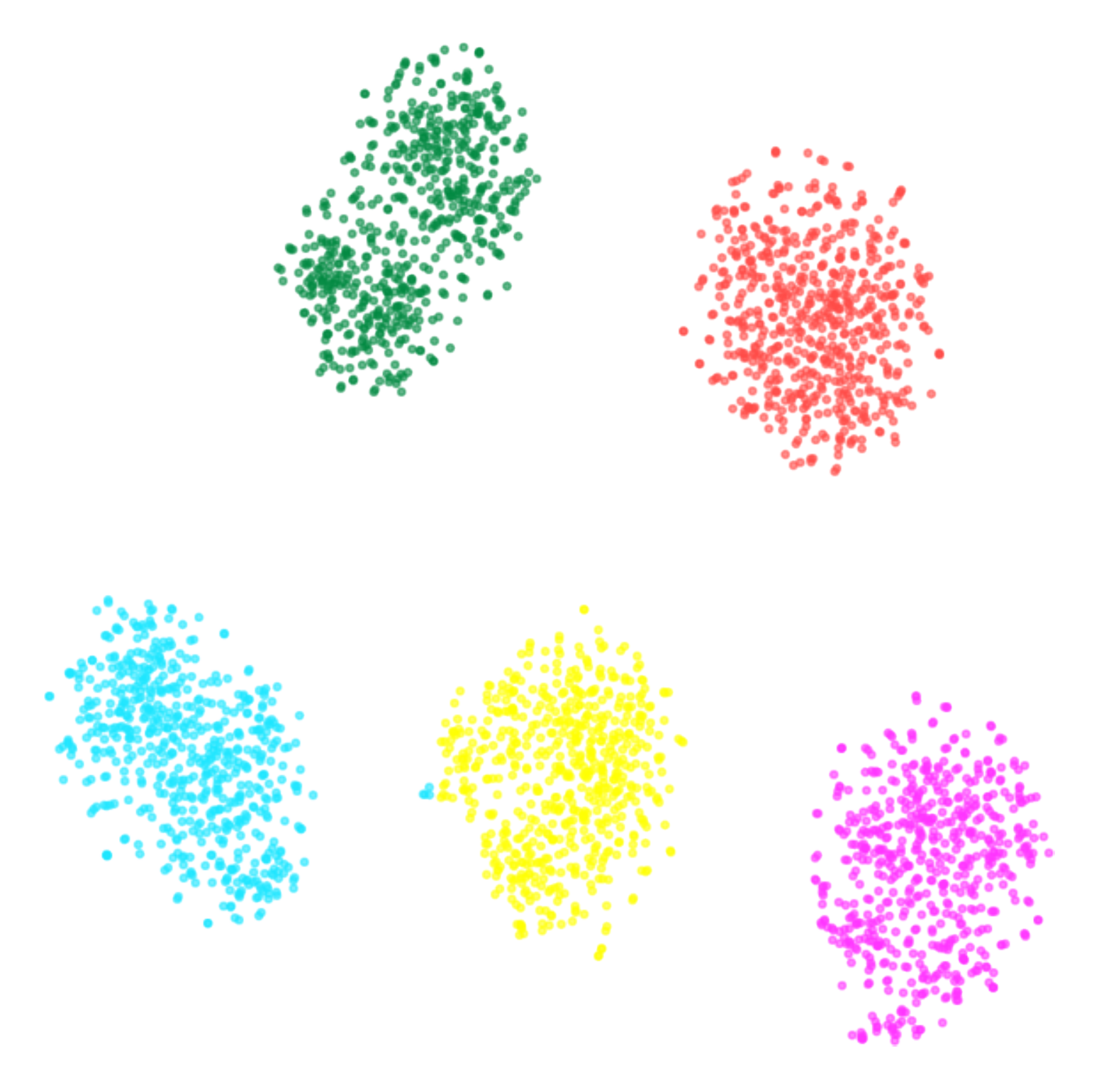}
          \label{fig:tsne_baseline_train_mini}
  } \hspace{-4mm} \subfigure[\emph{train}, \emph{tiered}, Baseline]{
      \includegraphics[width=1.5in]{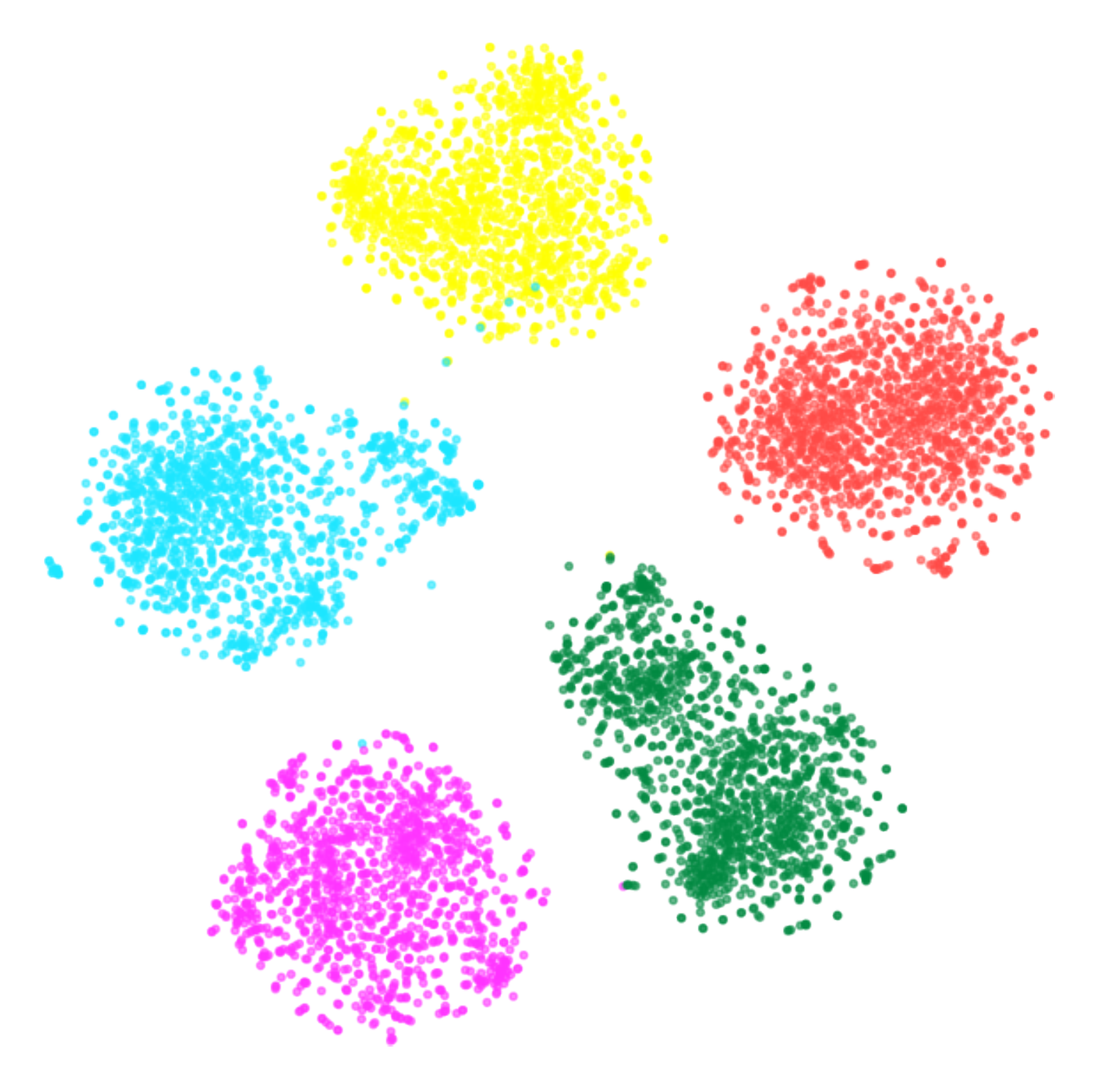}
      \label{fig:tsne_baseline_train_tiered}
  } \hspace{-4mm} \subfigure[\emph{train}, CIFAR, Baseline]{
      \includegraphics[width=1.5in]{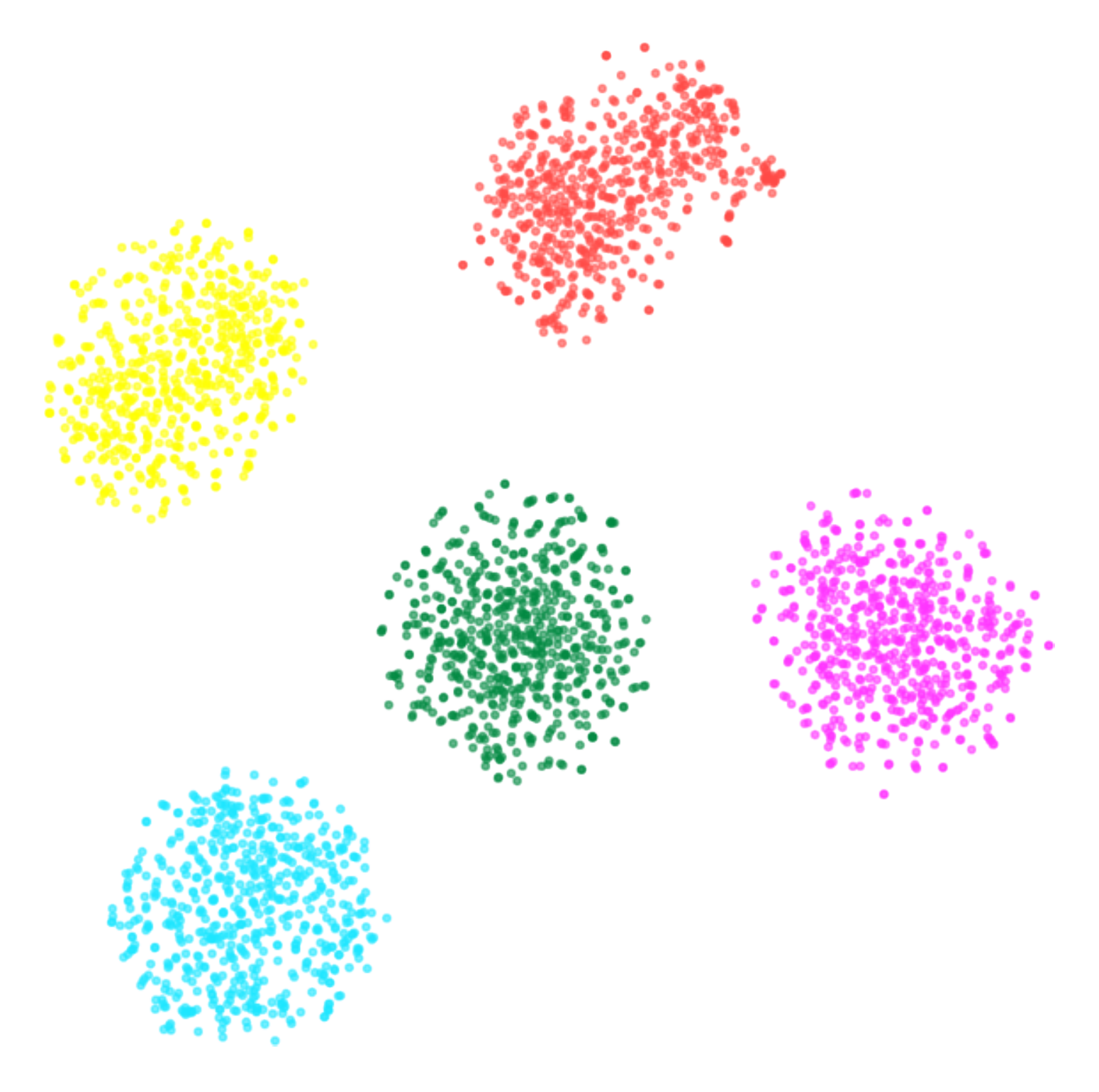}
          \label{fig:tsne_baseline_train_cifarfs}
  } \\ \vspace{-1em}
  \subfigure[\emph{train}, \emph{mini}, \textbf{SUN}]{
          \includegraphics[width=1.5in]{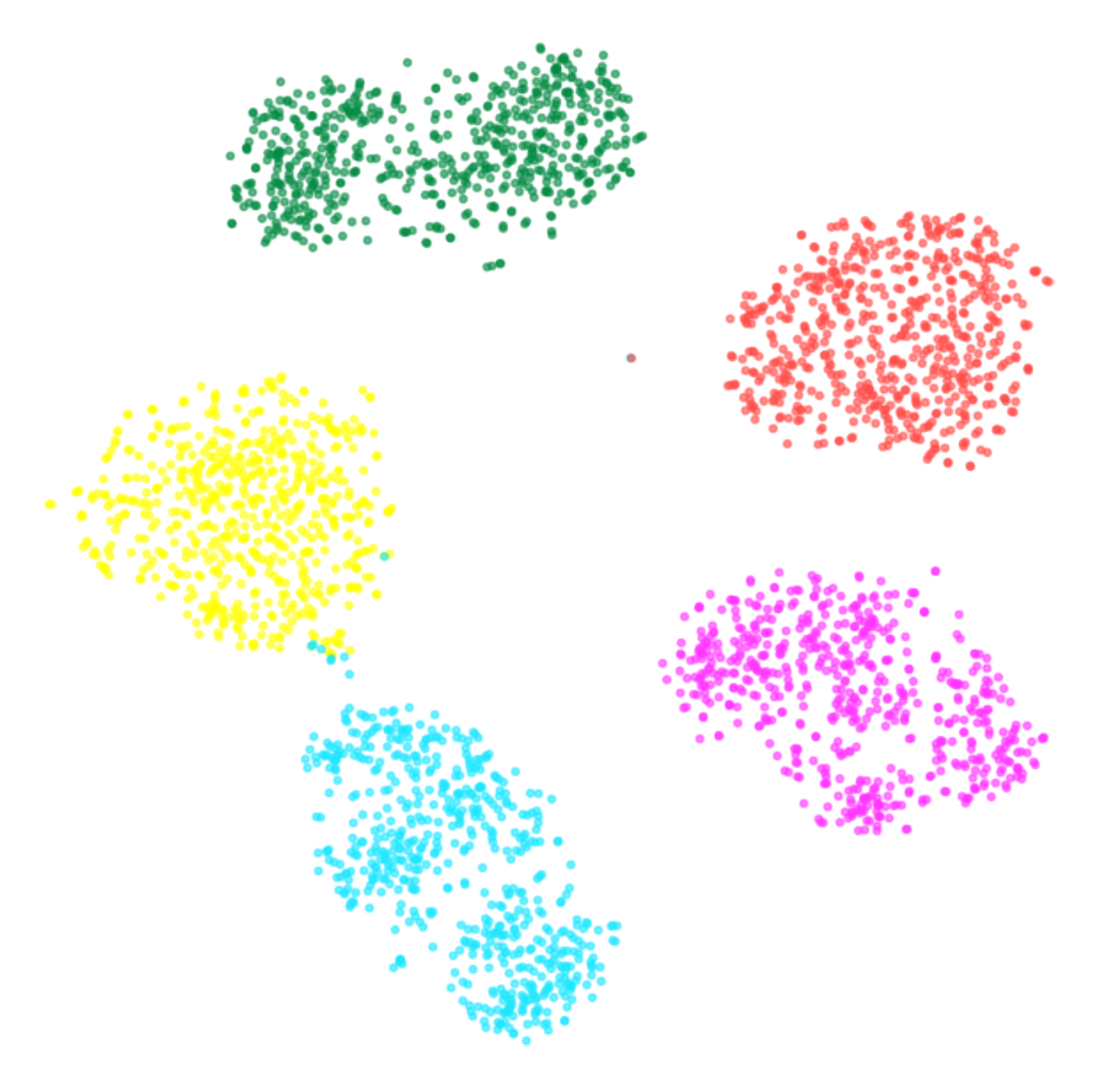}
          \label{fig:tsne_sun_train_mini}
  } \hspace{-4mm} \subfigure[\emph{train}, \emph{tiered}, \textbf{SUN}]{
      \includegraphics[width=1.5in]{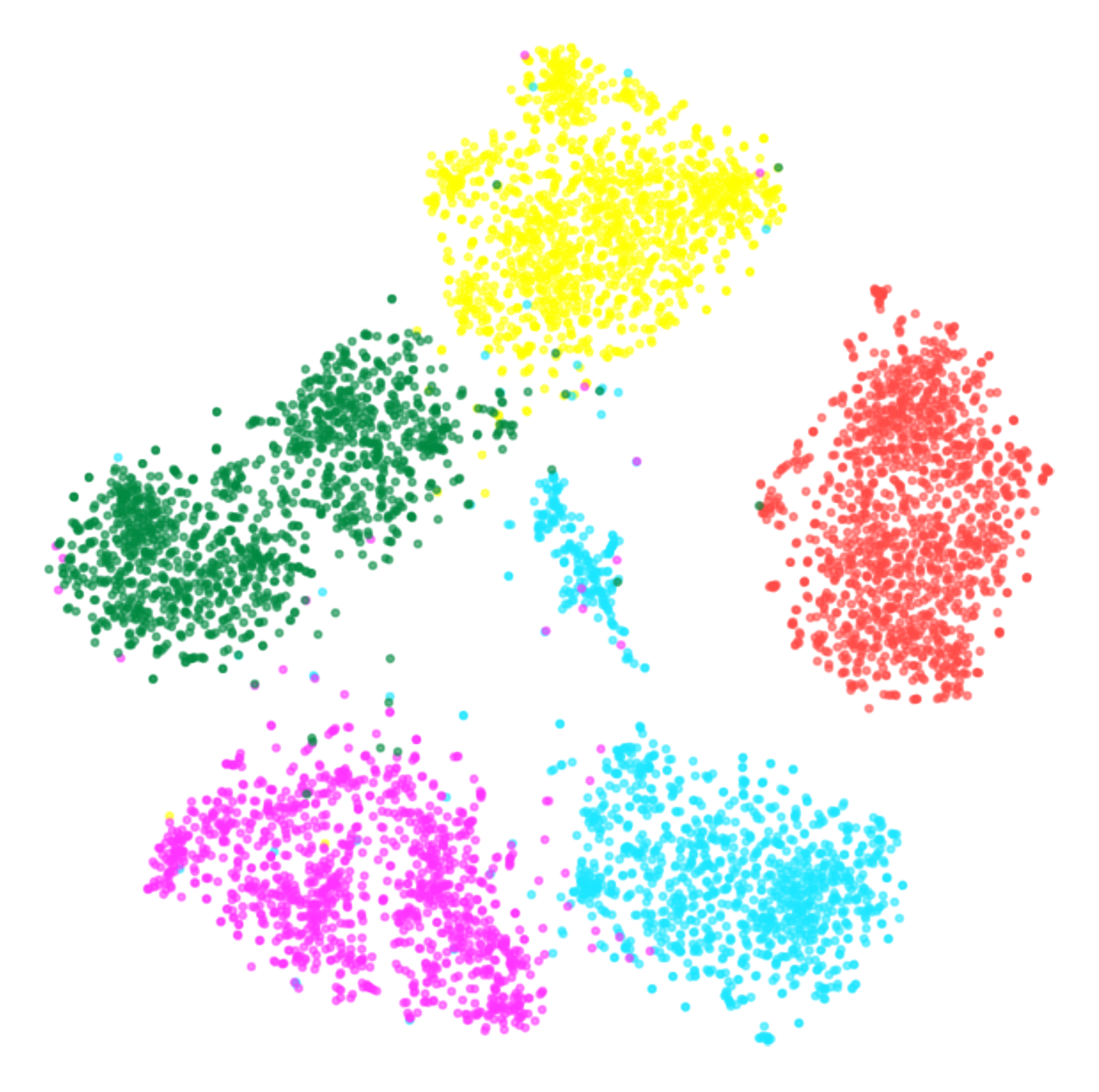}
      \label{fig:tsne_sun_train_tiered}
  } \hspace{-4mm} \subfigure[\emph{train}, CIFAR, \textbf{SUN}]{
      \includegraphics[width=1.5in]{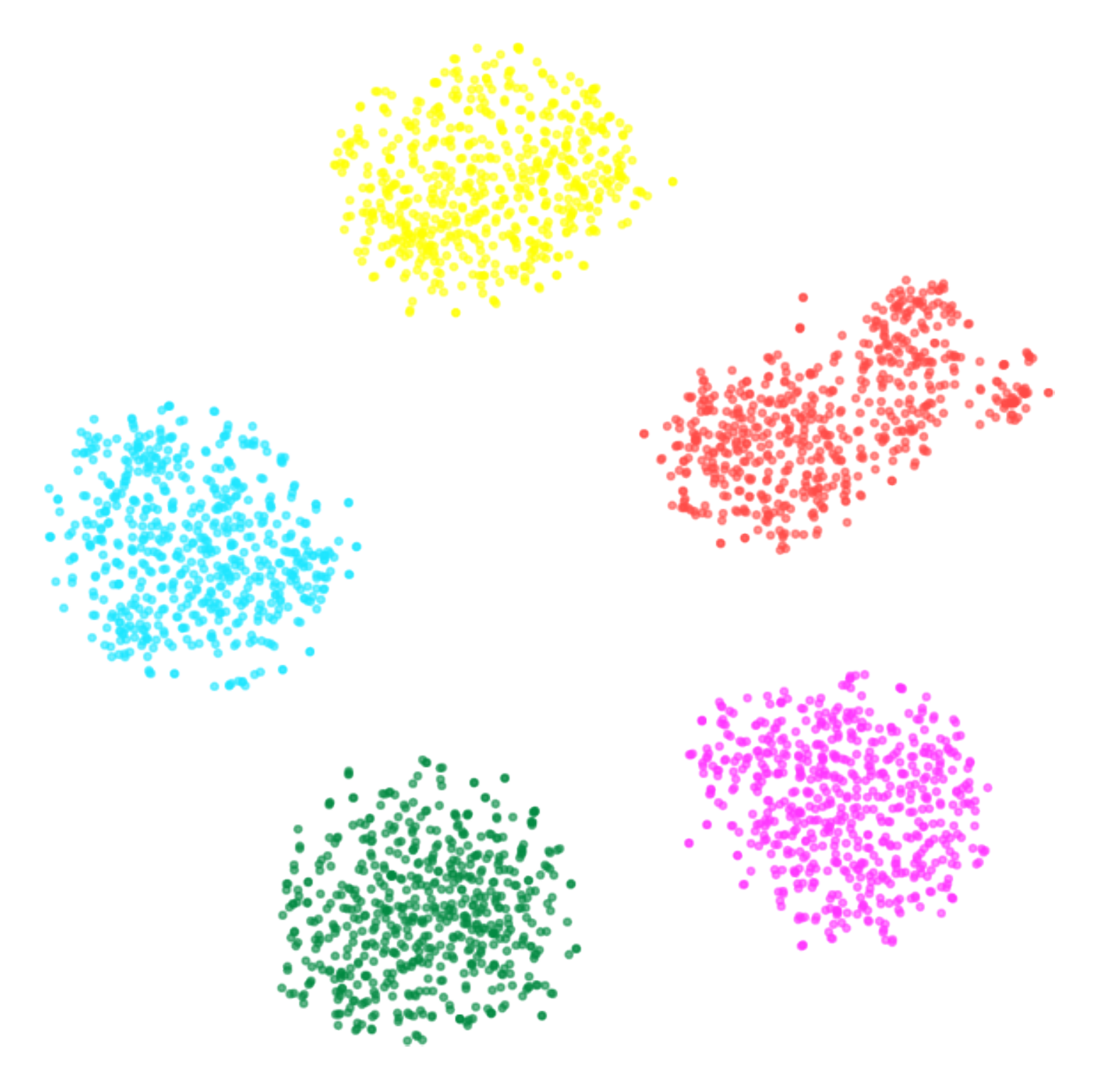}
          \label{fig:tsne_sun_train_cifarfs}
  } \\ \vspace{-1em}

  \subfigure[\emph{test}, \emph{mini}, Baseline]{
          \includegraphics[width=1.5in]{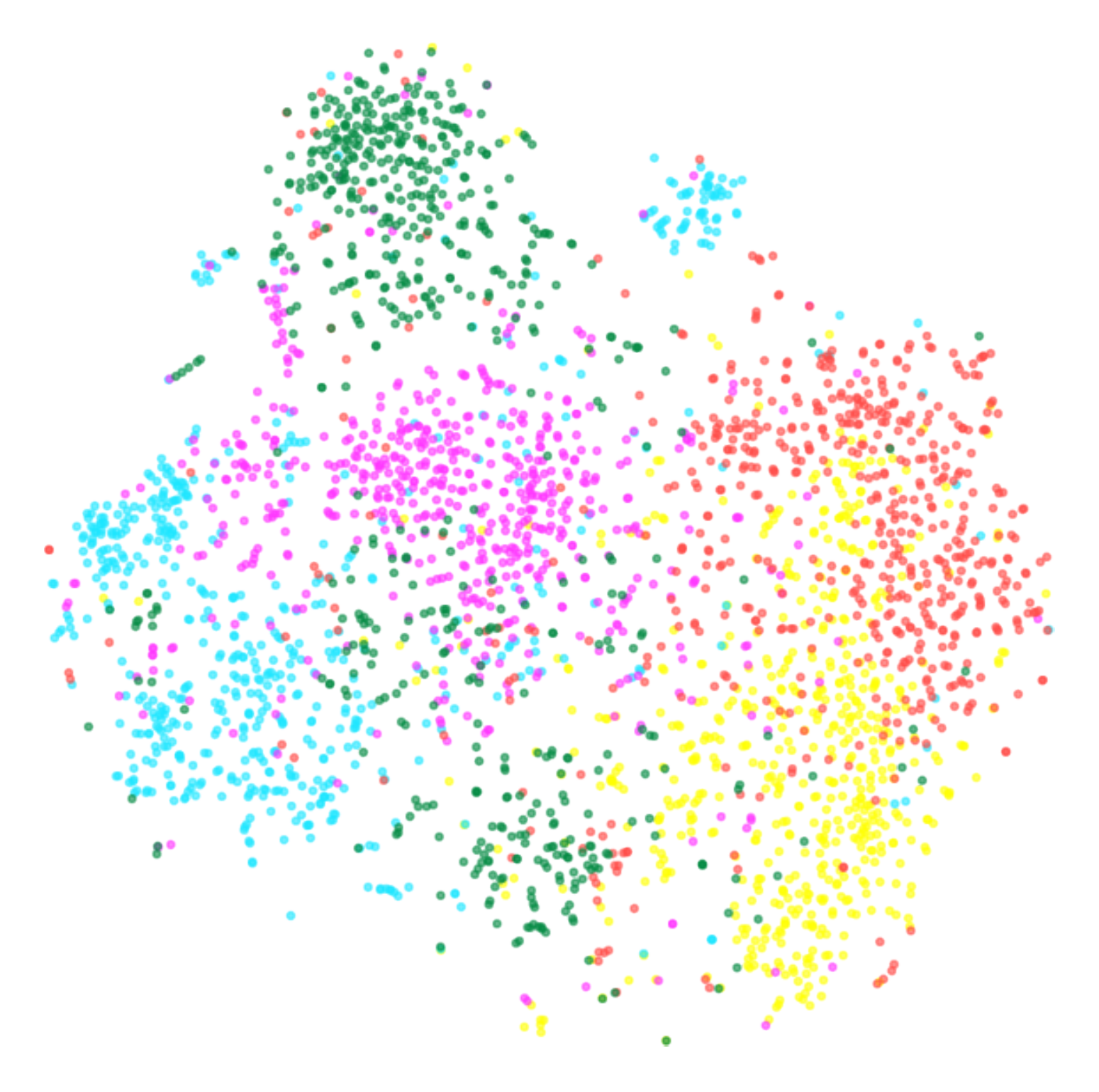}
          \label{fig:tsne_baseline_test_mini}
  } \hspace{-4mm} \subfigure[\emph{test}, \emph{tiered}, Baseline]{
      \includegraphics[width=1.5in]{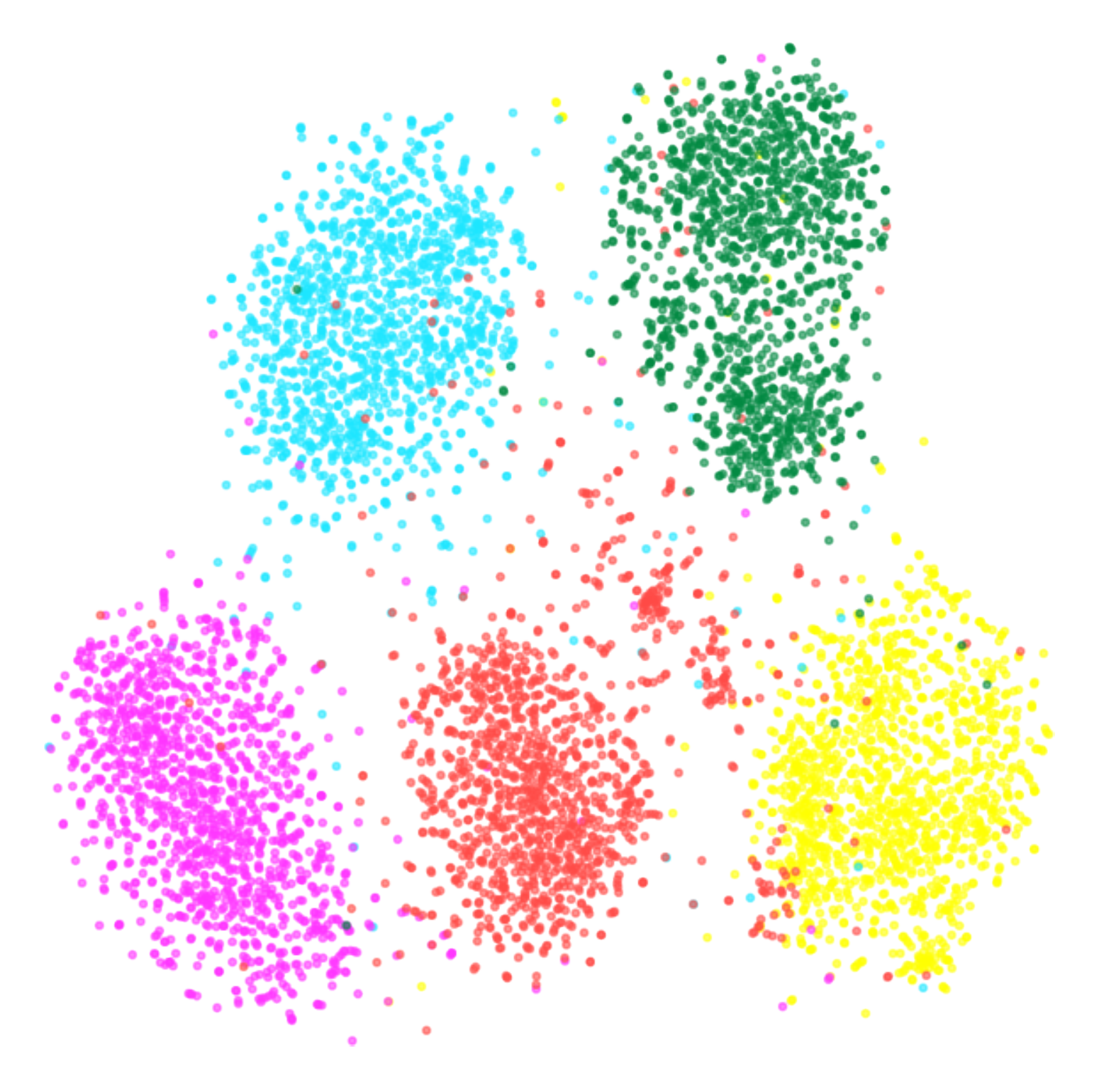}
      \label{fig:tsne_baseline_test_tiered}
  } \hspace{-4mm} \subfigure[\emph{test}, CIFAR, Baseline]{
      \includegraphics[width=1.5in]{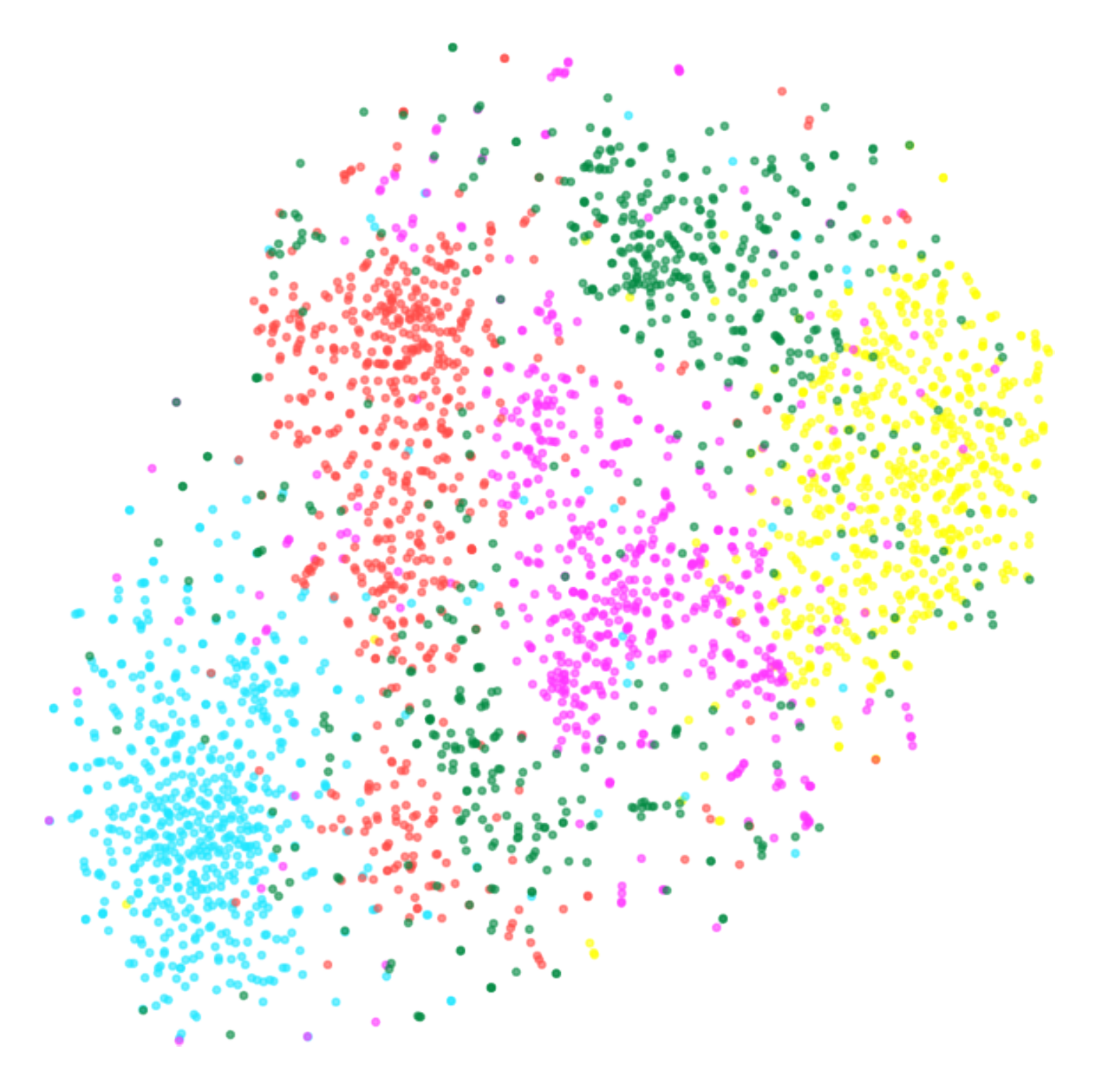}
          \label{fig:tsne_baseline_test_cifarfs}
  } \\ \vspace{-1em}
  \subfigure[\emph{test}, \emph{mini}, \textbf{SUN}]{
          \includegraphics[width=1.5in]{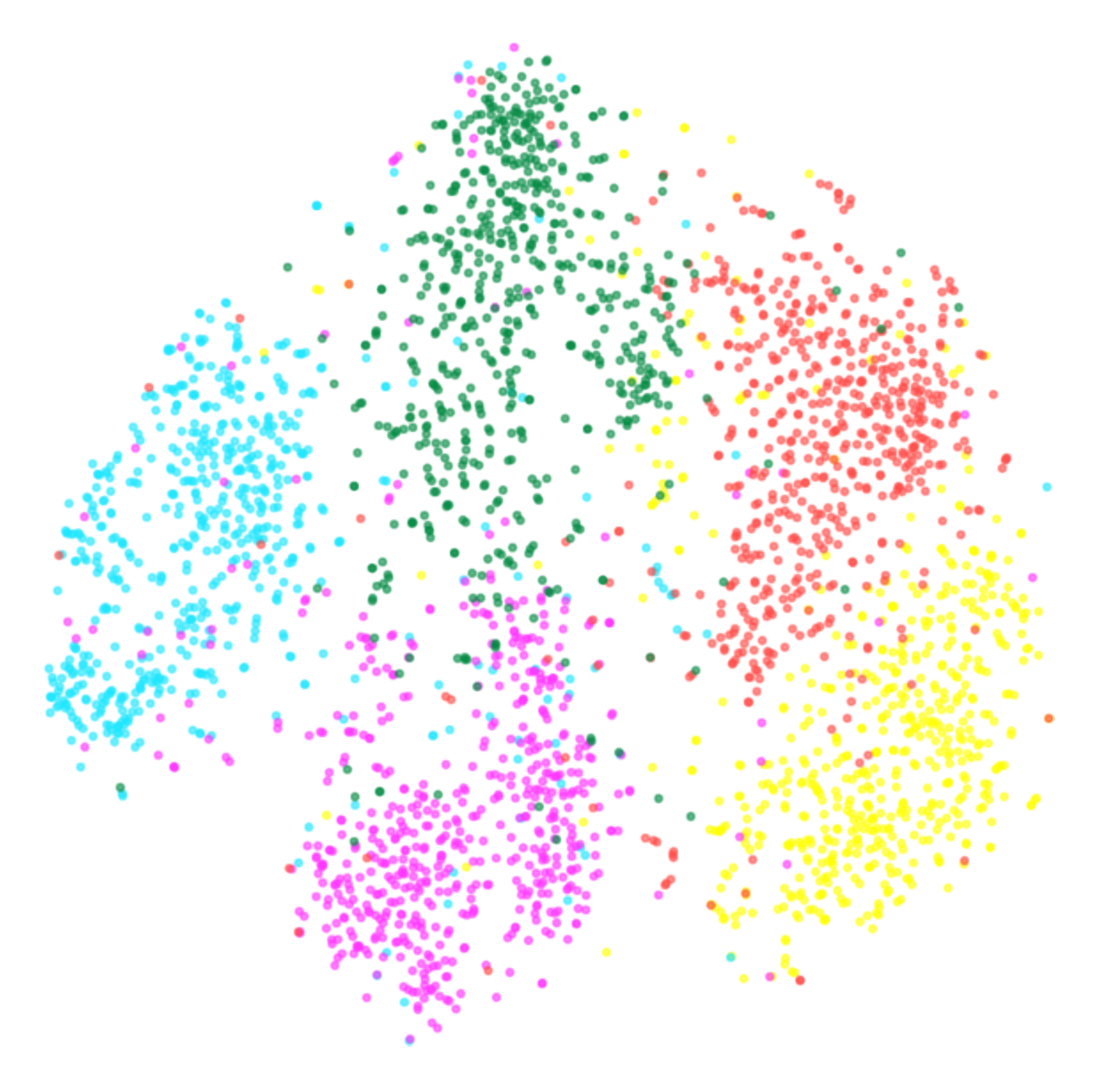}
          \label{fig:tsne_sun_test_mini}
  } \hspace{-4mm} \subfigure[\emph{test}, \emph{tiered}, \textbf{SUN}]{
      \includegraphics[width=1.5in]{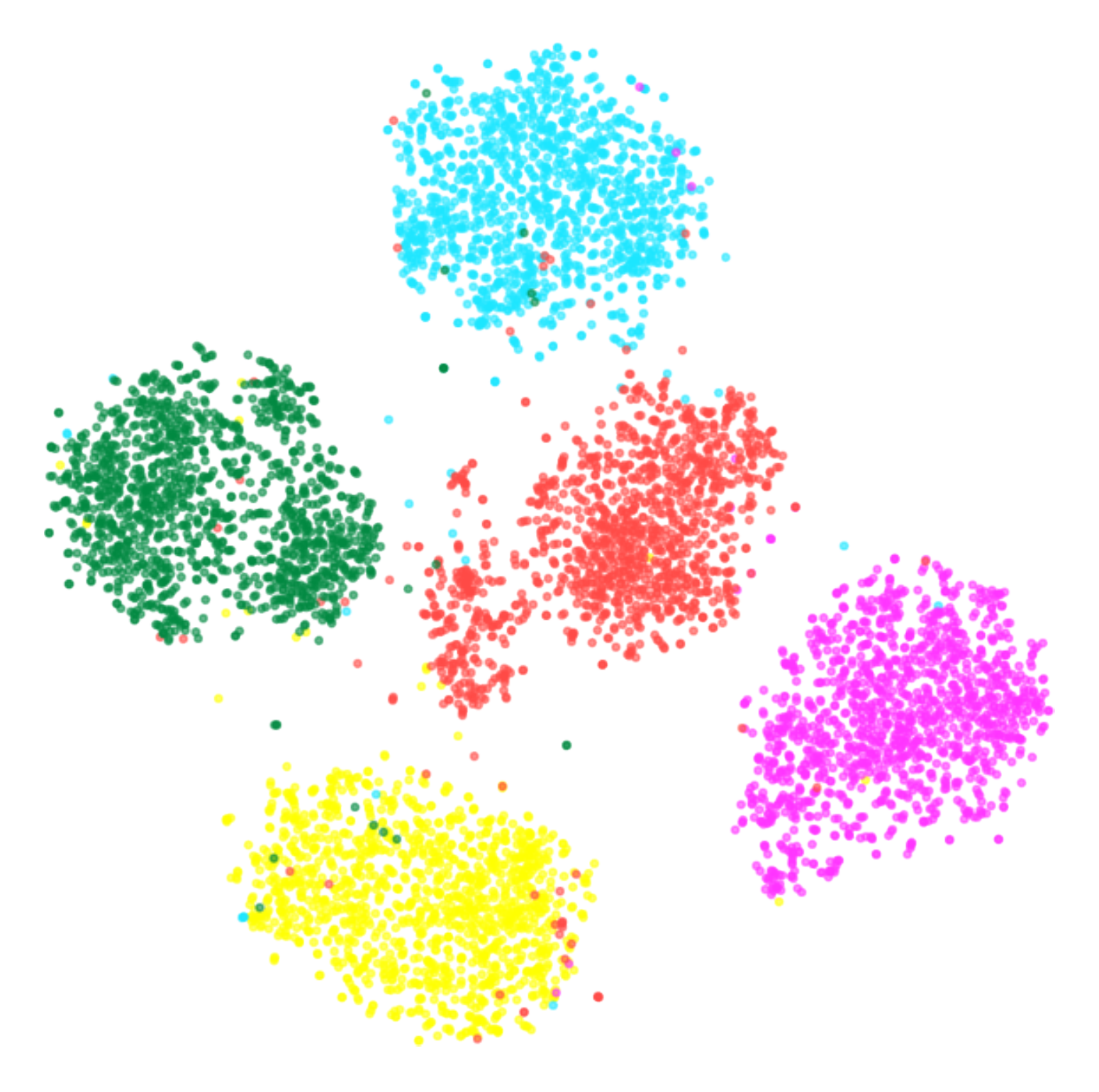}
      \label{fig:tsne_sun_test_tiered}
  } \hspace{-4mm} \subfigure[\emph{test}, CIFAR, \textbf{SUN}]{
      \includegraphics[width=1.5in]{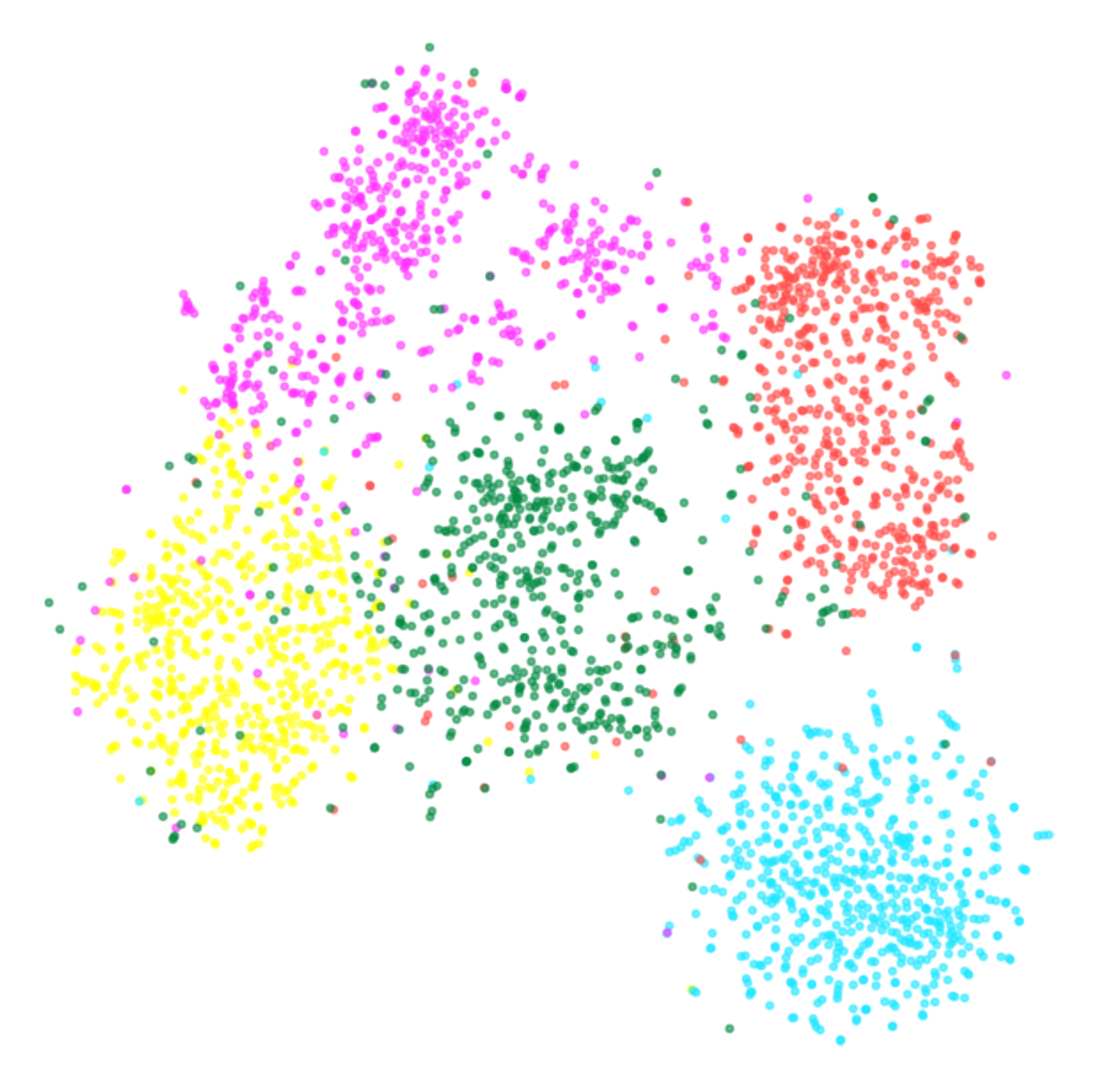}
          \label{fig:tsne_sun_test_cifarfs}
  } \\
  \caption{t-SNE visualization results of ViT without SUN (i.e. Baseline) and ViT with SUN (i.e. \textbf{SUN}) on three different datasets, where fig~(a)$\sim$(f) demonstrate the results of base classes from \emph{train} set and fig~(g)$\sim$(l) demonstrate those of novel classes from \emph{test} set. ViT with SUN performs better on all datasets.}
  \label{fig:tsne_fig}
\end{figure*}

%
Additionally, to qualitatively analyze the effects of our SUN, we also illustrate the t-SNE~\cite{vandermaaten08tsne} results of ViT with Meta-Baseline~\cite{Chen_2021_ICCV_metabaseline} (short for ``Baseline'' in Fig.~\ref{fig:tsne_fig}) and ViT with our SUN (short for ``\textbf{SUN}'' in Fig.~\ref{fig:tsne_fig}). Following Sec.~5.2 and Sec.~5.5 in the main paper, we use NesT~\cite{zhang2021aggregating} as our ViT feature extractor.

Detailed visualization results are shown in Fig.~\ref{fig:tsne_fig}.
Specifically, for each comparison pair (\eg, Fig.~\ref{fig:tsne_fig}(g) and Fig.~\ref{fig:tsne_fig}(j)), we keep using the same 5 categories for visualization. And for each category, we select the same 300 samples for visualization. According to these t-SNE visualization results, our ViT with SUN achieves the similar embedding grouping ability on base classes from the training set, but obtains better grouping ability for novel classes on test set. Especially, as shown in Fig.~\ref{fig:tsne_fig}(g) and Fig.~\ref{fig:tsne_fig}(j), the embeddings from novel classes extracted by ``baseline'' tend to be mixed together, but those from ViT with ``SUN'' can be separated into approximately 5 different groups. These observations also demonstrate the generalization ability of ViT with SUN on both base and novel categories, further verifying the effectiveness of SUN. 
\clearpage
\vspace{-2em}
\end{document}